\pdfoutput=1

\documentclass[11pt]{article}

\usepackage[final]{acl}

\usepackage{times}
\usepackage{latexsym}

\usepackage[T1]{fontenc}

\usepackage[utf8]{inputenc}

\usepackage{microtype}

\usepackage{inconsolata}

\usepackage{graphicx}

\usepackage[normalem]{ulem}

\usepackage{caption}
\usepackage{subcaption}
\usepackage{booktabs}
\usepackage{colortbl}
\usepackage{multirow}
\usepackage{hyperref}
\usepackage[normalem]{ulem}
\usepackage{footmisc}
\usepackage{tabularx}
\usepackage{array} 

\usepackage{xurl}
\usepackage{pdflscape}
\usepackage{geometry}
\usepackage{longtable}
\usepackage{enumitem}
\usepackage{tipa}
\usepackage{xcolor}
\usepackage{lipsum}

\usepackage{amssymb}
\usepackage{amsmath}
\usepackage{amsfonts}

\newcommand{\todo}[1]{\textcolor{blue}{\textbf{TODO}}}
\newcommand{\done}[1]{\textcolor{brown}{\textbf{DONE}}}

\usepackage{soul}
\definecolor{highlightBlue}{HTML}{00FFFF} 
\definecolor{highlightRed}{HTML}{EA9999} 
\useunder{\uline}{\ul}{}

\newcommand{\abbrvmetric}[1]{ILO}

\title{High-Dimensional Interlingual Representations of \\Large Language Models}

\author{
    \textbf{Bryan Wilie$^{\dagger}$},
    \textbf{Samuel Cahyawijaya$^{\ddagger}$},
    \textbf{Junxian He$^{\dagger}$},
    \textbf{Pascale Fung$^{\dagger}$}
    \\
    $^{\dagger}$Hong Kong University of Science and Technology $^{\ddagger}$Cohere\\
    \texttt{bwilie@connect.ust.hk} 
}

\begin{document}
\maketitle
\begin{abstract}
Large language models (LLMs) trained on massive multilingual datasets hint at the formation of interlingual constructs--a shared region in the representation space. However, evidence regarding this phenomenon is mixed, leaving it unclear whether these models truly develop unified interlingual representations, or present a partially aligned constructs. We explore 31 diverse languages varying on their resource-levels, typologies, and geographical regions; and find that multilingual LLMs exhibit inconsistent cross-lingual alignments. To address this, we propose an interlingual representation framework identifying both the shared interlingual semantic region and fragmented components, existed due to representational limitations. We introduce Interlingual Local Overlap (\abbrvmetric{}) score to quantify interlingual alignment by comparing the local neighborhood structures of high-dimensional representations.
We utilize \abbrvmetric{} to investigate the impact of single-language fine-tuning on the interlingual alignment in multilingual LLMs. Our results indicate that training exclusively on a single language disrupts the alignment in early layers, while freezing these layers preserves the alignment of interlingual representations, leading to improved cross-lingual generalization.
These results validate our framework and metric\footnote{\url{https://github.com/HLTCHKUST/interlingua}} for evaluating interlingual representation, and further underscore that interlingual alignment is crucial for scalable multilingual learning.
\end{abstract}

\section{Introduction}

\begin{figure}[!t]
    \centering
    \includegraphics[trim={7, 7, 7, 7}, clip, width=0.95\columnwidth]{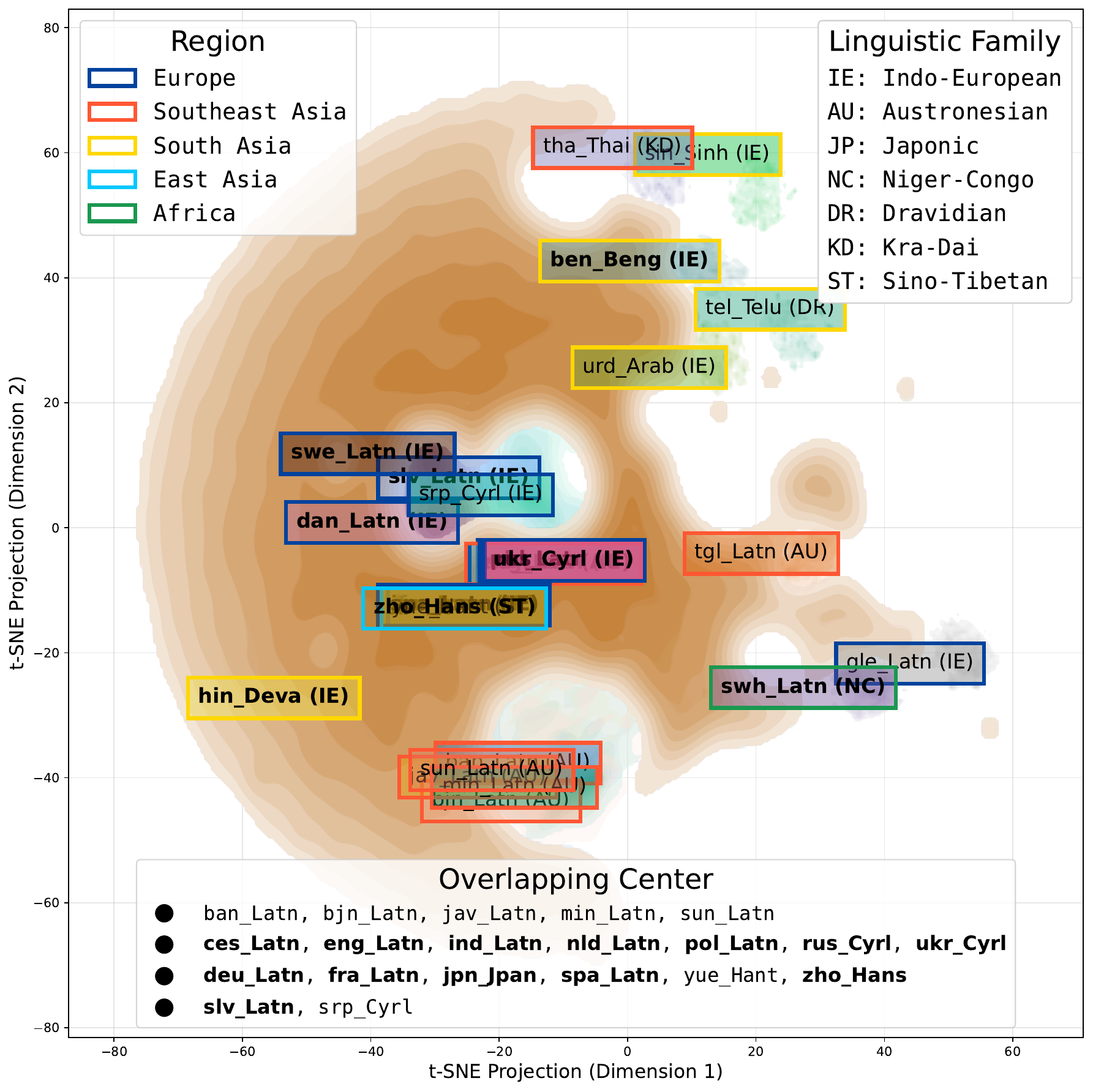}
    \vspace{-5pt}
    \caption{
    Interlingual overlaps transcending familial and regional boundaries in the intermediate layer, observed in a t-SNE visualization on the middle layer (16) of Aya Expanse (8B) hidden-state embeddings (HRLs in \textbf{bold}).
    }
    \label{fig:tsne_1}
    \vspace{-10pt}
\end{figure}

Interlingua, a universal language-neutral representation, is pivotal for cross-lingual generalization. Grounded in both linguistic theories and computational practice, this concept aims to treat languages equitably and capture universal semantic structures independent of any specific language~\cite{richens1958interlingualmt,Vauquois1968ASO,schubert1989interlinguistics,rayner2010bootstrapped, johnson2017google}. 
The advent of LLMs trained on extensive multilingual corpora suggests the potential of interlingual constructs naturally emerging without any explicit objectives~\cite{conneau-etal-2020-unsupervised,chang-etal-2022-geometry,moschella2023relative,wendler-etal-2024-llamas}. 
This is attributed to their ability to map representations from different languages into a shared multilingual representation space~\cite{pires-etal-2019-multilingual,libovicky2020language,conneau2020emerging,muller2021first,zhao2024how,zeng2025converging}.

However, evidence remains mixed on whether they converge all language-specific representations into a unified single interlingual representation space, and
raising questions about whether LLMs can retain the interlingual representations in diverse linguistic typology, geographical distribution, and resource-level settings. It is unclear whether LLMs form a unified interlingual construct or if fragmentation occurs across different language groups. A critical question persists: Do LLMs develop a universal interlingua representation, or present a partially aligned construct toward certain languages?

Our preliminary experiments reveal that LLMs represent parallel semantic input differently across languages. Notably, their neuron activations align better within high-resource pairs and the same familial or regional roots, suggesting that LLMs exhibit varying alignment consistencies across differing language groups.
Building upon these insights, we introduce a novel interlingual representation framework aimed at enhancing the understanding of how LLMs encapsulate interlingual semantics. Our framework identifies both the core region that captures shared semantics across languages, and addresses fragmented components due to representational limitations underscoring the importance of interlingual alignment across diverse linguistic contexts.
With the framework, we introduce a novel metric, Interlingual Local Overlap (\abbrvmetric{}), which quantifies intrinsic interlingual alignment consistencies by comparing the local neighborhood structures of high-dimensional representations.
Inspired by graph theory~\cite{guimera2005cartography,freeman2002centrality,borgatti2006graph}, the \abbrvmetric{} score is derived from the harmonic mean of two measurements, on 
the extent to which representations of a given language within the multilingual space: individually neighboring diverse other languages (\textbf{bridge}) and collectively connect diversely with other languages (\textbf{reachability}).

We demonstrate the effectiveness our framework and metric through an in-depth analysis of LLMs' internal states on a multilingual mathematical reasoning task, chosen for its language-agnostic properties. 
We first observe that training multilingual LLMs on a single-language causes catastrophic forgetting~\cite{mccloskey1989catastrophic, french1999catastrophic, biesialska2020continual} degrading their cross-lingual generalization~\cite{liu2021preserving, winata2023overcoming}. 
These degradations are correlated with the disruption of interlingual alignment that originate in the early layers of LLMs.
To ensure the preservation of interlingual alignments, we adopt a strategy of selectively-freezing parameters during the single-language fine-tuning. Evaluations using \abbrvmetric{} highlight that this approach effectively safeguards the interlingual alignments across all layers and maintains the levels observed prior,
which results in significant improvements in cross-lingual generalization. Ultimately, our findings 
underscore the pivotal role of interlingual semantic alignment in the pursuit of scalable multilingual learning. 

\begin{table}[!t]
\centering
\small
\resizebox{\columnwidth}{!}{%
\begin{tabularx}{\columnwidth}{p{1.5cm}X}
\toprule
\textbf{Properties} & \textbf{Details} \\
\midrule
Resources & High: 18 / Low: 13 \\
Families & Indo-European: 18 / Austronesian: 7 / Sino-Tibetan: 2 / Japonic: 1 / Niger-Congo: 1 / Dravidian: 1 / Kra-Dai: 1 \\
Regions & Europe: 14 / Southeast Asia: 8 / South Asia: 5 / East Asia: 3 / Africa: 1  \\
\bottomrule
\end{tabularx}
}
\vspace{-5pt}
\caption{Distribution of the 31 languages across families, regions, and resource-levels in our analysis, sampled from Flores-200 (see Appendix~\ref{app:lang_details} for complete details).}
\label{tab:anc_lang_stats}
\vspace{-10pt}
\end{table}

\begin{figure*}[!t]
  \centering
  \hspace*{-10pt}
  \begin{subfigure}[t]{0.48\linewidth}  
    \centering
      \includegraphics[trim={7, 7, 7, 7}, clip, width=\linewidth]{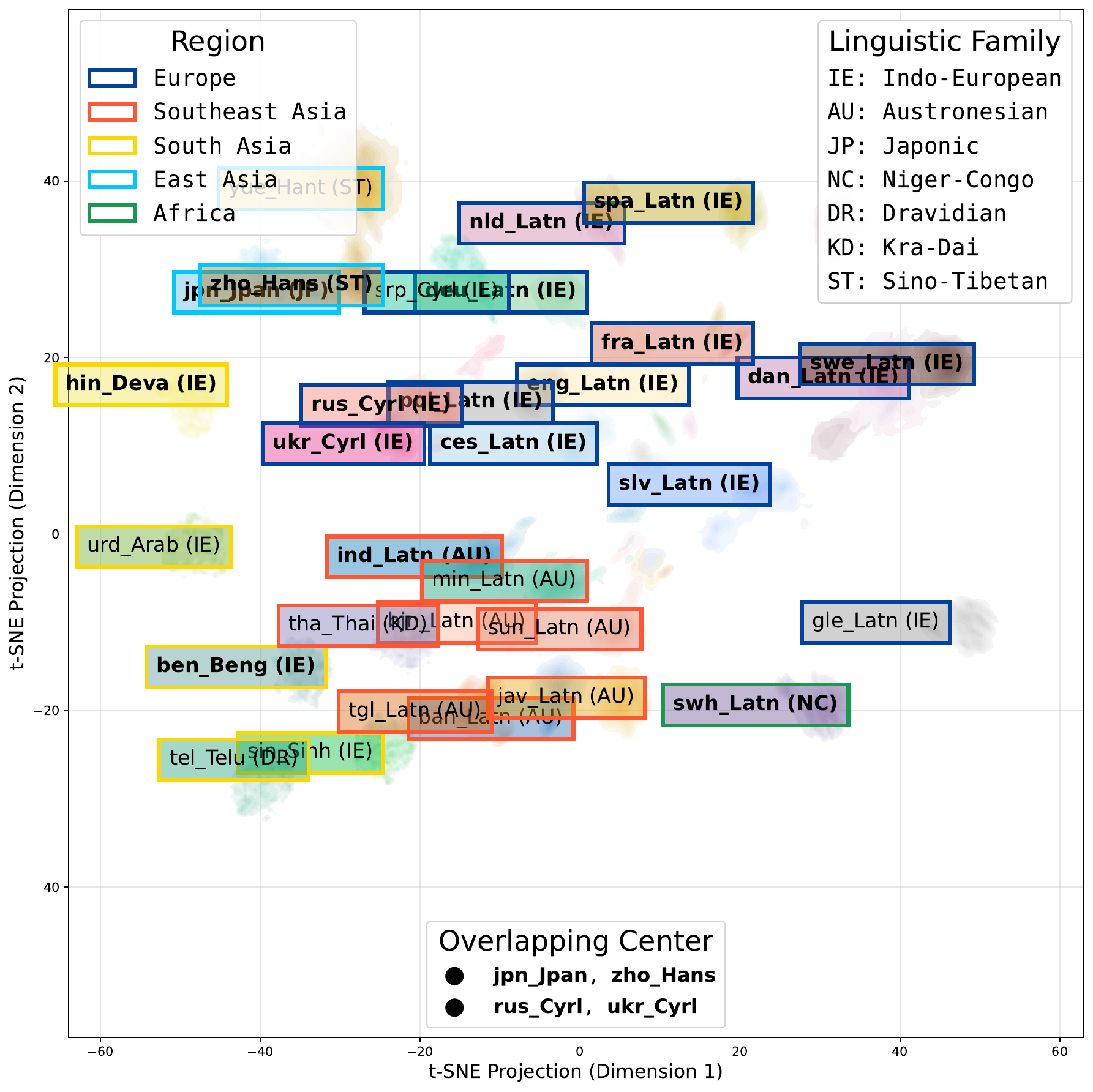}
      \vspace{-15pt}
      \caption{Early (Layer 0)}
      \vspace{-5pt}
      \label{fig:tsne_earlylayer}
  \end{subfigure}
  \hspace{5pt}
  \begin{subfigure}[t]{0.48\linewidth}
    \centering
      \includegraphics[trim={7, 7, 7, 7}, clip, width=\linewidth]{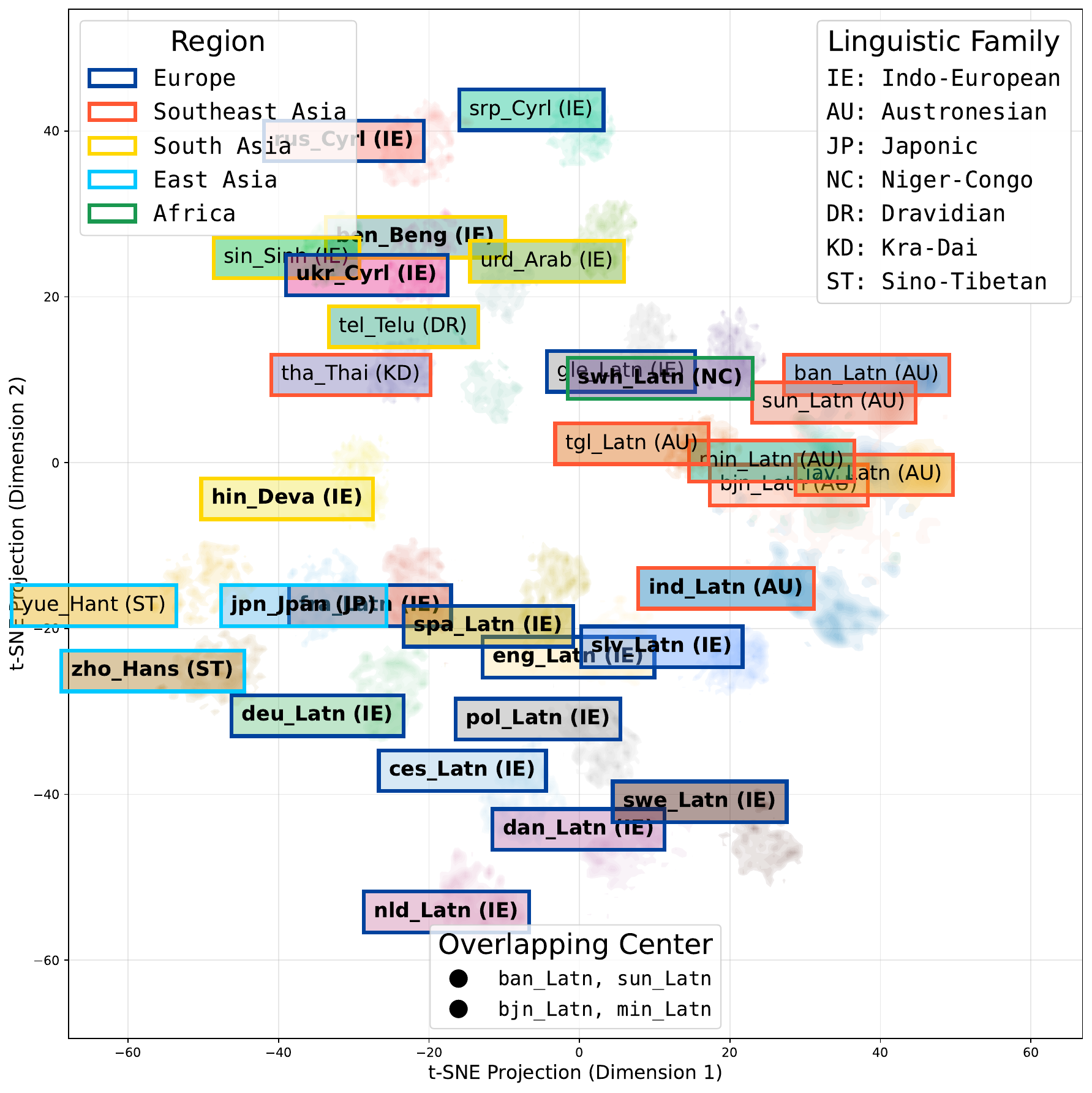}
      \vspace{-15pt}
      \caption{Late (Layer 32)}
      \vspace{-5pt}
      \label{fig:tsne_latelayer}
  \end{subfigure}
  \caption{Hidden-state embeddings of Aya Expanse (8B) projected in t-SNE dimensions (HRLs in \textbf{bold}). In these early and late layers, the representations cluster w.r.t resource levels and linguistic features, and minimally overlap.}
  \label{fig:tsne_2}
  \vspace{-10pt}
\end{figure*}

\section{Related Works}

\paragraph{Syntactical Interlingua Representations}

Interlingua has played a huge role throughout the development of NLP. Various representations of interlingua have been developed along with the advancement of NLP. In the early years, a logically formalized interlingua representation for mechanical translation has been proposed~\cite{richens1958interlingualmt,Vauquois1968ASO}. In the early days, interlingua is presented as delexicalized grammar extracted from the original text that can be mapped to other language interlingua delexicalized grammar. In this case, each language has its own interlingua form which can then be mapped into other language with a dictionary lookup~\cite{richens1958interlingualmt,rayner-etal-2010-bootstrapped}. A more sophisticated method involves interlingua representation as a common abstract syntax that are shared across all languages~\cite{rayner-etal-2008-almost,kanzaki-etal-2008-many}. This method has been applied in various systems such as Spoken Languge Translator~\cite{rayner2000spoken}, PARC’s XLE~\cite{riezler-etal-2002-parsing}, and Verbmobil~\cite{wahlster2013verbmobil}. Despite its advancement, this method tends to be incomplete and difficult to scale to new languages~\cite{ranta-etal-2020-abstract}.

\paragraph{Semantic Interlingua Representations}

With the rise of statistical machine translation~\cite{brown-etal-1990-statistical,och-etal-1999-improved,lopez2008pbsmt} and cross-lingual alignment~\cite{brown1991aligning,och-ney-2003-systematic,mikolov2013exploiting,miceli-barone-2016-towards,artetxe-schwenk-2019-massively}, methods for representing interlingua using latent semantic vectors become more prominent~\cite{fung2004biframenet,fung1994aligning,fung-church-1994-k,seneff-2006-combining}. Methods involving specialized objectives to construct better semantic interlingua representations have also been proposed~\cite{lu2018neural,al2019consistency, zhu2020language,wei2021learning,feng-etal-2022-language,cahyawijaya2023instructalign,cahyawijaya-etal-2024-llms}. In recent years, various studies have showcased that current LLMs inherit such interlingua representation~\cite{muller2021first,chang-etal-2022-geometry,moschella2023relative,zhao2024how,wendler-etal-2024-llamas} which enables LLMs to process sentences with a single shared representation across different languages.
However, the characteristics of this representation in LLMs remain unexplored. This research aims to explore the extent of this interlingua representation offering a novel perspective on interlingual representation in LLMs.

\section{Interlingual Representations in Multilingual LLMs}
\label{sec:theory}

To explore the emergence of interlingual representation in LLMs, we assess the semantic alignment of their hidden states to understand whether the latent structures capture universal semantics across languages. We presume that multilingual LLMs adhere to a ``first align, then predict'' pattern~\cite{muller2021first} and that their aligned states represent semantically similar features across languages.
Ideally, these features map parallel semantic inputs from many languages to similar vector representations that overlaps in the high-dimensional space.

Consider the high-dimensional representation space \(\mathcal{H} \subseteq \mathbb{R}^d\) learned by LLMs, where \(d\) is the model’s hidden-states dimension. For an input \(\mathbf{x}\) in language \(\ell\), the model uses language-specific encoding functions \(f_{\ell}(\mathbf{x}) \in \mathcal{H}\). Here, \(\mathcal{H}\) serves as a shared multilingual space where different encoding functions \(f_{\ell}(\mathbf{x})\) align semantic and syntactic patterns across languages. 
Building on this, we define semantic alignment \(\alpha\) of representations from parallel inputs \(\mathbf{x}\) and \(\mathbf{x}'\) in languages \(\ell\) and \(\ell'\) as:
\[
    \alpha(\ell, \ell') = \mathbb{E}_{(\mathbf{x}, \mathbf{x}') \sim \mathcal{D}_{\ell,\ell'}} \left[ \phi\left(f_\ell(\mathbf{x}), f_{\ell'}(\mathbf{x}')\right) \right].
\]
Here, \(\phi\) denotes a similarity function and \(\mathcal{D}_{\ell,\ell'}\) is the distribution of semantically equivalent input pairs. A higher \(\alpha(\ell, \ell')\) indicates better alignment.

\begin{figure*}[!t]
    \centering
    \includegraphics[trim={6, 8, -5, 7}, clip, width=\linewidth]    {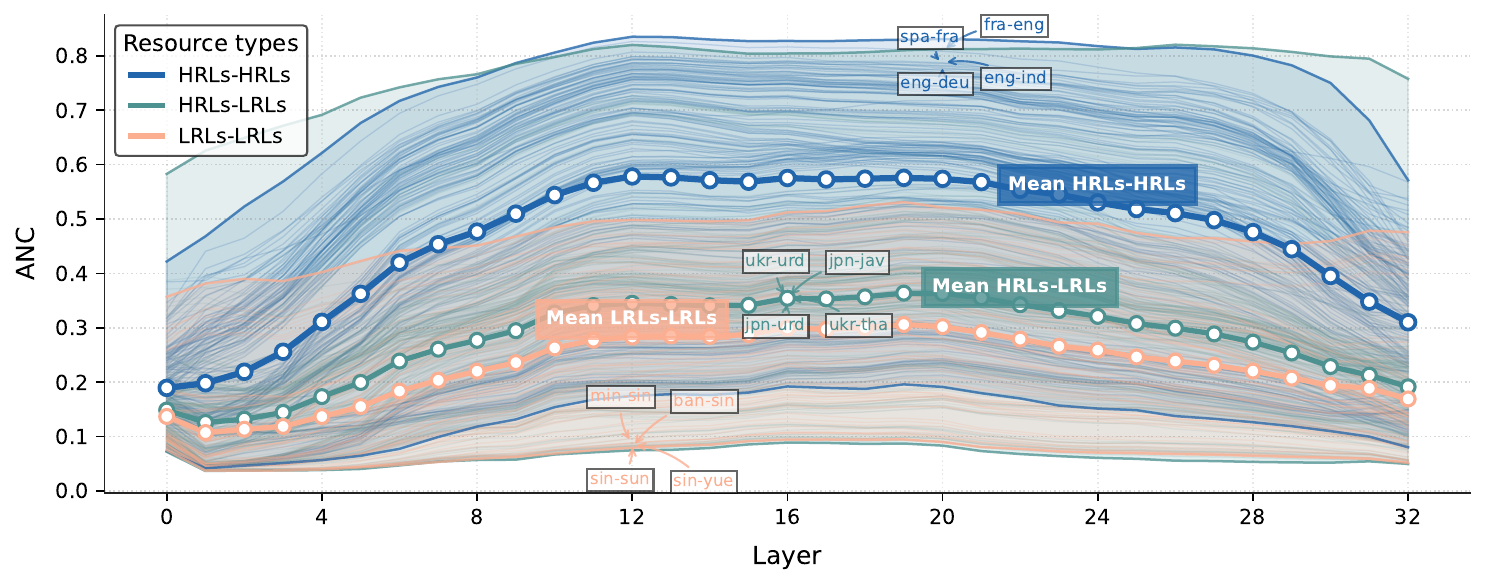}
    \caption{Comparisons of per-layer ANC scores on Aya Expanse (8B) with highlights on pairs w.r.t their resource-levels. Pairs of HRLs demonstrate strong correlations, while pairs involving LRLs exhibit lower ANC scores.}
    \vspace{-5pt}
    \label{fig:anc_hrllrl}
\end{figure*}

\subsection{Multilingual Shared Representation Space}

We posit an interlingual representation framework that incorporates an intricate internal structure influenced by inherent model representational limitations. This framework highlights that the quality of alignment among representations may vary, leading to latent discrepancies that may stem from differences in resource availability or language-specific properties. 
Formally, we conceptualize the representations from various languages as falling into one of two qualitative regions of \(\mathcal{H}\):
\[
\mathcal{H} \supset \mathcal{M}_c \cup \bigcup_{\ell\in F} \mathcal{M}_{f_\ell}.
\]
The component \(\mathcal{M}_c\) is an aligned core interlingual region, that predominantly encapsulates shared semantics across languages. In contrast, the fragmented \(\mathcal{M}_{f}\) represent regions where alignment with \(\mathcal{M}_c\) is challenging. This framework refines the ``first align, then predict'' paradigm, that while LLMs align inputs from languages to a shared interlingual region, some remain partially aligned.

\subsection{Core Interlingual Region} 
Conceptually, we define \(\mathcal{M}_{c}\) as a region that predominantly encodes universal semantic structures and syntactic abstractions. By positioning multilingual representations in this shared region, LLMs effectively learn interlingual semantic representations that facilitate multilingual performance, e.g. through emphasizing semantics while minimizing language signals, retaining them only for language-specific predictions. This is where key interlingual alignments form, enabling LLMs to leverage universal semantic patterns for multilingual tasks.

\subsection{Fragmented Region}
While some languages enjoy substantial overlaps in \(\mathcal{M}_{c}\), the less-aligned others occupy fragmented region \(\mathcal{M}_{f}\) as they reflect model’s representational limitation to embed the representation from these languages into \(\mathcal{M}_{c}\). Factors such as sparse training data, typological distance, and morphological complexity might lead to partial alignment of these representations. Consequently, representations in \(\mathcal{M}_{f}\) tend to be more weakly aligned to the universal semantics encoded by \(\mathcal{M}_{c}\). This misalignment can degrade multilingual performances: tasks that rely on inputs from the less-aligned languages may exhibit lower performance since they draw from 
semantics that loosely intersects with \(\mathcal{M}_{c}\).

\section{Semantic Alignment of Multilingual LLMs Representations}
\label{sec:tsne_anc}

We explore the presence and characteristics of the components \(\mathcal{M}_c\) and \(\mathcal{M}_{f}\) within multilingual LLMs through assessing the semantic alignment between its hidden-states, derived from parallel inputs on various languages. Initially, we project LLMs' internal hidden-state embeddings into a 2D space to broadly assess proximities of parallel language representations and observe whether parallel input pairs in different languages clusters or overlaps. We then measure the cross-lingual alignment across the parallel hidden-state embeddings through neuron activation consistency w.r.t their resource-level, linguistic features, and geographical region.

We sample 31 diverse language subsets of Flores-200~\cite{nllb2024scaling} varied on its resource-level, region, and family~\cite{ethnologue} (see Tables~\ref{tab:anc_lang_stats} and~\ref{tab:complete_langs}) as proxies to typological and morphological features~\cite{georgi2010comparing}. Over experiments, we assess several multilingual LLMs: Aya Expanse (8B)~\cite{dang2024aya}, Llama-3.1 (8B)~\cite{dubey2024llama}, Gemma-2 (9B)~\cite{team2024gemma}, Qwen-2.5 (7B)~\cite{yang2024qwen2}. We observe a universal phenomenon from these models, as described in the following sections. We put the further comparison details in Appendix~\ref{app:interlingual_alignment_various_model}.

\begin{figure*}[!t]
  \centering
  \begin{subfigure}[t]{0.49\linewidth}
      \includegraphics[trim={7, 7, -5, 6}, clip, width=\linewidth]{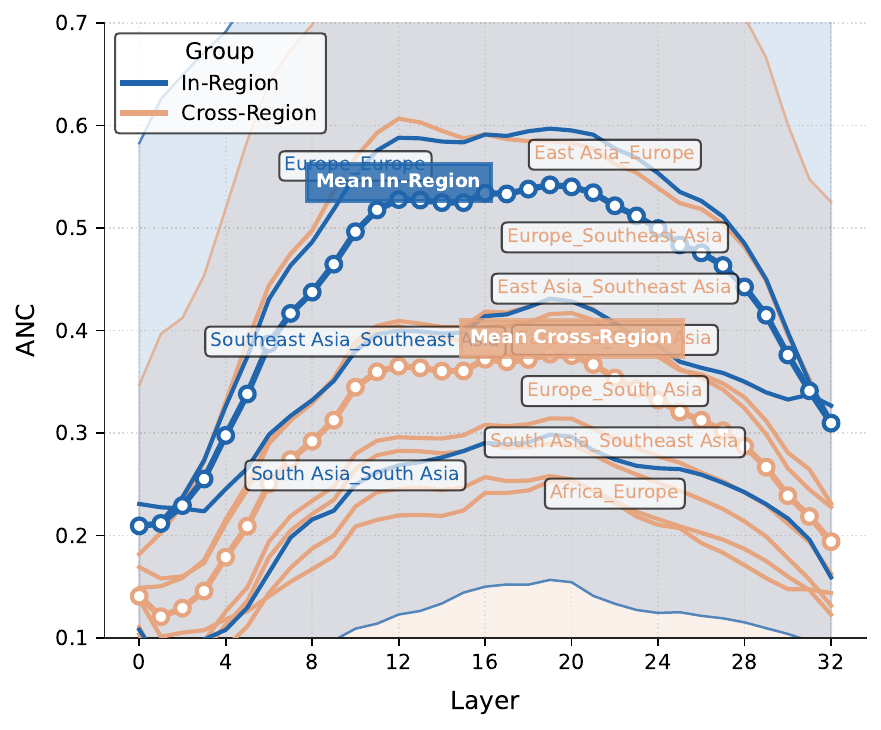}
      \vspace{-15pt}
      \caption{Highlights on pairs w.r.t their linguistic region}
      \label{fig:anc_region}
  \end{subfigure}
  \begin{subfigure}[t]{0.49\linewidth}
    \centering
      \includegraphics[trim={7, 7, -5, 6}, clip, width=\linewidth]{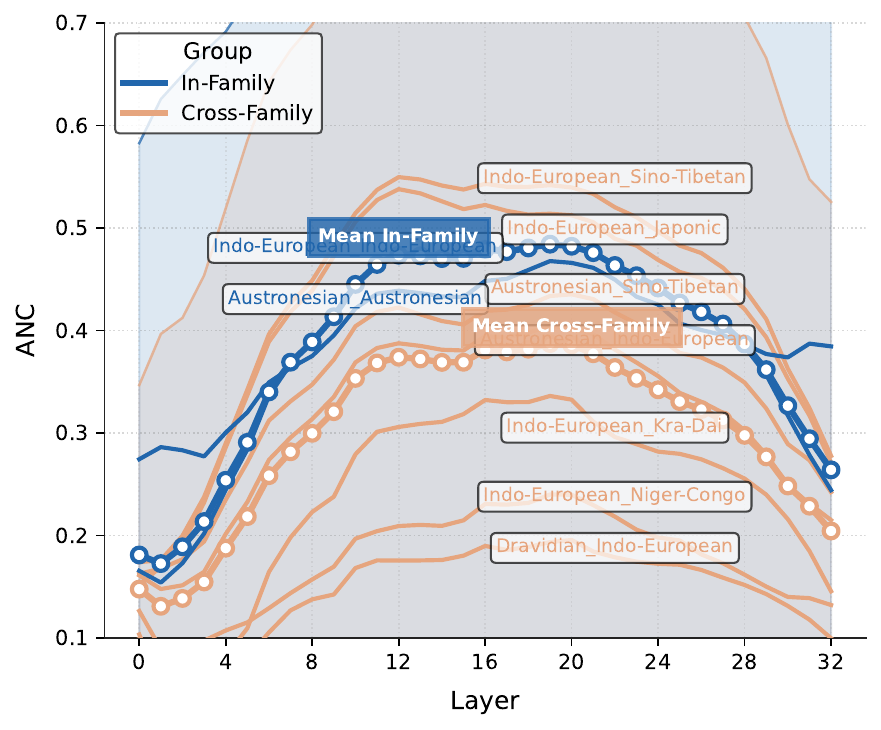}
      \vspace{-15pt}
      \caption{Highlights on pairs w.r.t their linguistic family}
      \label{fig:anc_family}
  \end{subfigure}
  \vspace{-3pt}
  \caption{Comparisons of per-layer ANC scores on Aya Expanse (8B) with highlights on pairs w.r.t their linguistic region and family. Consistently stronger alignments are observed between within-group mean correlations.}
  \label{fig:anc_region_family}
  \vspace{-7pt}
\end{figure*}

\subsection{Inherent Regional Clustering with Mid-Layers High-Resource Alignment}

We employ t-SNE~\cite{van2008visualizing} to project LLMs' hidden-state embeddings into a 2D space and assess the proximities across language clusters. As t-SNE retains local neighborhood structures, overlaps in this 2D space imply closeness in the original high-dimensional space. In scenarios where representations are interlingually aligned, their nearest neighbors should comprise of multiple languages. 
We visualize the cross-lingual comparisons on the early, middle, and late layers of Aya Expanse (8B) in Figures~\ref{fig:tsne_1} and~\ref{fig:tsne_2}, and others in Appendix~\ref{app:tsne}. We ran t-SNE with perplexity values of {5, 15, 30, 50}, and observe consistent trends. Results for perplexity 15 are shown here; others are in Appendix~\ref{sec:appendix_tsne_ppl}.

The t-SNE visualizations reveal distinct structural patterns across early (layer 0), intermediate (layer 16), and late (layer 32) layers (see Figures~\ref{fig:tsne_earlylayer},~\ref{fig:tsne_1},~\ref{fig:tsne_latelayer}, respectively). In early and late layers, parallel language representations cluster according to resource levels and linguistic features, with minimal overlap. In contrast, the intermediate layer shows interlingual overlaps that transcend familial and regional boundaries, such as English and Russian overlapping with Indonesian, and Chinese with French. While overlaps mainly involve high-resource languages (HRLs), low-resource languages (LRLs) also exhibit overlaps, often due to regional factors. Meanwhile, some parallel representations remain fragmented outside these overlaps. These intermediate layer observations show that the quality of alignment varies.
We further investigate the interactions in high-dimensional space to understand the alignment properties, in order to complement these low‐dimensional observations.

\subsection{Cross-lingual Alignments Depend on Resource-level and Linguistic Properties}

\paragraph{Measurement.} We further quantify the alignment characteristics by measuring neuron activation alignment for semantically identical inputs across different \(\ell\) through \textit{Average Neuron-wise Correlation} (ANC)~\cite{del2022cross}. The ANC score in a certain LLM layer is defined as:
\[
\operatorname{ANC}(\ell, \ell') = \frac{1}{d} \sum_{i\in d} \ corr\!\left(f_\ell^i(\mathbf{x}), f_{\ell'}^i(\mathbf{x}')\right),
\]
with 
\(f_\ell^i(\mathbf{x})\) as the activation of \(i\)-th neuron for language \(\ell\) and \(corr\) denotes Pearson correlation between corresponding activations in \(\ell \text{ and } \ell'\). 
We visualize layer-wise ANCs from Aya Expanse in Figure~\ref{fig:anc_hrllrl} and~\ref{fig:anc_region_family}, and others in Appendix~\ref{app:anc_figures}.

\paragraph{Findings.} We find the ``first align, then predict'' patterns varies across language pairs. Notably, pairs of HRLs demonstrate strong correlations, while pairs involving LRLs exhibit lower scores (see Figure~\ref{fig:anc_hrllrl}). Similarly, a consistent gap persists between within- and cross-group mean correlations, indicating stronger alignment within familial and regional language groups. Detailed analysis in Table~\ref{tab:anc_top_languages} illustrates that most correlated pairs among LLMs are similar on their HRLs. Despite differing rankings, instruction-tuned LLMs exhibit similar sets of top language pairs with its pre-trained counterparts. These significant alignment gaps in cross-lingual correlations indicates latent discrepancies between semantically identical representations that stem from sparse data, typological distance, and the morphological complexity of languages.

\section{Intrinsic Interlinguality of LLMs}


In Section~\ref{sec:tsne_anc}, we empirically demonstrated that multilingual LLMs' behavior aligns closely with the theoretical framework introduced in Section~\ref{sec:theory}. Building upon these theoretical insights and empirical validations, we propose the Interlingual Local Overlap (ILO) score to measure the consistency of interlingual alignment in multilingual LLMs. Specifically, ILO score considers the local neighborhoods of models' hidden-state embeddings of linguistically-diverse semantically-parallel inputs, to indicate and quantify their intrinsic interlingual alignment in the high-dimensional space.

\subsection{Interlingual Local Overlap Score}

Given \(N\) input samples from set of languages in \(\mathcal{L}\), \(\{\mathbf{x}_i^{\ell}\}_{\ell \in \mathcal{L}, i\in N}\), each sample \(\mathbf{x}_i^{\ell}\) is embedded in model space \(\mathcal{H}\) via \(f_{\ell}(\mathbf{x})\). Let's denote \(\mathcal{N}(\mathbf{x}_i^{\ell})\) as the set of  \(k\)-nearest neighboring languages of \(\mathbf{x}_i^{\ell}\), defined as \(\mathcal{N}(\mathbf{x}_i^{\ell}) = \{\ell' \neq \ell : \mathbf{x}_j^{\ell'} \in \operatorname{NN}_k(\mathbf{x}_i^{\ell})\}\).

\paragraph{Bridge.} 

The bridge score \(B_{\ell}\) determines the degree of local interlingual mixing, analogous to the participation coefficient in graph theory, which assesses a node's link distribution across modules~\cite{guimera2005cartography, mijalkov2017braph}. Bridge score measures the proportion of samples whose \(k\)-nearest neighbors in \(\mathcal{H}\) include at least \(\tau\) unique other languages, formally:
\[
B_{\ell} = \frac{1}{N} \sum_{i\in N} \mathbf{1} \big(|\mathcal{N}(\mathbf{x}_i^{\ell})| \geq \tau \big)
\]
A score of \(\approx\) 1 indicates that samples from \(\ell\) consistently neighboring with diverse other languages.

\paragraph{Reachability.} 

Inspired by classical degree of centrality in network analysis~\cite{freeman2002centrality, borgatti2006graph}, which quantifies a node's connections, we define reachability score to measure cross-lingual connectivity of \(\ell\) representations. We view the multilingual space \(\mathcal{H}\) as an undirected graph with each hidden-state embeddings as nodes linked to their \(k\)-nearest neighbors. The reachability score \(R_\ell\) quantifies the connectivity degree of \(\ell\) representations 
, defined as:
\[
R_\ell = \frac{1}{|\mathcal{L}| - 1} \bigg|\bigcup_{i \in N} \mathcal{N}(\mathbf{x}_i^{\ell}) \bigg|
\]
\(R_\ell\) enumerates the fraction of unique languages encountered across all samples of \(\ell\) in \(\mathcal{L}\), excluding itself. A high \( R_\ell \) suggests that \(\ell\) representations connect extensively within the multilingual space.

\begin{table}[!t]
\small
\centering
\resizebox{\linewidth}{!}{%
\begin{tabular}{llrr}
\toprule
\textbf{Dataset} & \textbf{Usage} & \multicolumn{1}{l}{\textbf{\# Lang}} & \multicolumn{1}{l}{\textbf{\# Sample}} \\ \midrule
Flores-200 & Analysis & 31 & 30,907 \\
GSM8KInstruct & Training & 10 & 73,559 \\
MGSM & Evaluation & 11 & 2,750 \\ \bottomrule
\end{tabular}}
\vspace{-3pt}
\caption{Dataset statistics. “\# Lang” indicates the number of languages represented in the dataset, and “\# Sample” signifies the total sample size included.}
\label{tab:dataset_list}
\vspace{-10pt}
\end{table}

\paragraph{Interlingual Local Overlap (\abbrvmetric{}).}
We then define an interlingual local overlap score \(\operatorname{ILO}_{\ell}\) to quantify the holistic interlingual alignment of language \(\ell\) within \(\mathcal{H}\), formally:
\[
 \operatorname{ILO}_{\ell} = 2\cdot\frac{B_\ell\cdot R_\ell}{B_\ell +R_\ell}
\]
with the harmonic mean emphasizes the requirement of strong assessments in both the mixing and connectivity for the representations of \(\ell\) to be considered as locally overlapping with other languages. Consequently, aggregated \(\bar{\operatorname{ILO}}_{\mathcal{L}}\) of high \(\operatorname{ILO}_{\ell}\) in
\[
\bar{\operatorname{ILO}}_{\mathcal{L}} = \frac{1}{|\mathcal{L}|} \sum_{\mathcal{L}} \operatorname{ILO}_{\ell},
\]
signals that multilingual LLMs effectively encode all of the diverse language inputs as aligned interlingual semantics within those in \(\mathcal{L}\).

\paragraph{Preserving Interlinguality of LLMs.}

We demonstrate how \abbrvmetric{} illuminate the performance variations in cross-lingual transfer and concurrently underscore the critical role of semantic interlingual alignment in multilingual LLMs. Cross-lingual transfer capitalizes on shared features to enhance multilingual capabilities~\cite{philippy-etal-2023-towards}, typically involving single-language fine-tuning on a source language and directly applying it to target languages without further tuning. Despite its success, LLMs can suffer from catastrophic forgetting~\cite{mccloskey1989catastrophic, french1999catastrophic, biesialska2020continual}, where their cross-lingual generalization may degrade~\cite{liu2021preserving, winata2023overcoming}. Research suggests LLMs align multilingual inputs into language-independent representations, then revert them back to the query's original language~\cite{muller2021first, zhao2024how}. 
Building on these insights, we conduct an experiment to preserve interlingual alignments by employing a \textbf{selective freezing} strategy, where we partially freeze parameters critical to language alignment.
Our aim is to assess the potential mitigation of cross-lingual disruption, evaluated through \abbrvmetric{} scores.

\begin{table*}[!ht]
\centering
\resizebox{\linewidth}{!}{%
\begin{tabular}{l|c|rrrrrrrrrrr|rr}
\toprule
& & \multicolumn{11}{c}{\textbf{Accuracy}} & \multicolumn{2}{|c}{\textbf{Average}} \\
\cmidrule(lr){3-13}\cmidrule(lr){14-15}
\multirow{-2.3}{*}{\textbf{Method}} & \multirow{-2.2}{*}{\shortstack{\textbf{Training} \\ \textbf{languages}}} & \multicolumn{1}{c}{\textbf{ben}} & \multicolumn{1}{c}{\textbf{tha*}} & \multicolumn{1}{c}{\textbf{swh}} & \multicolumn{1}{c}{\textbf{tel*}} & \multicolumn{1}{c}{\textbf{jpn}} & \multicolumn{1}{c}{\textbf{zho}} & \multicolumn{1}{c}{\textbf{deu}} & \multicolumn{1}{c}{\textbf{fra}} & \multicolumn{1}{c}{\textbf{rus}} & \multicolumn{1}{c}{\textbf{spa}} & \multicolumn{1}{c}{\textbf{eng}} & \multicolumn{1}{|c}{\textbf{All}} & \multicolumn{1}{c}{\textbf{XL}} \\ \midrule
Pre-trained & mixed & 11.6\% & 12.0\% & 7.2\% & 0.0\% & 10.4\% & 8.8\% & 16.0\% & 12.4\% & 14.0\% & 11.6\% & 17.6\% & 10.3\% & \multicolumn{1}{c}{-} \\ \midrule
\multirow{9.7}{*}{Fine-tuning} & ben & \cellcolor[HTML]{00FFFF}\textbf{23.2\%} & \cellcolor[HTML]{EA9999}4.8\% & \cellcolor[HTML]{EA9999}1.2\% & 3.2\% & \cellcolor[HTML]{EA9999}10.0\% & 9.6\% & \cellcolor[HTML]{EA9999}10.8\% & 13.6\% & \cellcolor[HTML]{EA9999}11.6\% & 14.8\% & \cellcolor[HTML]{EA9999}12.8\% & 10.5\% & 9.2\% \\
 & tha* & \cellcolor[HTML]{EA9999}1.6\% & \cellcolor[HTML]{00FFFF}\textbf{32.8\%} & \cellcolor[HTML]{EA9999}4.4\% & 1.6\% & 14.4\% & 14.8\% & 17.2\% & 19.2\% & 18.0\% & 20.4\% & 25.6\% & 15.5\% & 13.7\% \\
 & swh & \cellcolor[HTML]{EA9999}3.2\% & \cellcolor[HTML]{EA9999}6.4\% & \cellcolor[HTML]{00FFFF}{\ul\textbf{30.8\%}} & 2.8\% & 11.2\% & 12.4\% & 20.4\% & 19.6\% & 14.8\% & 22.4\% & 26.8\% & 15.5\% & 14.0\% \\
 & jpn & \cellcolor[HTML]{EA9999}3.6\% & \cellcolor[HTML]{EA9999}7.2\% & \cellcolor[HTML]{EA9999}2.8\% & 1.2\% & \cellcolor[HTML]{00FFFF}{\ul \textbf{32.8\%}} & 21.6\% & 19.6\% & 18.0\% & 18.4\% & 22.4\% & 28.8\% & 16.0\% & 14.4\% \\
 & zho & \cellcolor[HTML]{EA9999}0.8\% & \cellcolor[HTML]{EA9999}7.2\% & \cellcolor[HTML]{EA9999}2.4\% & 1.6\% & 22.0\% & \cellcolor[HTML]{00FFFF}{\ul \textbf{34.8\%}} & 19.6\% & 19.6\% & 21.6\% & 21.2\% & 27.6\% & 16.2\% & 14.4\% \\
 & deu & \cellcolor[HTML]{EA9999}8.0\% & 16.4\% & 8.0\% & \textbf{4.0\%} & 19.2\% & 19.6\% & \cellcolor[HTML]{00FFFF}{\ul \textbf{37.6\%}} & \textbf{34.4\%} & 23.6\% & 28.8\% & 36.4\% & 21.5\% & 19.8\% \\
 & fra & \cellcolor[HTML]{EA9999}4.8\% & \cellcolor[HTML]{EA9999}11.6\% & \cellcolor[HTML]{EA9999}4.0\% & 3.2\% & 16.0\% & 16.8\% & 31.6\% & \cellcolor[HTML]{00FFFF}\textbf{34.4\%} & 25.6\% & 34.4\% & 35.6\% & 19.8\% & 18.4\% \\
 & rus & \cellcolor[HTML]{EA9999}4.0\% & 14.0\% & \cellcolor[HTML]{EA9999}4.0\% & 1.2\% & 17.2\% & 16.4\% & 29.6\% & 28.4\% & \cellcolor[HTML]{00FFFF}\textbf{34.0\%} & 30.0\% & 26.4\% & 18.7\% & 17.1\% \\
 & spa & \cellcolor[HTML]{EA9999}4.8\% & 16.0\% & \cellcolor[HTML]{EA9999}2.8\% & 2.4\% & 14.4\% & 19.6\% & 28.4\% & 30.8\% & 31.2\% & \cellcolor[HTML]{00FFFF}{\ul \textbf{38.4\%}} & 38.4\% & 20.7\% & 18.9\% \\
 & eng & \cellcolor[HTML]{EA9999}6.4\% & 14.4\% & \cellcolor[HTML]{EA9999}6.0\% & 2.4\% & 18.8\% & 24.4\% & 37.2\% & 27.2\% & 33.6\% & 33.2\% & \cellcolor[HTML]{00FFFF}\textbf{43.2\%} & \textbf{22.4\%} & \textbf{20.4\%} \\ \midrule
\multirow{9.7}{*}{\shortstack[l]{Selective \\ Freezing}}  & ben & \cellcolor[HTML]{00FFFF}{\ul \textbf{26.4\%}} & 12.8\% & 11.6\% & 14.4\% & 13.6\% & 14.8\% & 19.6\% & 20.0\% & 20.0\% & 17.6\% & \cellcolor[HTML]{EA9999}17.2\% & 17.1\% & 16.2\% \\
 & tha* & 14.8\% & \cellcolor[HTML]{00FFFF}{\ul \textbf{34.0\%}} & 12.0\% & 12.4\% & 15.6\% & 21.6\% & 25.2\% & 22.0\% & 20.4\% & 24.4\% & 32.4\% & 21.3\% & 20.1\% \\
 & swh & \cellcolor[HTML]{EA9999}9.2\% & 16.4\% & \cellcolor[HTML]{00FFFF}\textbf{22.8\%} & 5.6\% & 14.0\% & 12.4\% & 18.4\% & 23.6\% & 19.2\% & 20.4\% & 27.6\% & 17.2\% & 16.7\% \\
 & jpn & 16.0\% & 17.6\% & 12.0\% & 11.2\% & \cellcolor[HTML]{00FFFF}\textbf{27.2\%} & 28.8\% & 24.4\% & 23.2\% & 24.0\% & 24.4\% & 29.6\% & 21.7\% & 21.1\% \\
 & zho & 17.2\% & 17.2\% & 12.4\% & 12.0\% & 22.4\% & \cellcolor[HTML]{00FFFF}{\ul\textbf{34.8\%}} & 29.6\% & 22.4\% & 27.6\% & 23.6\% & 37.2\% & 23.3\% & 22.2\% \\
 & deu & 12.8\% & 22.8\% & 14.4\% & 17.6\% & 20.0\% & 25.6\% & \cellcolor[HTML]{00FFFF}\textbf{36.0\%} & 29.6\% & 27.6\% & 32.8\% & 39.2\% & 25.3\% & 24.2\% \\
 & fra & 14.8\% & 24.8\% & 18.4\% & 12.0\% & 21.2\% & 21.2\% & 33.6\% & \cellcolor[HTML]{00FFFF}{\ul\textbf{37.2\%}} & 32.0\% & \textbf{36.8\%} & 36.8\% & 26.3\% & 25.2\% \\
 & rus & 20.4\% & 19.6\% & 11.6\% & {\ul \textbf{18.8\%}} & 22.0\% & 19.6\% & 28.8\% & 25.2\% & \cellcolor[HTML]{00FFFF}38.4\% & 28.8\% & 32.0\% & 24.1\% & 22.7\% \\
 & spa & 20.0\% & 24.0\% & 17.6\% & 16.8\% & 18.0\% & 27.2\% & 33.6\% & 33.6\% & 29.6\% & \cellcolor[HTML]{00FFFF}34.0\% & 36.4\% & 26.4\% & 25.7\% \\
 & eng & 20.4\% & 24.0\% & 18.0\% & 16.4\% & 20.4\% & 26.4\% & 35.2\% & 30.0\% & {\ul \textbf{43.6\%}} & 32.4\% & \cellcolor[HTML]{00FFFF}{\ul \textbf{46.8\%}} & {\ul\textbf{28.5\%}} & {\ul\textbf{26.7\%}} \\ \bottomrule
\end{tabular}}
\caption{Cross-lingual transfer performance on MGSM for Llama-3.1 (8B) without and with selective freezing. ``XL'' denotes average on languages that were not fine-tuned. Diagonal entries in \sethlcolor{highlightBlue}\hl{blue highlights} correspond to source language performances. \sethlcolor{highlightRed}\hl{Red highlights} indicate decrease from pre-trained baseline. \textbf{Bold} and {\ul underline} respectively denote the best within group and within column. The (*) marks languages classified as low-resource in Flores-200.}
\label{tab:xlt_frz_llama}
\vspace{-6pt}
\end{table*}

\subsection{Experiment Design}

To preserve the aligned semantics within multilingual model space, we experiment on freezing the parameters of the early layers on the first 4, 8, 12, and 16 layers. Additionally, we keep the token embedding, final layer normalization, and language modeling head (output projection layer) fixed. We identify these parameters as the language aligners.

\paragraph{Datasets.} We attend specifically to multilingual mathematical reasoning task, as it is inherently language-independent. We utilize the multilingual dataset \texttt{GSM8KInstruct}~\cite{chen-etal-2024-breaking}, which extends the English mathematical reasoning dataset \texttt{GSM8K}~\cite{cobbe2021training} by translating English instructions and chain-of-thought responses into 9 non-English languages via automatic translation and native-speaker human verification. To evaluate the model performance in this task, we utilize the \texttt{MGSM} benchmark~\cite{shilanguage}. We attach the complete dataset statistics in Table~\ref{tab:dataset_list}.

\paragraph{Evaluation.} We evaluate the accuracy of LLM greedy decoding zero-shot responses. Specifically, we employ the evaluation of~\citeauthor{zhu2024question} and determine answer accuracy by verifying that the final numerical value produced in the LLM’s output exactly matches the ground-truth. In addition, we utilize \abbrvmetric{} to investigate how changes in training impact LLMs’ interlingual semantic alignment. To compute the \abbrvmetric{} scores, we define a neighborhood size large enough to be informative and small enough to respect the local structures, while requiring each neighborhood to be rich in interlingual mixing. We experimented with Euclidean and cosine distance metric, with $k$, $\tau$ values of {(5,3), (10,5), (20,10)} and observe consistent trends. Results using $k=10, \tau=5$ and Euclidean distance are shown here; others in App.~\ref{sec:appendix_knn}. We evaluate the ILO scores using the same dataset from Section~\ref{sec:tsne_anc}.

\paragraph{Models.} 
We employ two multilingual LLMs: Llama-3.1 (8B) and Gemma-2 (9B).
We train both LLMs using the same hyperparameters with learning rate $8e-5$, batch size $8$, and gradient accumulation of $16$ for $3$ epochs using 4 A800 GPUs.

\begin{figure*}[!t]
    \centering
    \includegraphics[trim={0, 0, 0, 14}, clip, width=\linewidth]{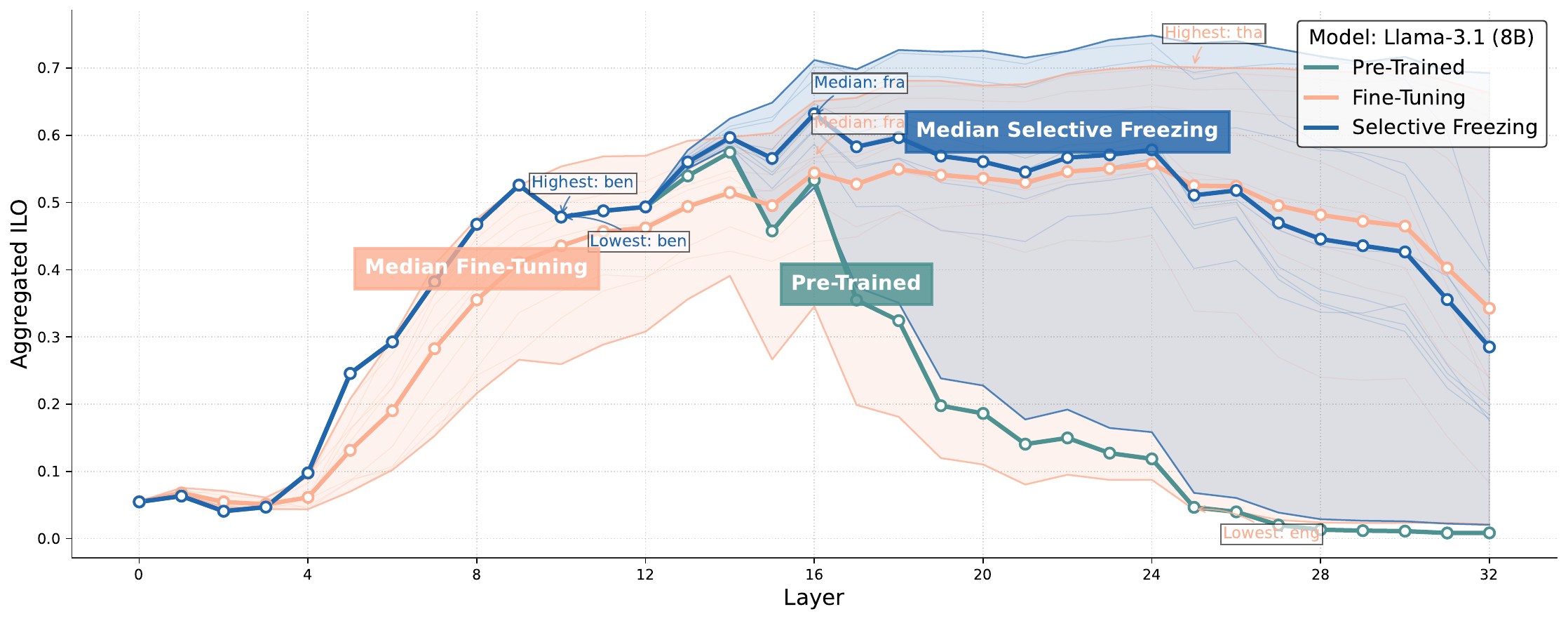}
    \vspace{-16pt}
    \caption{
    Layer-wise \(\bar{\operatorname{\abbrvmetric{}}}_{\mathcal{L}}\) scores for all of the source-languages in the single-language training on Llama-3.1 (8B) in \textbf{pre-trained}, \textbf{fine-tuning}, and \textbf{selective freezing} modes. Decrease in alignment from single-language \textbf{fine-tuning} is seen in the early layers, whereas \textbf{selective freezing} allows LLM to sustain its \textbf{pre-trained} semantic alignment.
    }
    \label{fig:interlingual_score_llama}
\end{figure*}

\subsection{Results and Analysis}

\paragraph{Cross-Lingual Transfer.}  We present findings from our cross-lingual transfer experiments, detailed in the Tables~\ref{tab:xlt_frz_llama} and~\ref{tab:xlt_frz_gemma} within the ``\textbf{fine-tuning}'' rows, where we evaluated the performance of the fine-tuned Llama-3.1 and Gemma-2 respectively. Consistent with the expectations, we observed substantial cross-lingual transfer signified by improved performance in both source and target languages, even without direct training in those languages. The transfer is notably more pronounced in HRLs and languages within the same families and regions, such as the Indo-European languages in Europe: English, Spanish, Russian, French, and German. Remarkably, in some instances, performances on the target languages paralleled the accuracies in the source language, as exemplified by Spanish-to-English achieving \(38.4\%\), which is on par with the Spanish-to-Spanish performance.

Despite the transfer, performance degradations are also observed on some of target languages. We conjecture that this issue stems from disruptions in the functionality of the aligner module. To investigate this hypothesis, we compute per-layer aggregated \(\bar{\operatorname{\abbrvmetric{}}}_{\mathcal{L}}\) scores, and visualize them in Figures~\ref{fig:interlingual_score_llama} and~\ref{fig:interlingual_score_gemma}, for all of the source-languages trained on each the Llama-3.1 (8B) and Gemma-2 (9B) models. Both figures show a notable decrease in interlingual semantic alignment post fine-tuning that appears as early as in the $4^{th}$ layer for Llama and the $6^{th}$ layer for Gemma. Critically, the degree of alignment does not recover to the height of its pre-trained levels even after additional computational stages in subsequent layers. Furthermore, the interlingual overlaps initially present in the pretrained models become disrupted following single-language fine-tuning, as evidenced by reduced overlapping centers and loosened language clusters (Figs (b) of \ref{fig:tsne_llama31_ft} vs \ref{fig:tsne_llama31}, and \ref{fig:tsne_gemma2_ft} vs \ref{fig:tsne_gemma2}).



\paragraph{Preservation of LLMs' Interlinguality.}
Here we analyze the impact on freezing the first 12 layers, since it provides the best aggregated improvements (see Appendix~\ref{sec:appendix_layer} for details).
The quantitative analysis through the lens of the aggregated \(\bar{\operatorname{\abbrvmetric{}}}_{\mathcal{L}}\)  reveals that multilingual LLMs trained with \textbf{selective-freezing} mechanism sustain their prior semantic alignment levels in the early layers, and across all layers, as demonstrated in Figures~\ref{fig:interlingual_score_llama} and \ref{fig:interlingual_score_gemma}. Empirical findings in Tables~\ref{tab:xlt_frz_llama} and~\ref{tab:xlt_frz_gemma} further corroborate these insights, highlighting the substantial impact of maintaining interlingual semantic alignments on enhancing multilingual performances. Through keeping the aligner parameters unchanged, both LLMs understudy gain improved cross-lingual generalization compared to their post fine-tuning performances on source languages. Enhanced transfers can be observed on languages within-families and within-regions, with improvements and nearly no degradation towards the low-resource, cross-family, and cross-regional languages. Additionally, models fine-tuned with selective freezing effectively retain their original interlingual alignment, with overlapping centers largely preserved and clusters remaining tight (see Figs (b) of \ref{fig:tsne_llama31_freeze} vs \ref{fig:tsne_llama31}, \ref{fig:tsne_gemma2_freeze} vs \ref{fig:tsne_gemma2}, and App.~\ref{app:ilo_in_tsne}). These findings indicate that preserving the interlingual alignment in LLMs is essential for scalable multilingual learning. They emphasize the critical role of interlingual representation alignments in enhancing the multilingual capabilities of LLMs.

\section{Conclusion}

The emergence of multilingual LLMs demonstrates that interlingual constructs naturally arise, even in the absence of explicit objectives. We introduce a conceptual framework to understand interlingual representations, identifying both the core interlingual region that captures shared semantics, and fragmented components that reveal representational limitations in aligning with this core region. To advance the understanding of interlingual semantic alignment, we propose the Interlingual Local Overlap (\abbrvmetric{}) score which quantifies alignment in the local neighborhood structures of interlingual high-dimensional representations. Our proposed framework and metric illuminates the critical role of semantic alignment, offering a quantitative view into the high-dimensional alignment of multilingual representations. This study emphasizes interlingual semantic alignment and provides critical insights to optimize multilingual LLMs in the context of diverse linguistic tasks.

\section*{Limitations}

\paragraph{Bias on linguistic family.} In our analysis of interlingual regions, we sample 31 diverse languages from the Flores-200 set, representing various resource levels, geographical regions, and language families. We note, however, that there is a predominance of Indo-European languages within our HRLs subset. This distribution reflects the broader availability of linguistic data, as evidenced by web crawl statistics from CommonCrawl, where Indo-European languages are disproportionately represented. This imbalance is not intentional but rather an inherent limitation arising from existing data availability. Consequently, the observed stronger correlations among HRLs may partially reflect this underlying bias. We encourage future works to account for this, since observed correlations among HRLs may partially reflect this underlying bias.

\paragraph{Broader multilingual evaluations.}  Additionally, our study of cross-lingual transfer primarily utilizes multilingual mathematical reasoning task due to their largely language-agnostic nature. Such task allow us to simultaneously asses the linguistic understanding and logical reasoning capabilities of multilingual LLMs. We argue that the cross-lingual transfer capabilities evaluated within this work offer significant insights into general multilingual performance. Nonetheless, we encourage future studies to broaden evaluations to other tasks to extend the insights into interlingual alignment.


\paragraph{Expanding the core interlingual region.}  
Our works presumes the existence of the core interlingual region where semantically aligned representations shared across languages, and others that only partially aligned to this core. Future works could explore on expanding this core interlingual region to encompass a broader range of languages, i.e. to introduce learning techniques that explicitly encourage deeper and more diverse interlingual mixing. Incorporating a larger, more heterogeneous multilingual datasets and leveraging linguistic priors might further strengthen the core region, and in turn, enhancing the universality of the core interlingual representations.

\paragraph{Bridging fragmented regions.}  
A significant limitation of existing multilingual LLMs is that certain languages, particularly the underrepresented or typologically distant ones, most likely form fragmented region rather than being integrated fully with the core cluster. To address this, future work could aim to develop targeted strategies to encourage the integration of these regions and to narrow these gaps, i.e. under conditions of extremely limited data. Such interventions could facilitate the alignments of interlingual representation, thereby improving overall inclusivity and richness in linguistic diversity of the multilingual LLMs.

\paragraph{Predicting cross-lingual transfer.}  
Although our work provides valuable insights into the local alignment of multilingual embeddings, it does not predict downstream cross-lingual transfer performance. One key limitation, for example, is that our proposals captures generic interlingual mixing of hidden-states representations and not the alignments of task vectors~\cite{ilharcoediting} that might be integral for effective transfer. This disconnect may arise when models achieve strong interlingual alignment while simultaneously losing critical nuances required for task performance. Future work could explore the integration of our proposals with task-aware signals, to develop quantifiers that are more designed to predict cross-lingual transfer.

\paragraph{Towards pure semantic representations.}  
While our current work focuses solely on textual embeddings, a major frontier for future research lies in extending the framework of quantifying alignment via the local neighborhood structures of high-dimensional representations, to multimodal settings. Considering information from another modalities, it may be beneficial to disentangle and measure pure semantic content from modality-specific biases effectively. Exploring this direction not only hints promises to elucidate and improve modality-transfer but also potentially advance our understanding of how different forms of information interact to shape a universal semantic space. We envision our work, upon many others (e.g.~\citet{cahyawijaya2024high, engels2025not, ji2024llm, liu2024universal, grosse2023studying}), to foster explorations towards the study of LLMs' semantic space.

\section*{Ethical Considerations}


The exploration of interlingual representation in multilingual LLMs presents a unique opportunity to foster diversity and inclusivity in the field of NLP. Our work introduces framework and metrics to inspect interlingual representations in multilingual LLMs. They enable the analysis of interlingual alignment of different languages in the naturally emerging interlingual constructs within LLMs. We use publicly available parallel corpora and adhere to best practices in data handling, ensuring that no sensitive or personally identifiable information is involved. While our proposals help reveal disparities in representation, through this work, we instead leverage these insights to drive proactive interventions—ensuring future multilingual LLMs are not only more inclusive but also more reflective of the rich linguistic diversity they aim to serve. We hope our results contributes to more equitable model development and encourages further investigation into mitigating potential representational gaps across underrepresented languages.

\paragraph{Embracing Language Diversity} Our work aims to create a universal representation that respects and preserves the unique characteristics of each language. Our findings highlight the importance of consistent interlingual alignments. By recognizing and capturing shared semantic structures through interlingua representations, LLMs can contribute to the preservation of linguistic diversity, ensuring that no single language or language group dominates the representation space. We envision LLMs to effectively represent and understand diverse languages, to be truly inclusive in language technology (e.g.~\citet{cahyawijaya2024llm}). This is particularly crucial for underrepresented languages and communities, enabling them to have their voices heard and enabling them equal access of information, for example to their language-agnostic applications.

\paragraph{Addressing Bias and Fairness} The study's observation of varying alignment consistencies across language groups underscores the need for careful consideration of bias. By identifying and addressing fragmented components due to representational limitations, we can work towards creating fairer representations. This is essential to prevent the reinforcement of existing biases and ensure equitable treatment of all languages. When LLMs effectively bridge the gap between languages, they enable seamless communication and understanding, benefiting diverse communities and fostering a more inclusive digital information systems.


\bibliography{custom}

\begin{thebibliography}{71}
\providecommand{\natexlab}[1]{#1}

\bibitem[{Al-Shedivat and Parikh(2019)}]{al2019consistency}
Maruan Al-Shedivat and Ankur Parikh. 2019.
\newblock Consistency by agreement in zero-shot neural machine translation.
\newblock In \emph{Proceedings of the 2019 Conference of the North American Chapter of the Association for Computational Linguistics: Human Language Technologies, Volume 1 (Long and Short Papers)}, pages 1184--1197.

\bibitem[{Artetxe and Schwenk(2019)}]{artetxe-schwenk-2019-massively}
Mikel Artetxe and Holger Schwenk. 2019.
\newblock \href {https://doi.org/10.1162/tacl_a_00288} {Massively multilingual sentence embeddings for zero-shot cross-lingual transfer and beyond}.
\newblock \emph{Transactions of the Association for Computational Linguistics}, 7:597--610.

\bibitem[{Biesialska et~al.(2020)Biesialska, Biesialska, and Costa-juss{\`a}}]{biesialska2020continual}
Magdalena Biesialska, Katarzyna Biesialska, and Marta~R Costa-juss{\`a}. 2020.
\newblock Continual lifelong learning in natural language processing: A survey.
\newblock In \emph{Proceedings of the 28th International Conference on Computational Linguistics}, pages 6523--6541.

\bibitem[{Borgatti and Everett(2006)}]{borgatti2006graph}
Stephen~P Borgatti and Martin~G Everett. 2006.
\newblock A graph-theoretic perspective on centrality.
\newblock \emph{Social networks}, 28(4):466--484.

\bibitem[{Brown et~al.(1990)Brown, Cocke, Della~Pietra, Della~Pietra, Jelinek, Lafferty, Mercer, and Roossin}]{brown-etal-1990-statistical}
Peter~F. Brown, John Cocke, Stephen~A. Della~Pietra, Vincent~J. Della~Pietra, Fredrick Jelinek, John~D. Lafferty, Robert~L. Mercer, and Paul~S. Roossin. 1990.
\newblock \href {https://aclanthology.org/J90-2002} {A statistical approach to machine translation}.
\newblock \emph{Computational Linguistics}, 16(2):79--85.

\bibitem[{Brown et~al.(1991)Brown, Lai, and Mercer}]{brown1991aligning}
Peter~F Brown, Jennifer~C Lai, and Robert~L Mercer. 1991.
\newblock Aligning sentences in parallel corpora.
\newblock In \emph{29th Annual Meeting of the Association for Computational Linguistics}, pages 169--176.

\bibitem[{Cahyawijaya(2024)}]{cahyawijaya2024llm}
Samuel Cahyawijaya. 2024.
\newblock \emph{Llm for everyone: Representing the underrepresented in large language models}.
\newblock Ph.D. thesis, Hong Kong University of Science and Technology (Hong Kong).

\bibitem[{Cahyawijaya et~al.(2024{\natexlab{a}})Cahyawijaya, Chen, Bang, Khalatbari, Wilie, Ji, Ishii, and Fung}]{cahyawijaya2024high}
Samuel Cahyawijaya, Delong Chen, Yejin Bang, Leila Khalatbari, Bryan Wilie, Ziwei Ji, Etsuko Ishii, and Pascale Fung. 2024{\natexlab{a}}.
\newblock High-dimension human value representation in large language models.
\newblock \emph{arXiv preprint arXiv:2404.07900}.

\bibitem[{Cahyawijaya et~al.(2024{\natexlab{b}})Cahyawijaya, Lovenia, and Fung}]{cahyawijaya-etal-2024-llms}
Samuel Cahyawijaya, Holy Lovenia, and Pascale Fung. 2024{\natexlab{b}}.
\newblock \href {https://doi.org/10.18653/v1/2024.naacl-long.24} {{LLM}s are few-shot in-context low-resource language learners}.
\newblock In \emph{Proceedings of the 2024 Conference of the North American Chapter of the Association for Computational Linguistics: Human Language Technologies (Volume 1: Long Papers)}, pages 405--433, Mexico City, Mexico. Association for Computational Linguistics.

\bibitem[{Cahyawijaya et~al.(2023)Cahyawijaya, Lovenia, Yu, Chung, and Fung}]{cahyawijaya2023instructalign}
Samuel Cahyawijaya, Holy Lovenia, Tiezheng Yu, Willy Chung, and Pascale Fung. 2023.
\newblock Instructalign: High-and-low resource language alignment via continual crosslingual instruction tuning.
\newblock In \emph{Proceedings of the First Workshop in South East Asian Language Processing}, pages 55--78.

\bibitem[{Chang et~al.(2022)Chang, Tu, and Bergen}]{chang-etal-2022-geometry}
Tyler Chang, Zhuowen Tu, and Benjamin Bergen. 2022.
\newblock \href {https://doi.org/10.18653/v1/2022.emnlp-main.9} {The geometry of multilingual language model representations}.
\newblock In \emph{Proceedings of the 2022 Conference on Empirical Methods in Natural Language Processing}, pages 119--136, Abu Dhabi, United Arab Emirates. Association for Computational Linguistics.

\bibitem[{Chen et~al.(2024)Chen, Zheng, Wu, Gong, Zhang, and Li}]{chen-etal-2024-breaking}
Nuo Chen, Zinan Zheng, Ning Wu, Ming Gong, Dongmei Zhang, and Jia Li. 2024.
\newblock \href {https://doi.org/10.18653/v1/2024.findings-emnlp.411} {Breaking language barriers in multilingual mathematical reasoning: Insights and observations}.
\newblock In \emph{Findings of the Association for Computational Linguistics: EMNLP 2024}, pages 7001--7016, Miami, Florida, USA. Association for Computational Linguistics.

\bibitem[{Cobbe et~al.(2021)Cobbe, Kosaraju, Bavarian, Chen, Jun, Kaiser, Plappert, Tworek, Hilton, Nakano et~al.}]{cobbe2021training}
Karl Cobbe, Vineet Kosaraju, Mohammad Bavarian, Mark Chen, Heewoo Jun, Lukasz Kaiser, Matthias Plappert, Jerry Tworek, Jacob Hilton, Reiichiro Nakano, et~al. 2021.
\newblock Training verifiers to solve math word problems.
\newblock \emph{arXiv preprint arXiv:2110.14168}.

\bibitem[{Conneau et~al.(2020{\natexlab{a}})Conneau, Khandelwal, Goyal, Chaudhary, Wenzek, Guzm{\'a}n, Grave, Ott, Zettlemoyer, and Stoyanov}]{conneau-etal-2020-unsupervised}
Alexis Conneau, Kartikay Khandelwal, Naman Goyal, Vishrav Chaudhary, Guillaume Wenzek, Francisco Guzm{\'a}n, Edouard Grave, Myle Ott, Luke Zettlemoyer, and Veselin Stoyanov. 2020{\natexlab{a}}.
\newblock \href {https://doi.org/10.18653/v1/2020.acl-main.747} {Unsupervised cross-lingual representation learning at scale}.
\newblock In \emph{Proceedings of the 58th Annual Meeting of the Association for Computational Linguistics}, pages 8440--8451, Online. Association for Computational Linguistics.

\bibitem[{Conneau et~al.(2020{\natexlab{b}})Conneau, Wu, Li, Zettlemoyer, and Stoyanov}]{conneau2020emerging}
Alexis Conneau, Shijie Wu, Haoran Li, Luke Zettlemoyer, and Veselin Stoyanov. 2020{\natexlab{b}}.
\newblock Emerging cross-lingual structure in pretrained language models.
\newblock In \emph{Proceedings of the 58th Annual Meeting of the Association for Computational Linguistics}, pages 6022--6034.

\bibitem[{Dang et~al.(2024)Dang, Singh, D'souza, Ahmadian, Salamanca, Smith, Peppin, Hong, Govindassamy, Zhao et~al.}]{dang2024aya}
John Dang, Shivalika Singh, Daniel D'souza, Arash Ahmadian, Alejandro Salamanca, Madeline Smith, Aidan Peppin, Sungjin Hong, Manoj Govindassamy, Terrence Zhao, et~al. 2024.
\newblock Aya expanse: Combining research breakthroughs for a new multilingual frontier.
\newblock \emph{arXiv preprint arXiv:2412.04261}.

\bibitem[{Del and Fishel(2022)}]{del2022cross}
Maksym Del and Mark Fishel. 2022.
\newblock Cross-lingual similarity of multilingual representations revisited.
\newblock In \emph{Proceedings of the 2nd Conference of the Asia-Pacific Chapter of the Association for Computational Linguistics and the 12th International Joint Conference on Natural Language Processing (Volume 1: Long Papers)}, pages 185--195.

\bibitem[{Dubey et~al.(2024)Dubey, Jauhri, Pandey, Kadian, Al-Dahle, Letman, Mathur, Schelten, Yang, Fan et~al.}]{dubey2024llama}
Abhimanyu Dubey, Abhinav Jauhri, Abhinav Pandey, Abhishek Kadian, Ahmad Al-Dahle, Aiesha Letman, Akhil Mathur, Alan Schelten, Amy Yang, Angela Fan, et~al. 2024.
\newblock The llama 3 herd of models.
\newblock \emph{arXiv preprint arXiv:2407.21783}.

\bibitem[{Eberhard et~al.(2024)Eberhard, Simons, and Fennig}]{ethnologue}
David~M. Eberhard, Gary~F. Simons, and Charles~D. Fennig. 2024.
\newblock \href {http://www.ethnologue.com} {\emph{Ethnologue: Languages of the World}}.
\newblock SIL International, Dallas, Texas.

\bibitem[{Engels et~al.(2025)Engels, Michaud, Liao, Gurnee, and Tegmark}]{engels2025not}
Joshua Engels, Eric~J Michaud, Isaac Liao, Wes Gurnee, and Max Tegmark. 2025.
\newblock \href {https://openreview.net/forum?id=d63a4AM4hb} {Not all language model features are linear}.
\newblock In \emph{The Thirteenth International Conference on Learning Representations}.

\bibitem[{Feng et~al.(2022)Feng, Yang, Cer, Arivazhagan, and Wang}]{feng-etal-2022-language}
Fangxiaoyu Feng, Yinfei Yang, Daniel Cer, Naveen Arivazhagan, and Wei Wang. 2022.
\newblock \href {https://doi.org/10.18653/v1/2022.acl-long.62} {Language-agnostic {BERT} sentence embedding}.
\newblock In \emph{Proceedings of the 60th Annual Meeting of the Association for Computational Linguistics (Volume 1: Long Papers)}, pages 878--891, Dublin, Ireland. Association for Computational Linguistics.

\bibitem[{Freeman et~al.(2002)}]{freeman2002centrality}
Linton~C Freeman et~al. 2002.
\newblock Centrality in social networks: Conceptual clarification.
\newblock \emph{Social network: critical concepts in sociology. Londres: Routledge}, 1:238--263.

\bibitem[{French(1999)}]{french1999catastrophic}
Robert~M French. 1999.
\newblock Catastrophic forgetting in connectionist networks.
\newblock \emph{Trends in cognitive sciences}, 3(4):128--135.

\bibitem[{Fung and Chen(2004)}]{fung2004biframenet}
Pascale Fung and Benfeng Chen. 2004.
\newblock Biframenet: bilingual frame semantics resource construction by cross-lingual induction.
\newblock In \emph{COLING 2004: Proceedings of the 20th International Conference on Computational Linguistics}, pages 931--937.

\bibitem[{Fung and Church(1994)}]{fung-church-1994-k}
Pascale Fung and Kenneth~Ward Church. 1994.
\newblock \href {https://aclanthology.org/C94-2178} {K-vec: A new approach for aligning parallel texts}.
\newblock In \emph{{COLING} 1994 Volume 2: The 15th {I}nternational {C}onference on {C}omputational {L}inguistics}.

\bibitem[{Fung and Mckeown(1994)}]{fung1994aligning}
Pascale Fung and Kathleen Mckeown. 1994.
\newblock Aligning noisy parallel corpora across language groups: Word pair feature matching by dynamic time warping.
\newblock In \emph{Proceedings of the First Conference of the Association for Machine Translation in the Americas}.

\bibitem[{Georgi et~al.(2010)Georgi, Xia, and Lewis}]{georgi2010comparing}
Ryan Georgi, Fei Xia, and William Lewis. 2010.
\newblock Comparing language similarity across genetic and typologically-based groupings.
\newblock In \emph{Proceedings of the 23rd international conference on computational linguistics (Coling 2010)}, pages 385--393.

\bibitem[{Grosse et~al.(2023)Grosse, Bae, Anil, Elhage, Tamkin, Tajdini, Steiner, Li, Durmus, Perez et~al.}]{grosse2023studying}
Roger Grosse, Juhan Bae, Cem Anil, Nelson Elhage, Alex Tamkin, Amirhossein Tajdini, Benoit Steiner, Dustin Li, Esin Durmus, Ethan Perez, et~al. 2023.
\newblock Studying large language model generalization with influence functions.
\newblock \emph{arXiv preprint arXiv:2308.03296}.

\bibitem[{Guimera and Amaral(2005)}]{guimera2005cartography}
Roger Guimera and Lu{\'\i}s A~Nunes Amaral. 2005.
\newblock Cartography of complex networks: modules and universal roles.
\newblock \emph{Journal of Statistical Mechanics: Theory and Experiment}, 2005(02):P02001.

\bibitem[{Ilharco et~al.(2022)Ilharco, Ribeiro, Wortsman, Schmidt, Hajishirzi, and Farhadi}]{ilharcoediting}
Gabriel Ilharco, Marco~Tulio Ribeiro, Mitchell Wortsman, Ludwig Schmidt, Hannaneh Hajishirzi, and Ali Farhadi. 2022.
\newblock Editing models with task arithmetic.
\newblock In \emph{The Eleventh International Conference on Learning Representations}.

\bibitem[{Ji et~al.(2024)Ji, Chen, Ishii, Cahyawijaya, Bang, Wilie, and Fung}]{ji2024llm}
Ziwei Ji, Delong Chen, Etsuko Ishii, Samuel Cahyawijaya, Yejin Bang, Bryan Wilie, and Pascale Fung. 2024.
\newblock Llm internal states reveal hallucination risk faced with a query.
\newblock In \emph{Proceedings of the 7th BlackboxNLP Workshop: Analyzing and Interpreting Neural Networks for NLP}, pages 88--104.

\bibitem[{Johnson et~al.(2017)Johnson, Schuster, Le, Krikun, Wu, Chen, Thorat, Vi{\'e}gas, Wattenberg, Corrado et~al.}]{johnson2017google}
Melvin Johnson, Mike Schuster, Quoc~V Le, Maxim Krikun, Yonghui Wu, Zhifeng Chen, Nikhil Thorat, Fernanda Vi{\'e}gas, Martin Wattenberg, Greg Corrado, et~al. 2017.
\newblock Google’s multilingual neural machine translation system: Enabling zero-shot translation.
\newblock \emph{Transactions of the Association for Computational Linguistics}, 5:339--351.

\bibitem[{Kanzaki et~al.(2008)Kanzaki, Nakao, Rayner, Santaholma, Starlander, and Tsourakis}]{kanzaki-etal-2008-many}
Kyoko Kanzaki, Yukie Nakao, Manny Rayner, Marianne Santaholma, Marianne Starlander, and Nikos Tsourakis. 2008.
\newblock \href {https://aclanthology.org/2008.amta-govandcom.4/} {Many-to-many multilingual medical speech translation on a {PDA}}.
\newblock In \emph{Proceedings of the 8th Conference of the Association for Machine Translation in the Americas: Government and Commercial Uses of MT}, Waikiki, USA. Association for Machine Translation in the Americas.

\bibitem[{Libovick{\`y} et~al.(2020)Libovick{\`y}, Rosa, and Fraser}]{libovicky2020language}
Jind{\v{r}}ich Libovick{\`y}, Rudolf Rosa, and Alexander Fraser. 2020.
\newblock On the language neutrality of pre-trained multilingual representations.
\newblock In \emph{Findings of the Association for Computational Linguistics: EMNLP 2020}, pages 1663--1674.

\bibitem[{Liu et~al.(2024)Liu, Chen, Cheng, and He}]{liu2024universal}
Junteng Liu, Shiqi Chen, Yu~Cheng, and Junxian He. 2024.
\newblock On the universal truthfulness hyperplane inside llms.
\newblock In \emph{Proceedings of the 2024 Conference on Empirical Methods in Natural Language Processing}, pages 18199--18224.

\bibitem[{Liu et~al.(2021)Liu, Winata, Madotto, and Fung}]{liu2021preserving}
Zihan Liu, Genta~Indra Winata, Andrea Madotto, and Pascale Fung. 2021.
\newblock Preserving cross-linguality of pre-trained models via continual learning.
\newblock In \emph{Proceedings of the 6th Workshop on Representation Learning for NLP (RepL4NLP-2021)}, pages 64--71.

\bibitem[{Lopez(2008)}]{lopez2008pbsmt}
Adam Lopez. 2008.
\newblock \href {https://doi.org/10.1145/1380584.1380586} {Statistical machine translation}.
\newblock \emph{ACM Comput. Surv.}, 40(3).

\bibitem[{Lu et~al.(2018)Lu, Keung, Ladhak, Bhardwaj, Zhang, and Sun}]{lu2018neural}
Yichao Lu, Phillip Keung, Faisal Ladhak, Vikas Bhardwaj, Shaonan Zhang, and Jason Sun. 2018.
\newblock A neural interlingua for multilingual machine translation.
\newblock \emph{arXiv preprint arXiv:1804.08198}.

\bibitem[{McCloskey and Cohen(1989)}]{mccloskey1989catastrophic}
Michael McCloskey and Neal~J Cohen. 1989.
\newblock Catastrophic interference in connectionist networks: The sequential learning problem.
\newblock In \emph{Psychology of learning and motivation}, volume~24, pages 109--165. Elsevier.

\bibitem[{Miceli~Barone(2016)}]{miceli-barone-2016-towards}
Antonio~Valerio Miceli~Barone. 2016.
\newblock \href {https://doi.org/10.18653/v1/W16-1614} {Towards cross-lingual distributed representations without parallel text trained with adversarial autoencoders}.
\newblock In \emph{Proceedings of the 1st Workshop on Representation Learning for {NLP}}, pages 121--126, Berlin, Germany. Association for Computational Linguistics.

\bibitem[{Mijalkov et~al.(2017)Mijalkov, Kakaei, Pereira, Westman, Volpe, and Initiative}]{mijalkov2017braph}
Mite Mijalkov, Ehsan Kakaei, Joana~B Pereira, Eric Westman, Giovanni Volpe, and Alzheimer's Disease~Neuroimaging Initiative. 2017.
\newblock Braph: a graph theory software for the analysis of brain connectivity.
\newblock \emph{PloS one}, 12(8):e0178798.

\bibitem[{Mikolov et~al.(2013)Mikolov, Le, and Sutskever}]{mikolov2013exploiting}
Tomas Mikolov, Quoc~V Le, and Ilya Sutskever. 2013.
\newblock Exploiting similarities among languages for machine translation.
\newblock \emph{arXiv preprint arXiv:1309.4168}.

\bibitem[{Moschella et~al.(2023)Moschella, Maiorca, Fumero, Norelli, Locatello, and Rodol{\`a}}]{moschella2023relative}
Luca Moschella, Valentino Maiorca, Marco Fumero, Antonio Norelli, Francesco Locatello, and Emanuele Rodol{\`a}. 2023.
\newblock \href {https://openreview.net/forum?id=SrC-nwieGJ} {Relative representations enable zero-shot latent space communication}.
\newblock In \emph{The Eleventh International Conference on Learning Representations}.

\bibitem[{Muller et~al.(2021)Muller, Elazar, Sagot, and Seddah}]{muller2021first}
Benjamin Muller, Yanai Elazar, Beno{\^\i}t Sagot, and Djam{\'e} Seddah. 2021.
\newblock First align, then predict: Understanding the cross-lingual ability of multilingual bert.
\newblock In \emph{Proceedings of the 16th Conference of the European Chapter of the Association for Computational Linguistics: Main Volume}, pages 2214--2231.

\bibitem[{Och and Ney(2003)}]{och-ney-2003-systematic}
Franz~Josef Och and Hermann Ney. 2003.
\newblock \href {https://doi.org/10.1162/089120103321337421} {A systematic comparison of various statistical alignment models}.
\newblock \emph{Computational Linguistics}, 29(1):19--51.

\bibitem[{Och et~al.(1999)Och, Tillmann, and Ney}]{och-etal-1999-improved}
Franz~Josef Och, Christoph Tillmann, and Hermann Ney. 1999.
\newblock \href {https://aclanthology.org/W99-0604} {Improved alignment models for statistical machine translation}.
\newblock In \emph{1999 Joint {SIGDAT} Conference on Empirical Methods in Natural Language Processing and Very Large Corpora}.

\bibitem[{Philippy et~al.(2023)Philippy, Guo, and Haddadan}]{philippy-etal-2023-towards}
Fred Philippy, Siwen Guo, and Shohreh Haddadan. 2023.
\newblock \href {https://doi.org/10.18653/v1/2023.acl-long.323} {Towards a common understanding of contributing factors for cross-lingual transfer in multilingual language models: A review}.
\newblock In \emph{Proceedings of the 61st Annual Meeting of the Association for Computational Linguistics (Volume 1: Long Papers)}, pages 5877--5891, Toronto, Canada. Association for Computational Linguistics.

\bibitem[{Pires et~al.(2019)Pires, Schlinger, and Garrette}]{pires-etal-2019-multilingual}
Telmo Pires, Eva Schlinger, and Dan Garrette. 2019.
\newblock \href {https://doi.org/10.18653/v1/P19-1493} {How multilingual is multilingual {BERT}?}
\newblock In \emph{Proceedings of the 57th Annual Meeting of the Association for Computational Linguistics}, pages 4996--5001, Florence, Italy. Association for Computational Linguistics.

\bibitem[{Ranta et~al.(2020)Ranta, Angelov, Gruzitis, and Kolachina}]{ranta-etal-2020-abstract}
Aarne Ranta, Krasimir Angelov, Normunds Gruzitis, and Prasanth Kolachina. 2020.
\newblock \href {https://doi.org/10.1162/coli_a_00378} {Abstract syntax as interlingua: Scaling up the grammatical framework from controlled languages to robust pipelines}.
\newblock \emph{Computational Linguistics}, 46(2):425--486.

\bibitem[{Rayner(2000)}]{rayner2000spoken}
Manny Rayner. 2000.
\newblock \emph{The spoken language translator}.
\newblock Cambridge University Press.

\bibitem[{Rayner et~al.(2008)Rayner, Bouillon, Hockey, and Nakao}]{rayner-etal-2008-almost}
Manny Rayner, Pierrette Bouillon, Beth~Ann Hockey, and Yukie Nakao. 2008.
\newblock \href {https://aclanthology.org/C08-1090/} {Almost flat functional semantics for speech translation}.
\newblock In \emph{Proceedings of the 22nd International Conference on Computational Linguistics (Coling 2008)}, pages 713--720, Manchester, UK. Coling 2008 Organizing Committee.

\bibitem[{Rayner et~al.(2010{\natexlab{a}})Rayner, Estrella, and Bouillon}]{rayner2010bootstrapped}
Manny Rayner, Paula Estrella, and Pierrette Bouillon. 2010{\natexlab{a}}.
\newblock A bootstrapped interlingua-based smt architecture.
\newblock In \emph{Proceedings of the 14th Annual Conference of the European Association for Machine Translation}.

\bibitem[{Rayner et~al.(2010{\natexlab{b}})Rayner, Estrella, and Bouillon}]{rayner-etal-2010-bootstrapped}
Manny Rayner, Paula Estrella, and Pierrette Bouillon. 2010{\natexlab{b}}.
\newblock \href {https://aclanthology.org/2010.eamt-1.39/} {A bootstrapped interlingua-based {SMT} architecture}.
\newblock In \emph{Proceedings of the 14th Annual Conference of the European Association for Machine Translation}, Saint Rapha{\"e}l, France. European Association for Machine Translation.

\bibitem[{Richens(1958)}]{richens1958interlingualmt}
R.~H. Richens. 1958.
\newblock \href {https://doi.org/10.1093/comjnl/1.3.144} {Interlingual machine translation}.
\newblock \emph{The Computer Journal}, 1(3):144--147.

\bibitem[{Riezler et~al.(2002)Riezler, King, Kaplan, Crouch, Maxwell~III, and Johnson}]{riezler-etal-2002-parsing}
Stefan Riezler, Tracy~H. King, Ronald~M. Kaplan, Richard Crouch, John~T. Maxwell~III, and Mark Johnson. 2002.
\newblock \href {https://doi.org/10.3115/1073083.1073129} {Parsing the {W}all {S}treet {J}ournal using a {L}exical-{F}unctional {G}rammar and discriminative estimation techniques}.
\newblock In \emph{Proceedings of the 40th Annual Meeting of the Association for Computational Linguistics}, pages 271--278, Philadelphia, Pennsylvania, USA. Association for Computational Linguistics.

\bibitem[{Schubert(1989)}]{schubert1989interlinguistics}
Klaus Schubert. 1989.
\newblock Interlinguistics--its aims, its achievements, and its place in language science.
\newblock \emph{Interlinguistics: Aspects of the Science of Planned Languages. Trends in Linguistics}, 42:7--44.

\bibitem[{Seneff(2006)}]{seneff-2006-combining}
Stephanie Seneff. 2006.
\newblock \href {https://aclanthology.org/2006.amta-panel2.4/} {Combining interlingua with {SMT}}.
\newblock In \emph{Proceedings of the 7th Conference of the Association for Machine Translation in the Americas: Panel on hybrid machine translation: why and how?}, Cambridge, Massachusetts, USA. Association for Machine Translation in the Americas.

\bibitem[{Shi et~al.(2022)Shi, Suzgun, Freitag, Wang, Srivats, Vosoughi, Chung, Tay, Ruder, Zhou et~al.}]{shilanguage}
Freda Shi, Mirac Suzgun, Markus Freitag, Xuezhi Wang, Suraj Srivats, Soroush Vosoughi, Hyung~Won Chung, Yi~Tay, Sebastian Ruder, Denny Zhou, et~al. 2022.
\newblock Language models are multilingual chain-of-thought reasoners.
\newblock In \emph{The Eleventh International Conference on Learning Representations}.

\bibitem[{Team et~al.(2024)Team, Riviere, Pathak, Sessa, Hardin, Bhupatiraju, Hussenot, Mesnard, Shahriari, Ram{\'e} et~al.}]{team2024gemma}
Gemma Team, Morgane Riviere, Shreya Pathak, Pier~Giuseppe Sessa, Cassidy Hardin, Surya Bhupatiraju, L{\'e}onard Hussenot, Thomas Mesnard, Bobak Shahriari, Alexandre Ram{\'e}, et~al. 2024.
\newblock Gemma 2: Improving open language models at a practical size.
\newblock \emph{arXiv preprint arXiv:2408.00118}.

\bibitem[{Team(2024)}]{nllb2024scaling}
NLLB Team. 2024.
\newblock Scaling neural machine translation to 200 languages.
\newblock \emph{Nature}, pages 1--6.

\bibitem[{Van~der Maaten and Hinton(2008)}]{van2008visualizing}
Laurens Van~der Maaten and Geoffrey Hinton. 2008.
\newblock Visualizing data using t-sne.
\newblock \emph{Journal of machine learning research}, 9(11).

\bibitem[{Vauquois(1968)}]{Vauquois1968ASO}
Bernard Vauquois. 1968.
\newblock \href {https://api.semanticscholar.org/CorpusID:23199204} {A survey of formal grammars and algorithms for recognition and transformation in mechanical translation}.
\newblock In \emph{IFIP Congress}.

\bibitem[{Wahlster(2013)}]{wahlster2013verbmobil}
Wolfgang Wahlster. 2013.
\newblock \emph{Verbmobil: foundations of speech-to-speech translation}.
\newblock Springer Science \& Business Media.

\bibitem[{Wei et~al.(2021)Wei, Weng, Hu, Xing, Yu, and Luo}]{wei2021learning}
Xiangpeng Wei, Rongxiang Weng, Yue Hu, Luxi Xing, Heng Yu, and Weihua Luo. 2021.
\newblock On learning universal representations across languages.
\newblock In \emph{International Conference on Learning Representations}.

\bibitem[{Wendler et~al.(2024)Wendler, Veselovsky, Monea, and West}]{wendler-etal-2024-llamas}
Chris Wendler, Veniamin Veselovsky, Giovanni Monea, and Robert West. 2024.
\newblock \href {https://doi.org/10.18653/v1/2024.acl-long.820} {Do llamas work in {E}nglish? on the latent language of multilingual transformers}.
\newblock In \emph{Proceedings of the 62nd Annual Meeting of the Association for Computational Linguistics (Volume 1: Long Papers)}, pages 15366--15394, Bangkok, Thailand. Association for Computational Linguistics.

\bibitem[{Winata et~al.(2023)Winata, Xie, Radhakrishnan, Wu, Jin, Cheng, Kulkarni, and Preo{\c{t}}iuc-Pietro}]{winata2023overcoming}
Genta Winata, Lingjue Xie, Karthik Radhakrishnan, Shijie Wu, Xisen Jin, Pengxiang Cheng, Mayank Kulkarni, and Daniel Preo{\c{t}}iuc-Pietro. 2023.
\newblock Overcoming catastrophic forgetting in massively multilingual continual learning.
\newblock In \emph{Findings of the Association for Computational Linguistics: ACL 2023}, pages 768--777.

\bibitem[{Yang et~al.(2024)Yang, Yang, Zhang, Hui, Zheng, Yu, Li, Liu, Huang, Wei et~al.}]{yang2024qwen2}
An~Yang, Baosong Yang, Beichen Zhang, Binyuan Hui, Bo~Zheng, Bowen Yu, Chengyuan Li, Dayiheng Liu, Fei Huang, Haoran Wei, et~al. 2024.
\newblock Qwen2. 5 technical report.
\newblock \emph{arXiv preprint arXiv:2412.15115}.

\bibitem[{Zeng et~al.(2025)Zeng, Han, Chen, and Yu}]{zeng2025converging}
Hongchuan Zeng, Senyu Han, Lu~Chen, and Kai Yu. 2025.
\newblock Converging to a lingua franca: Evolution of linguistic regions and semantics alignment in multilingual large language models.
\newblock In \emph{Proceedings of the 31st International Conference on Computational Linguistics}, pages 10602--10617.

\bibitem[{Zhao et~al.(2024)Zhao, Zhang, Chen, Kawaguchi, and Bing}]{zhao2024how}
Yiran Zhao, Wenxuan Zhang, Guizhen Chen, Kenji Kawaguchi, and Lidong Bing. 2024.
\newblock \href {https://openreview.net/forum?id=ctXYOoAgRy} {How do large language models handle multilingualism?}
\newblock In \emph{The Thirty-eighth Annual Conference on Neural Information Processing Systems}.

\bibitem[{Zhu et~al.(2020)Zhu, Yu, Cheng, and Luo}]{zhu2020language}
Changfeng Zhu, Heng Yu, Shanbo Cheng, and Weihua Luo. 2020.
\newblock Language-aware interlingua for multilingual neural machine translation.
\newblock In \emph{Proceedings of the 58th Annual Meeting of the Association for Computational Linguistics}, pages 1650--1655.

\bibitem[{Zhu et~al.(2024)Zhu, Huang, Yuan, She, Chen, and Birch}]{zhu2024question}
Wenhao Zhu, Shujian Huang, Fei Yuan, Shuaijie She, Jiajun Chen, and Alexandra Birch. 2024.
\newblock Question translation training for better multilingual reasoning.
\newblock \emph{arXiv preprint arXiv:2401.07817}.

\end{thebibliography}

\newpage
\appendix

\setcounter{table}{0}
\renewcommand{\thetable}{A\arabic{table}}
\setcounter{figure}{0}
\renewcommand{\thefigure}{A\arabic{figure}}

\clearpage
\section*{Appendix}

\section{Details on Linguistic Properties}
\label{app:lang_details}

We provide the detail of the region and linguistic properties of the language subsets sampled from Flores-200 in~\ref{tab:complete_langs}. Here, while most of them are extracted from~\citet{nllb2024scaling}, we refer to~\cite{ethnologue} for the details on linguistic families.

\begin{table}[!ht]
\centering
\resizebox{\linewidth}{!}{%
\begin{tabular}{lllllc}
\toprule
\textbf{Code} & \textbf{Language} & \textbf{Script} & \textbf{Region} & \textbf{Family} & \textbf{Res.} \\
\midrule
ban\_Latn & Balinese & Latin & Southeast Asia & Austronesian & Low \\
ben\_Beng & Bengali & Bengali & South Asia & Indo-European & High \\
bjn\_Latn & Banjar & Latin & Southeast Asia & Austronesian & Low \\
ces\_Latn & Czech & Latin & Europe & Indo-European & High \\
dan\_Latn & Danish & Latin & Europe & Indo-European & High \\
deu\_Latn & German & Latin & Europe & Indo-European & High \\
eng\_Latn & English & Latin & Europe & Indo-European & High \\
fra\_Latn & French & Latin & Europe & Indo-European & High \\
gle\_Latn & Irish & Latin & Europe & Indo-European & Low \\
hin\_Deva & Hindi & Devanagari & South Asia & Indo-European & High \\
ind\_Latn & Indonesian & Latin & Southeast Asia & Austronesian & High \\
jav\_Latn & Javanese & Latin & Southeast Asia & Austronesian & Low \\
jpn\_Jpan & Japanese & Japanese & East Asia & Japonic & High \\
min\_Latn & Minangkabau & Latin & Southeast Asia & Austronesian & Low \\
nld\_Latn & Dutch & Latin & Europe & Indo-European & High \\
pol\_Latn & Polish & Latin & Europe & Indo-European & High \\
rus\_Cyrl & Russian & Cyrillic & Europe & Indo-European & High \\
sin\_Sinh & Sinhala & Sinhala & South Asia & Indo-European & Low \\
slv\_Latn & Slovenian & Latin & Europe & Indo-European & High \\
spa\_Latn & Spanish & Latin & Europe & Indo-European & High \\
srp\_Cyrl & Serbian & Cyrillic & Europe & Indo-European & Low \\
sun\_Latn & Sundanese & Latin & Southeast Asia & Austronesian & Low \\
swe\_Latn & Swedish & Latin & Europe & Indo-European & High \\
swh\_Latn & Swahili & Latin & Africa & Niger-Congo & High \\
tel\_Telu & Telugu & Telugu & South Asia & Dravidian & Low \\
tgl\_Latn & Tagalog & Latin & Southeast Asia & Austronesian & Low \\
tha\_Thai & Thai & Thai & Southeast Asia & Kra-Dai & Low \\
ukr\_Cyrl & Ukrainian & Cyrillic & Europe & Indo-European & High \\
urd\_Arab & Urdu & Arabic & South Asia & Indo-European & Low \\
yue\_Hant & Yue Chinese & Han (Traditional) & East Asia & Sino-Tibetan & Low \\
zho\_Hans & Chinese (Simplified) & Han (Simplified) & East Asia & Sino-Tibetan & High \\
\bottomrule
\end{tabular}}
\caption{Complete distribution of the 31 languages across families, regions, and resource-levels in our analysis, sampled from Flores-200}
\label{tab:complete_langs}
\end{table}

\begin{table*}[!ht]
\centering
\scriptsize
\begin{tabularx}{\textwidth}{@{}>{\raggedright\arraybackslash}p{0.9cm}*{6}{>{\centering\arraybackslash}X}@{}}
\toprule
\textbf{Models} & \textbf{Gemma-2 (9B)} & \textbf{Gemma-2 It (8B)} & \textbf{Aya Expanse (8B)} & \textbf{Llama-3.1 (8B)} & \textbf{Llama-3.1 It (8B)} & \textbf{Qwen-2.5 (7B)} \\
\midrule
\multirow{12}{1cm}{\centering\parbox[t]{1cm}{Top \\ language pairs}} 
  & dan\_Latn - swe\_Latn   & eng\_Latn - fra\_Latn   & rus\_Cyrl - ukr\_Cyrl    & yue\_Hant - zho\_Hans    & yue\_Hant - zho\_Hans    & yue\_Hant - zho\_Hans \\[0.6ex]
  & eng\_Latn - fra\_Latn   & dan\_Latn - swe\_Latn   & eng\_Latn - fra\_Latn    & rus\_Cyrl - ukr\_Cyrl    & rus\_Cyrl - ukr\_Cyrl    & dan\_Latn - swe\_Latn \\[0.6ex]
  & rus\_Cyrl - ukr\_Cyrl   & rus\_Cyrl - ukr\_Cyrl   & yue\_Hant - zho\_Hans    & dan\_Latn - swe\_Latn    & dan\_Latn - swe\_Latn    & rus\_Cyrl - ukr\_Cyrl \\[0.6ex]
  & yue\_Hant - zho\_Hans   & deu\_Latn - eng\_Latn   & eng\_Latn - ind\_Latn    & eng\_Latn - fra\_Latn    & eng\_Latn - fra\_Latn    & fra\_Latn - spa\_Latn \\[0.6ex]
  & dan\_Latn - eng\_Latn   & yue\_Hant - zho\_Hans   & fra\_Latn - spa\_Latn    & fra\_Latn - spa\_Latn    & fra\_Latn - spa\_Latn    & eng\_Latn - fra\_Latn \\[0.6ex]
  & eng\_Latn - swe\_Latn   & eng\_Latn - swe\_Latn   & deu\_Latn - eng\_Latn    & deu\_Latn - swe\_Latn    & deu\_Latn - swe\_Latn    & fra\_Latn - rus\_Cyrl \\[0.6ex]
  & deu\_Latn - eng\_Latn   & dan\_Latn - eng\_Latn   & ces\_Latn - rus\_Cyrl    & deu\_Latn - fra\_Latn    & deu\_Latn - fra\_Latn    & rus\_Cyrl - spa\_Latn \\[0.6ex]
  & deu\_Latn - swe\_Latn   & deu\_Latn - fra\_Latn   & ces\_Latn - ukr\_Cyrl    & deu\_Latn - eng\_Latn    & deu\_Latn - eng\_Latn    & deu\_Latn - fra\_Latn \\[0.6ex]
  & deu\_Latn - fra\_Latn   & deu\_Latn - swe\_Latn   & deu\_Latn - fra\_Latn    & deu\_Latn - nld\_Latn    & eng\_Latn - swe\_Latn    & ces\_Latn - pol\_Latn \\[0.6ex]
  & dan\_Latn - deu\_Latn   & dan\_Latn - deu\_Latn   & fra\_Latn - ind\_Latn    & ces\_Latn - rus\_Cyrl    & eng\_Latn - spa\_Latn    & deu\_Latn - nld\_Latn \\
\midrule
\multirow{2.5}{1cm}{Unique languages} 
  & \texttt{swe, dan, fra, eng, ukr, rus, zho, yue, deu, spa}
  & \texttt{fra, eng, swe, dan, rus, ukr, deu, zho, yue, spa}
  & \texttt{rus, ukr, fra, eng, zho, yue, ind, spa, deu, ces}
  & \texttt{yue, zho, ukr, rus, swe, dan, fra, eng, spa, deu}
  & \texttt{zho, yue, rus, ukr, dan, swe, fra, eng, spa, deu}
  & \texttt{yue, zho, dan, swe, ukr, rus, spa, fra, eng, deu} \\
\bottomrule
\end{tabularx}
\caption{Top correlated language pairs from the layers with peak ANC scores and the unique languages from the top language pairs. Most correlated pairs among LLMs are similar on their HRLs. Despite differing rankings, instruction-tuned LLMs exhibit similar sets of top language pairs with its pre-trained counterparts.}
\label{tab:anc_top_languages}
\end{table*}

\section{Further Details on ANC Scores}
\label{app:anc_figures}

Here we provide a detailed view on the ANC comparison of the language pairs for all the model understudy. We compute aggregate peak score for each language pair as the mean over the peak layers. We identify the peak layer by computing the $75^{th}$ percentile of ANCs for each layer and select the top $3$ layers as the peak layers. We denote all the top correlated language pairs from the layers with peak ANC scores and the unique languages from the top language pairs in Table~\ref{tab:anc_top_languages}. We find that the top correlated pairs with high ANCs among the LLMs are similar on their HRLs. Instruction-tuned LLMs exhibit similar sets of top language pairs with its pre-trained counterparts, despite the differing rankings of them.

\section{Visualization and Comparisons For Other Multilingual LLMs}

\subsection{ANC Comparisons from Other LLMs}
\label{app:other_anc}

We attach the complete visualization on ANC scores derived from the hidden-state embeddings of Aya Expanse (8B), Llama-3.1 (8B), Llama-3.1-Instruct (8B), Gemma-2 (9B), Gemma-2-Instruct (9B), and Qwen (9B), respectively in Figures~\ref{fig:anc_aya_complete},~\ref{fig:anc_llama31_complete},~\ref{fig:anc_llama31instruct_complete},~\ref{fig:anc_gemma2_complete},~\ref{fig:anc_gemma2instruct_complete}, and~\ref{fig:anc_qwen_complete}.

\subsection{T-SNE Visualizations from Other LLMs}
\label{app:tsne}

We attach the complete t-SNE visualization projected from the hidden-state embeddings of Aya Expanse (8B), Qwen (9B), Llama-3.1 (8B), Llama-3.1-Instruct (8B), Gemma-2 (9B), and Gemma-2-Instruct (9B), respectively in Figures~\ref{fig:tsne_aya},~\ref{fig:tsne_qwen},~\ref{fig:tsne_llama31}~\ref{fig:tsne_llama31instruct},~\ref{fig:tsne_gemma2}, and~\ref{fig:tsne_gemma2instruct}.

\subsection{Reports on Cross-Lingual Transfer Experiments for Gemma-2 (9B)}
\label{app:ilo_gemma}

\begin{figure*}[!ht]
    \centering
    \includegraphics[trim={0, 0, 0, 0}, clip, width=\linewidth]{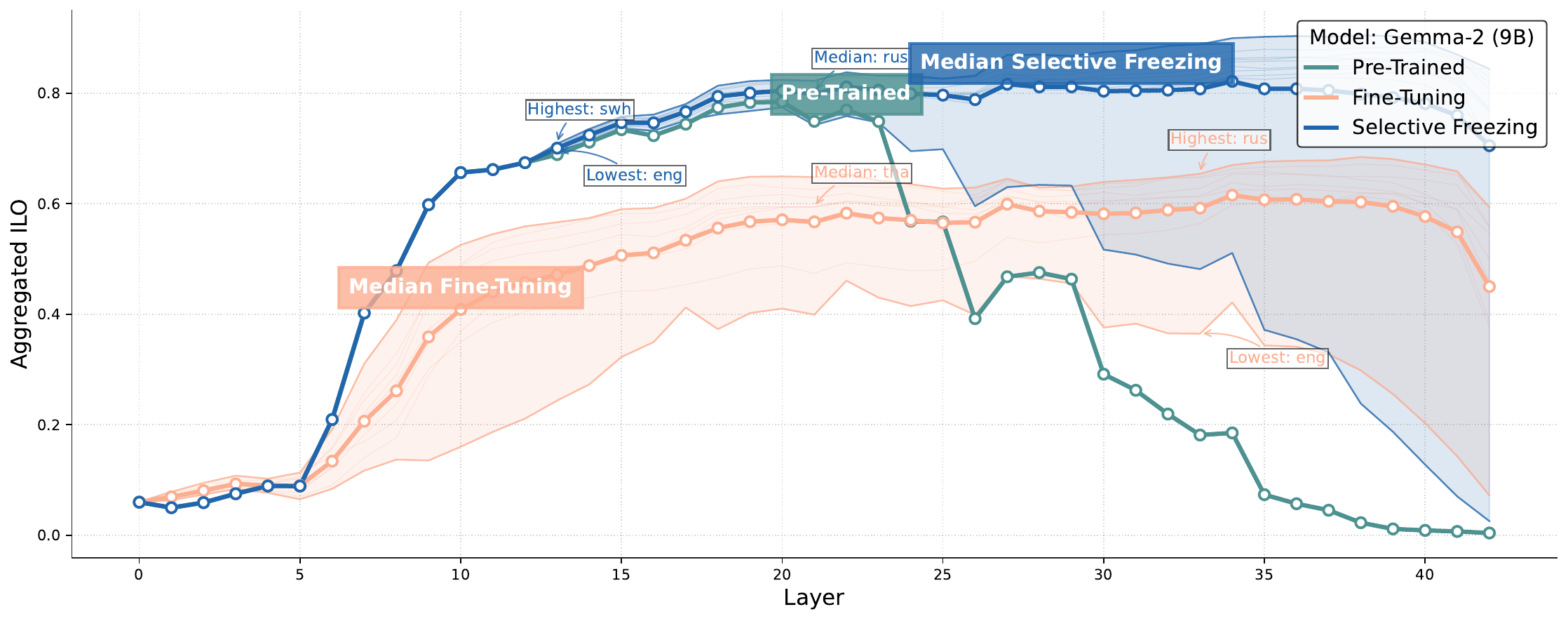}
    \caption{
    Layer-wise \(\bar{\operatorname{\abbrvmetric{}}}_{\mathcal{L}}\) scores for Gemma-2 (9B) in \textbf{pre-trained}, \textbf{fine-tuning}, and \textbf{selective freezing} modes. Notable decrease in alignment from single-language training is seen in the early layers on \textbf{fine-tuning}, whereas the \textbf{selective freezing} mechanism allows the model to sustain its \textbf{pre-trained} semantic alignment across layers.
    }
    \label{fig:interlingual_score_gemma}
\end{figure*}

We attach the cross-lingual transfer performance on MGSM and the layer-wise \(\bar{\operatorname{\abbrvmetric{}}}_{\mathcal{L}}\) scores, for Gemma-2 (9B) in its pre-trained, fine-tuning, and selective-freezing modes, in Table~\ref{tab:xlt_frz_gemma} and Figure~\ref{fig:interlingual_score_gemma}.

\section{Interlingual alignments of various multilingual LLMs}
\label{app:interlingual_alignment_various_model}

In this work, we observe a universal phenomenon that various multilingual LLMs, irrespective of their specific architecture or training data, exhibit a common behavior in constructing an interlingual representation region within their middle layers. However, amongst these similar general trend, we observe that there are different alignment levels across different LLMs in App \ref{app:anc_figures}, \ref{app:other_anc}, and \ref{app:tsne}) 

\begin{figure*}[!t]
  \centering
  \begin{subfigure}[t]{\linewidth}
      \includegraphics[clip, width=\linewidth]{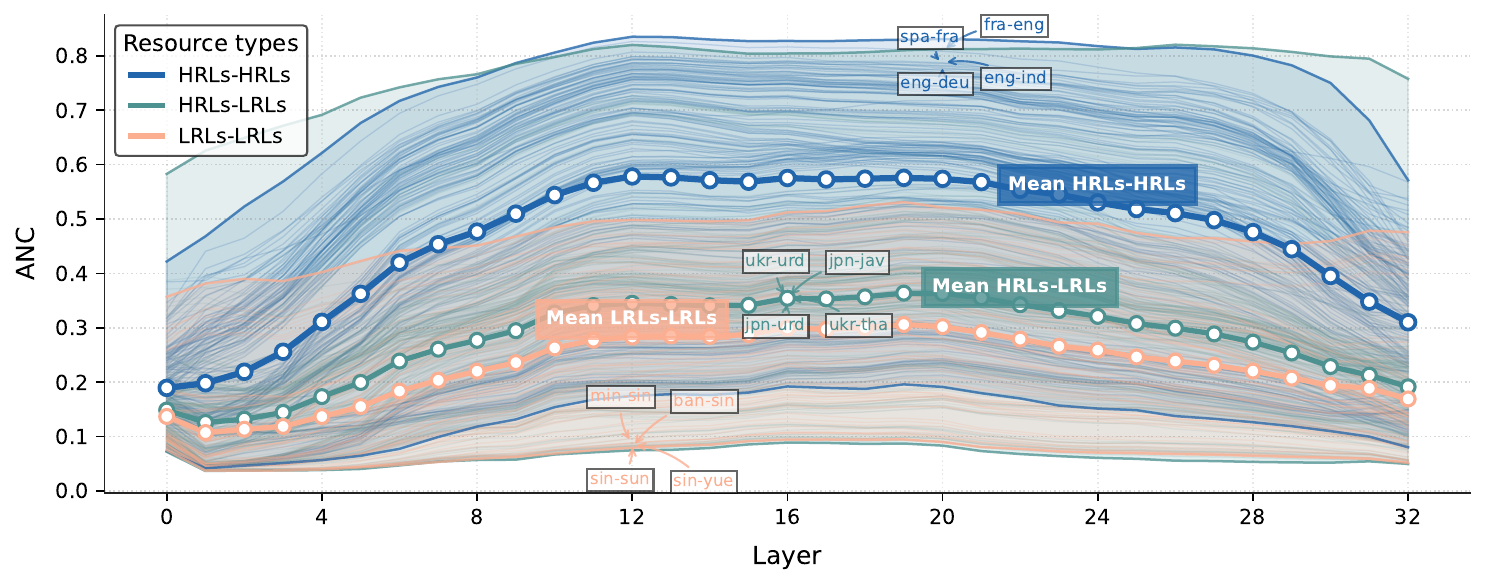}
      \caption{Highlights on pairs w.r.t their resource levels}
      \vspace{5pt}
  \end{subfigure}
  \begin{subfigure}[t]{\linewidth}
      \includegraphics[clip, width=\linewidth]{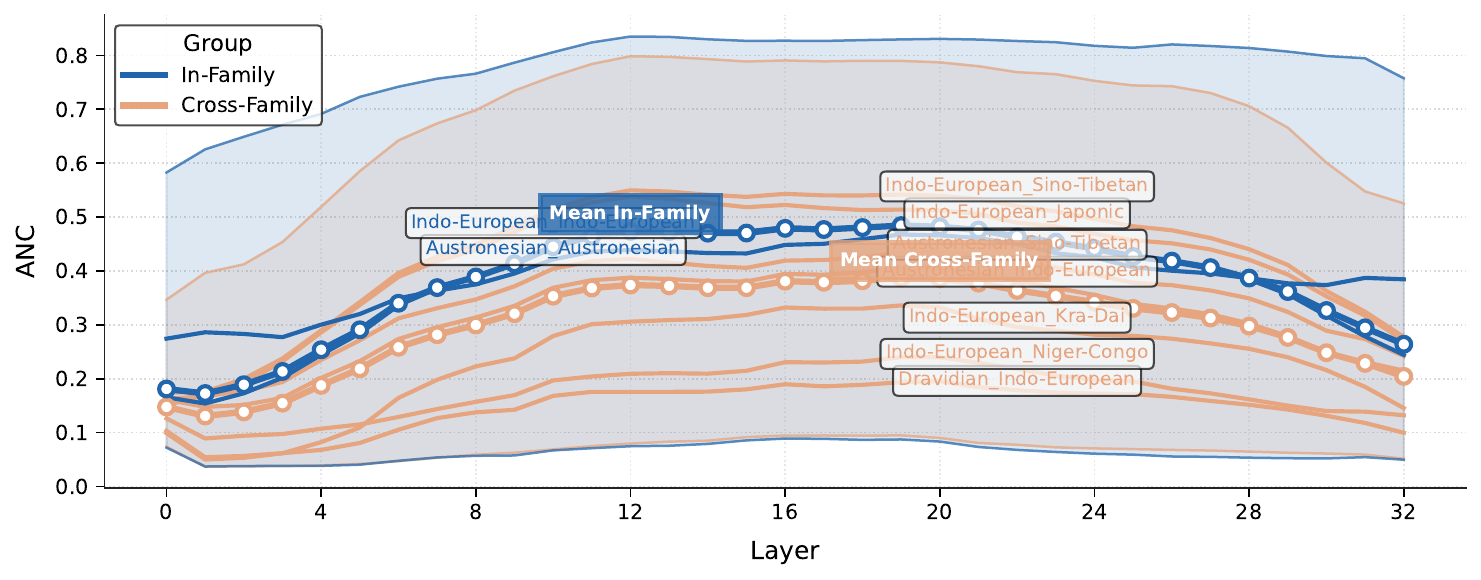}
      \caption{Highlights on pairs w.r.t their linguistic region}
      \vspace{5pt}
  \end{subfigure}
  \begin{subfigure}[t]{\linewidth}
    \centering
      \includegraphics[clip, width=\linewidth]{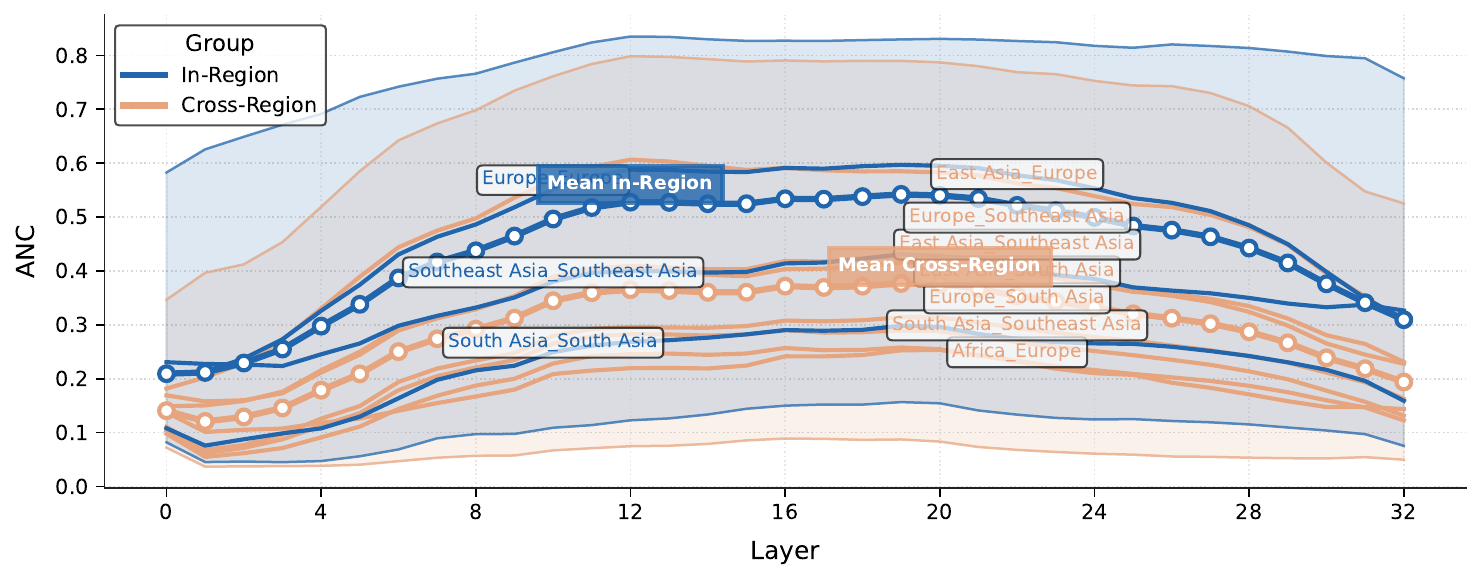}
      \caption{Highlights on pairs w.r.t their linguistic family}
      \vspace{5pt}
  \end{subfigure}
  \caption{Comparisons of per-layer ANC scores on Aya Expanse (8B) with highlights on pairs w.r.t their resource levels, linguistic region and family. Consistently stronger alignments are observed between HRLs pairs and within-group mean correlations.}
  \label{fig:anc_aya_complete}
  \vspace{20pt}
\end{figure*}

\begin{figure*}[!t]
  \centering
  \begin{subfigure}[t]{\linewidth}
      \includegraphics[clip, width=\linewidth]{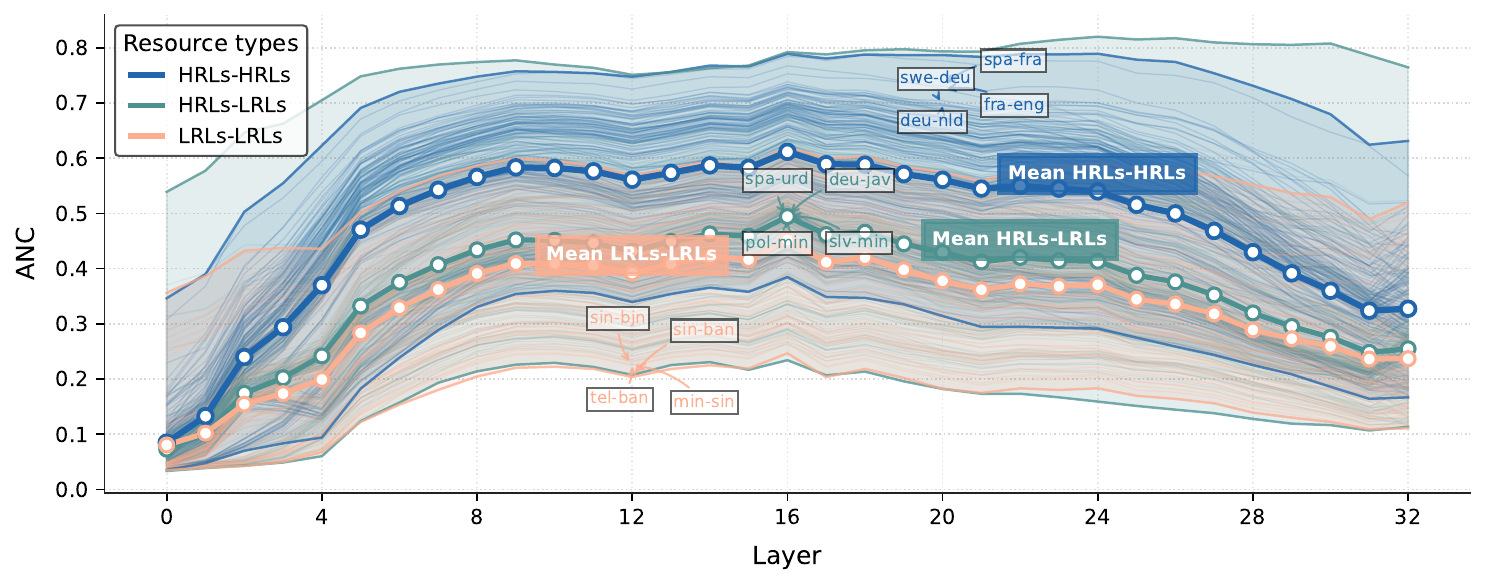}
      \caption{Highlights on pairs w.r.t their resource levels}
      \vspace{5pt}
  \end{subfigure}
  \begin{subfigure}[t]{\linewidth}
      \includegraphics[clip, width=\linewidth]{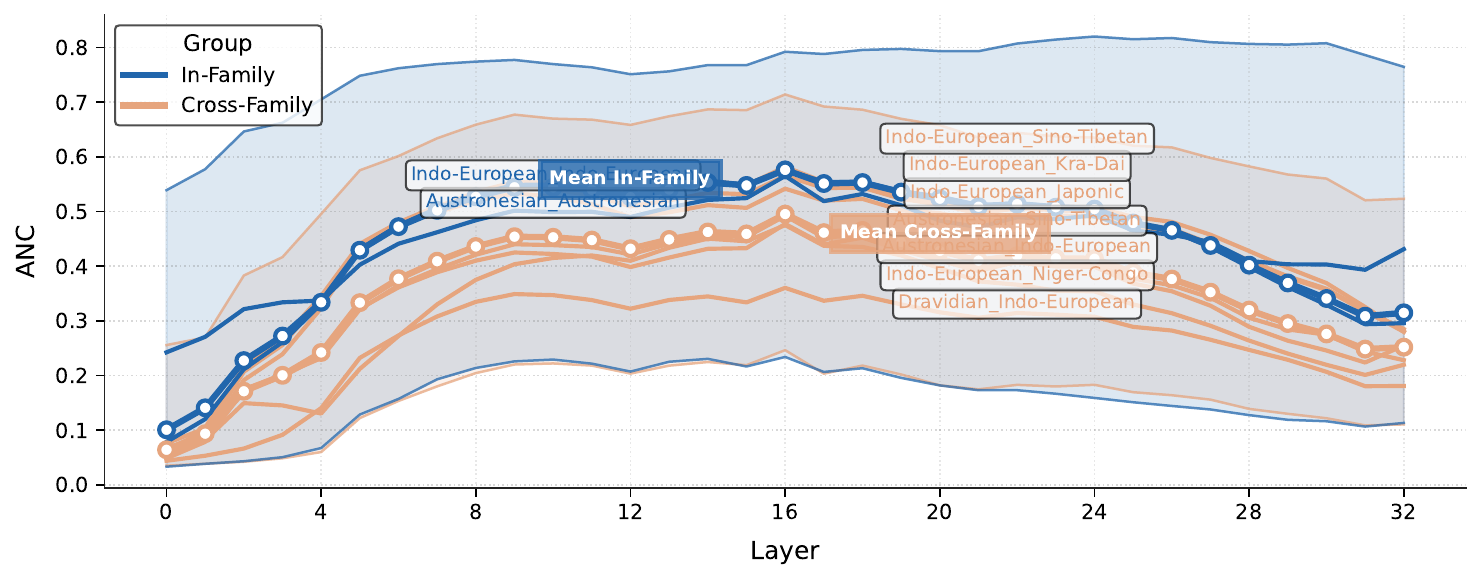}
      \caption{Highlights on pairs w.r.t their linguistic region}
      \vspace{5pt}
  \end{subfigure}
  \begin{subfigure}[t]{\linewidth}
    \centering
      \includegraphics[clip, width=\linewidth]{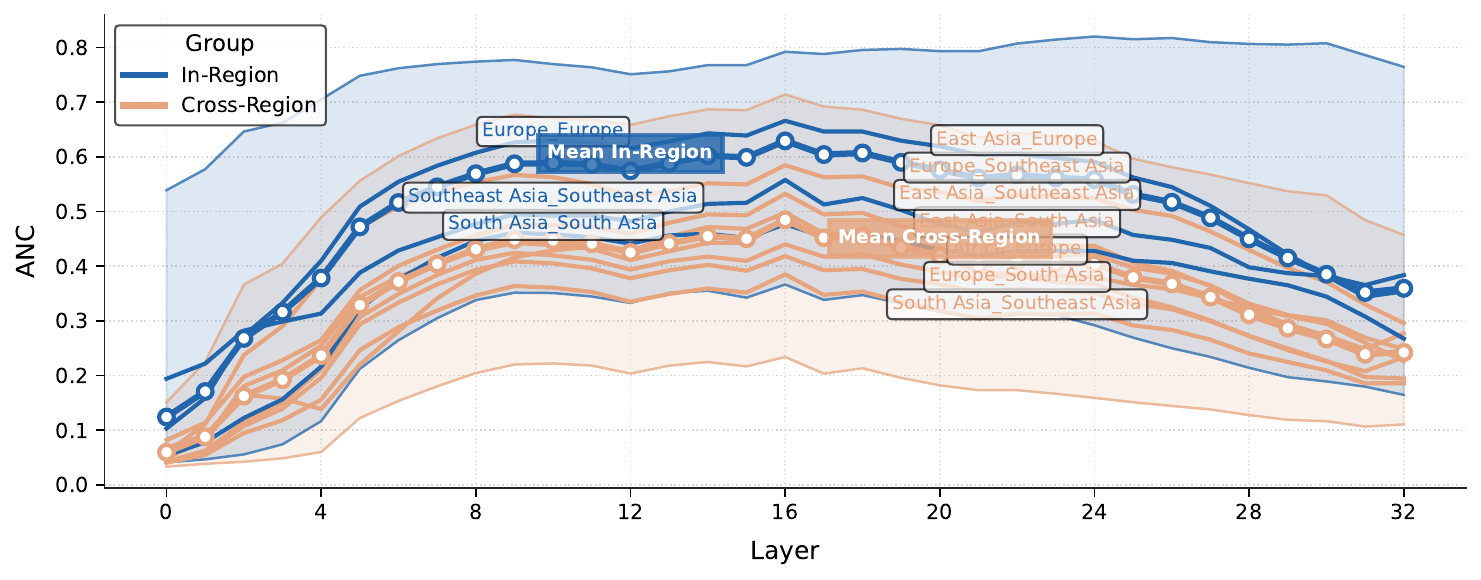}
      \caption{Highlights on pairs w.r.t their linguistic family}
      \vspace{5pt}
  \end{subfigure}
  \caption{Comparisons of per-layer ANC scores on Llama-3.1 (8B) with highlights on pairs w.r.t their resource levels, linguistic region and family. Consistently stronger alignments are observed between HRLs pairs and within-group mean correlations.}
  \label{fig:anc_llama31_complete}
  \vspace{20pt}
\end{figure*}

\begin{figure*}[!t]
  \centering
  \begin{subfigure}[t]{\linewidth}
      \includegraphics[clip, width=\linewidth]{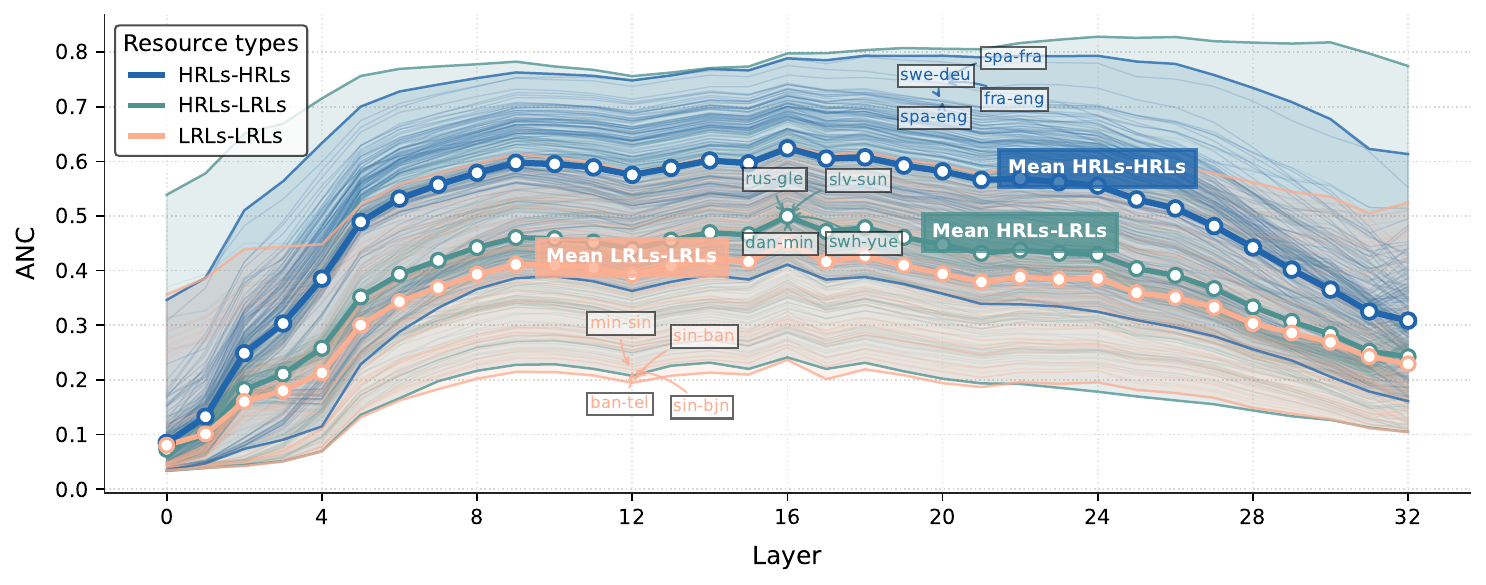}
      \caption{Highlights on pairs w.r.t their resource levels}
      \vspace{5pt}
  \end{subfigure}
  \begin{subfigure}[t]{\linewidth}
      \includegraphics[clip, width=\linewidth]{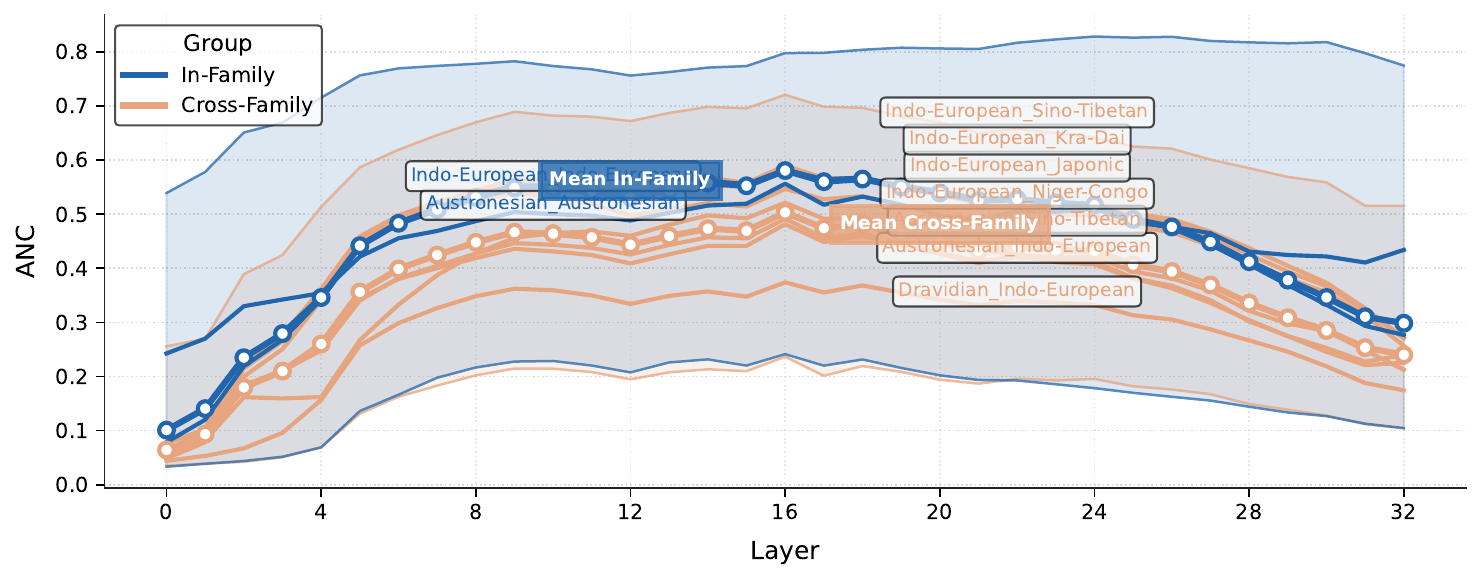}
      \caption{Highlights on pairs w.r.t their linguistic region}
      \vspace{5pt}
  \end{subfigure}
  \begin{subfigure}[t]{\linewidth}
    \centering
      \includegraphics[clip, width=\linewidth]{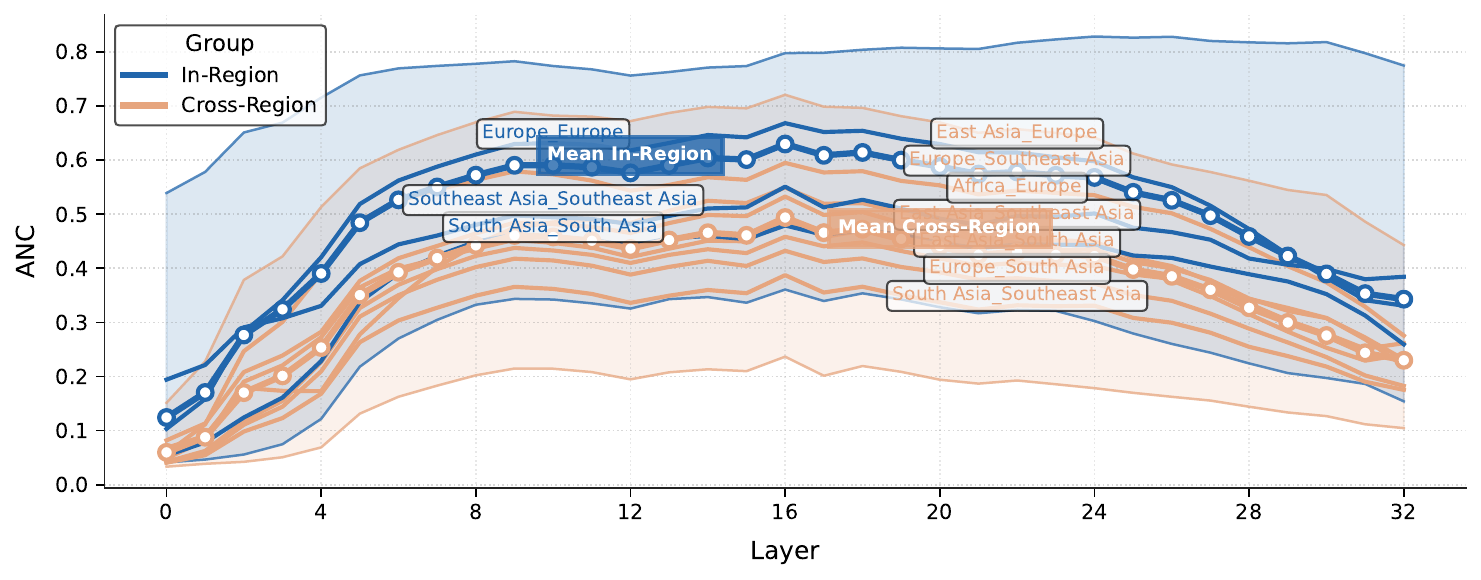}
      \caption{Highlights on pairs w.r.t their linguistic family}
      \vspace{5pt}
  \end{subfigure}
  \caption{Comparisons of per-layer ANC scores on Llama-3.1-Instruct (8B) with highlights on pairs w.r.t their resource levels, linguistic region and family. Consistently stronger alignments are observed between HRLs pairs and within-group mean correlations.}
  \label{fig:anc_llama31instruct_complete}
  \vspace{20pt}
\end{figure*}

\begin{figure*}[!t]
  \centering
  \begin{subfigure}[t]{\linewidth}
      \includegraphics[clip, width=\linewidth]{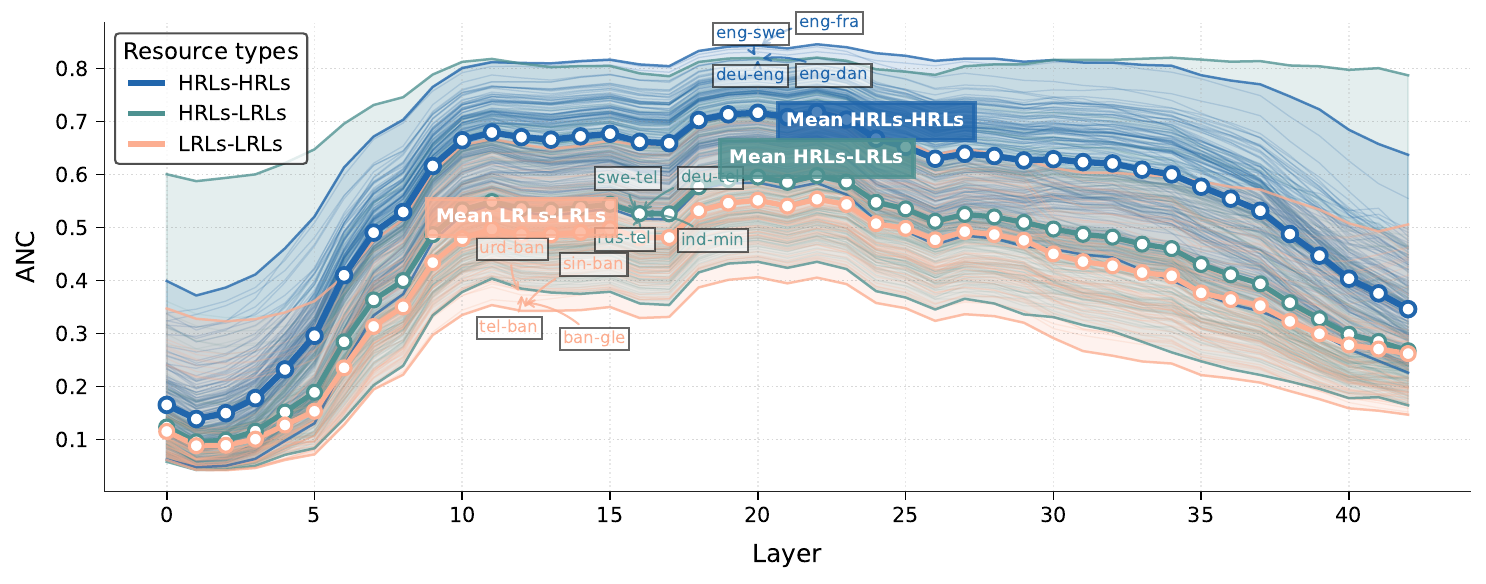}
      \caption{Highlights on pairs w.r.t their resource levels}
      \vspace{5pt}
  \end{subfigure}
  \begin{subfigure}[t]{\linewidth}
      \includegraphics[clip, width=\linewidth]{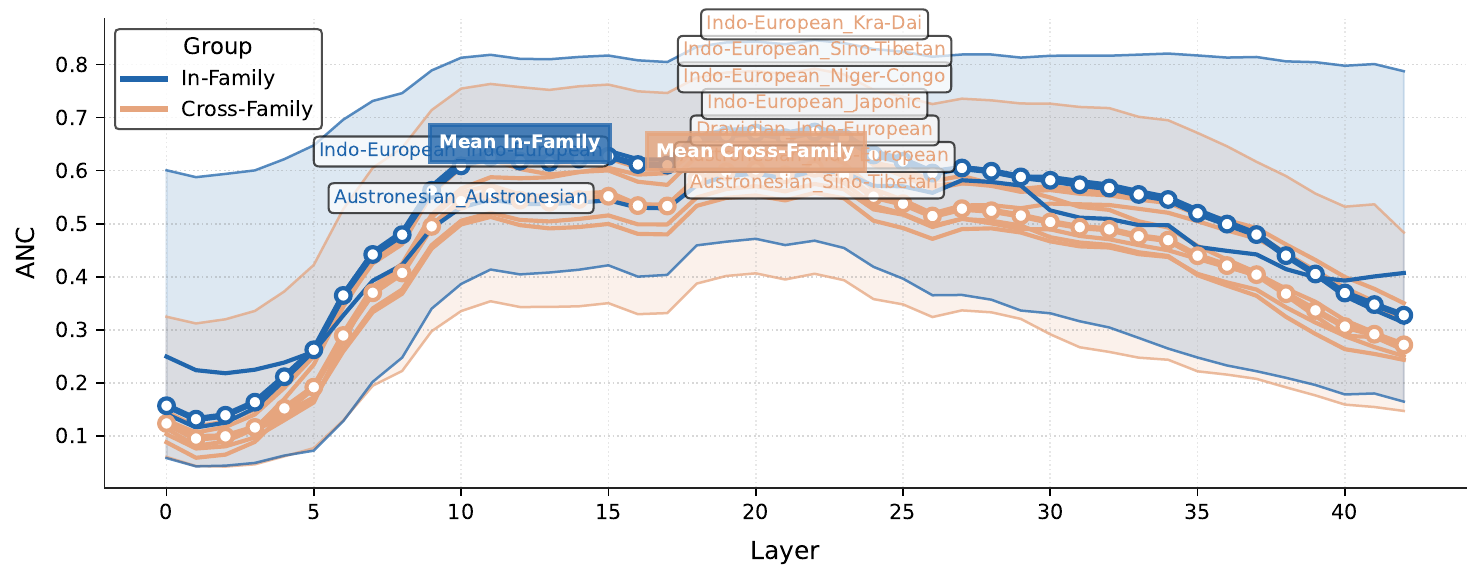}
      \caption{Highlights on pairs w.r.t their linguistic region}
      \vspace{5pt}
  \end{subfigure}
  \begin{subfigure}[t]{\linewidth}
    \centering
      \includegraphics[clip, width=\linewidth]{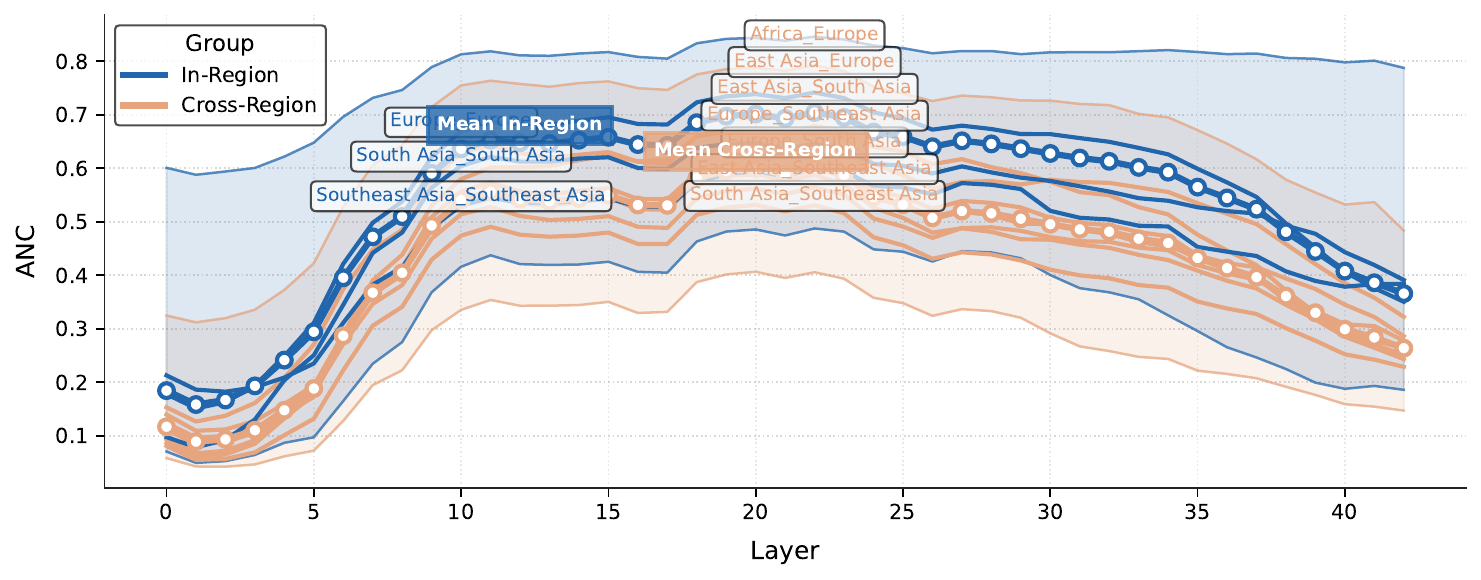}
      \caption{Highlights on pairs w.r.t their linguistic family}
      \vspace{5pt}
  \end{subfigure}
  \caption{Comparisons of per-layer ANC scores on Gemma-2 (9B) with highlights on pairs w.r.t their resource levels, linguistic region and family. Consistently stronger alignments are observed between HRLs pairs and within-group mean correlations.}
  \label{fig:anc_gemma2_complete}
  \vspace{20pt}
\end{figure*}

\begin{figure*}[!t]
  \centering
  \begin{subfigure}[t]{\linewidth}
      \includegraphics[clip, width=\linewidth]{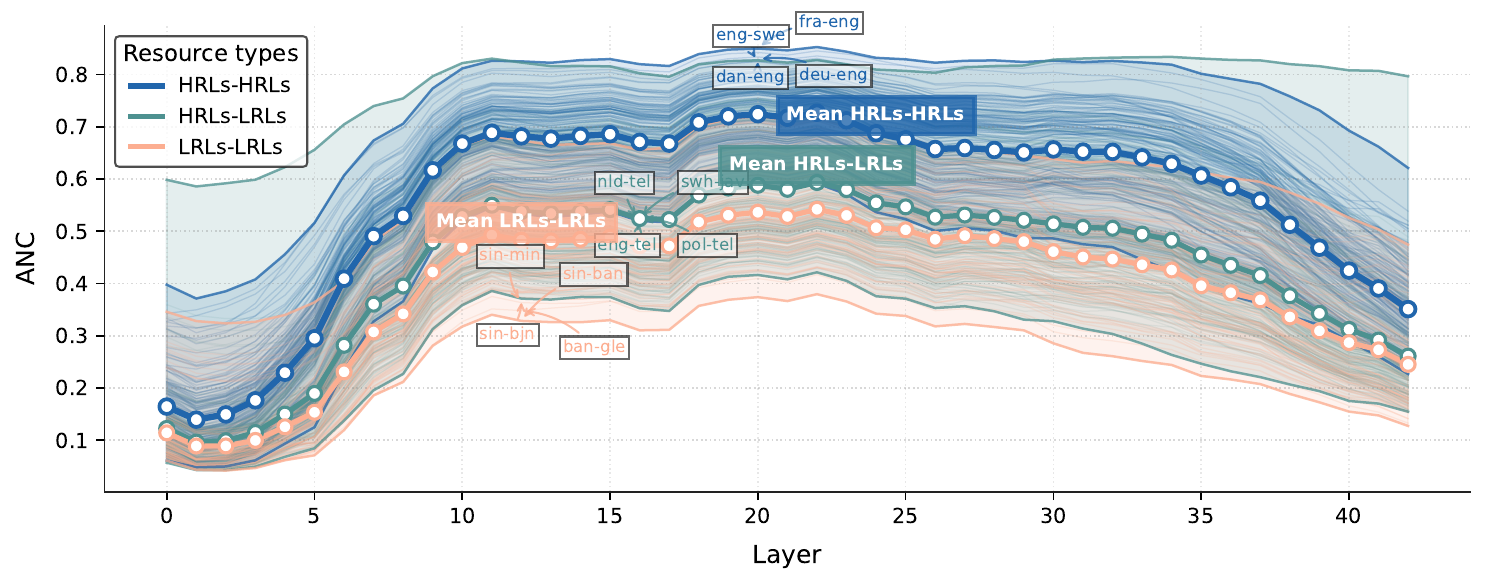}
      \caption{Highlights on pairs w.r.t their resource levels}
      \vspace{5pt}
  \end{subfigure}
  \begin{subfigure}[t]{\linewidth}
      \includegraphics[clip, width=\linewidth]{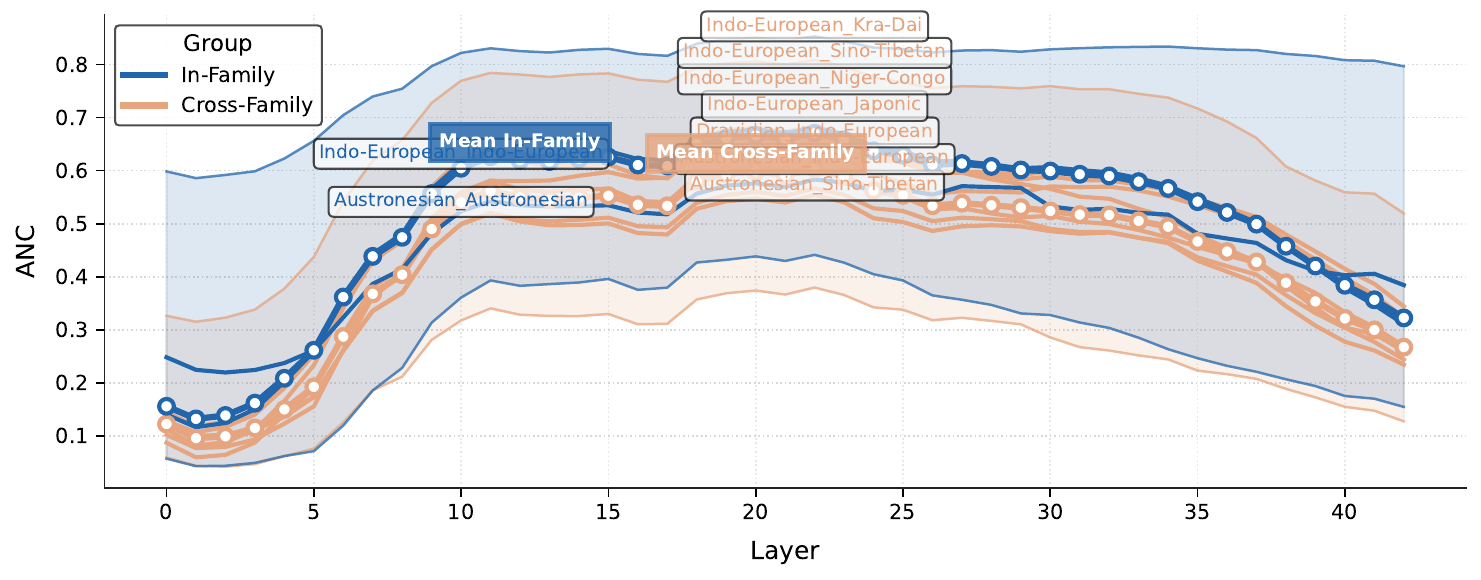}
      \caption{Highlights on pairs w.r.t their linguistic region}
      \vspace{5pt}
  \end{subfigure}
  \begin{subfigure}[t]{\linewidth}
    \centering
      \includegraphics[clip, width=\linewidth]{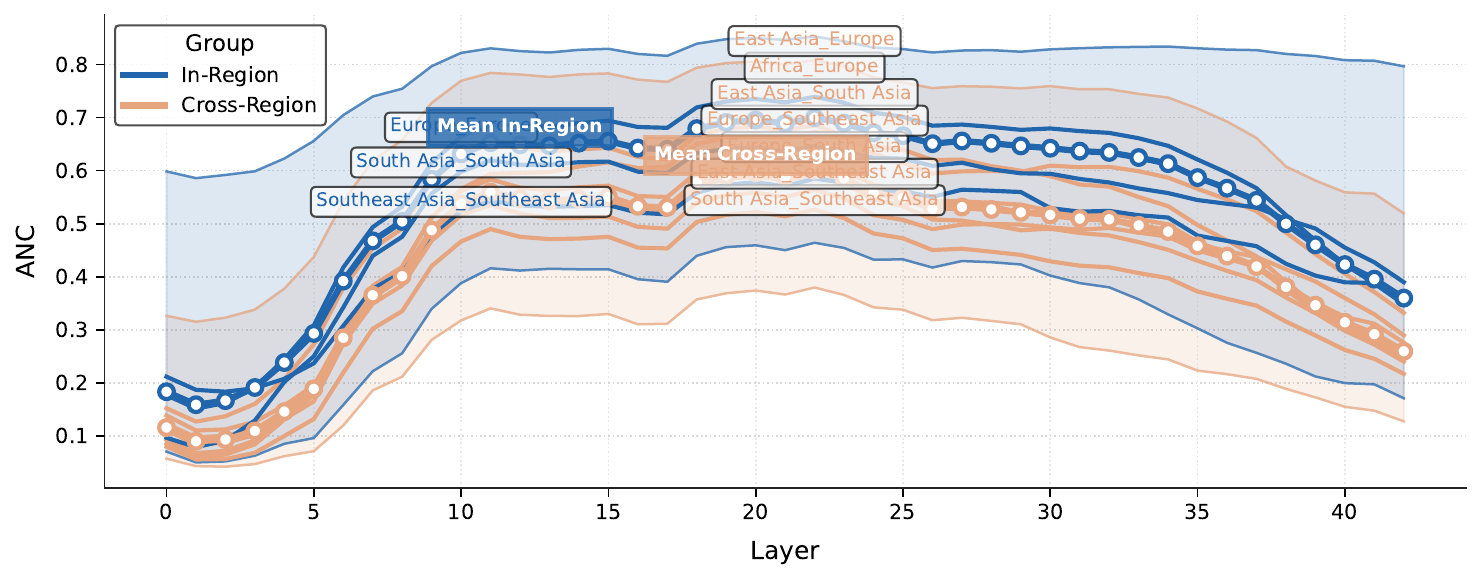}
      \caption{Highlights on pairs w.r.t their linguistic family}
      \vspace{5pt}
  \end{subfigure}
  \caption{Comparisons of per-layer ANC scores on Gemma-2-Instruct (9B) with highlights on pairs w.r.t their resource levels, linguistic region and family. Consistently stronger alignments are observed between HRLs pairs and within-group mean correlations.}
  \label{fig:anc_gemma2instruct_complete}
  \vspace{20pt}
\end{figure*}

\begin{figure*}[!t]
  \centering
  \begin{subfigure}[t]{\linewidth}
      \includegraphics[clip, width=\linewidth]{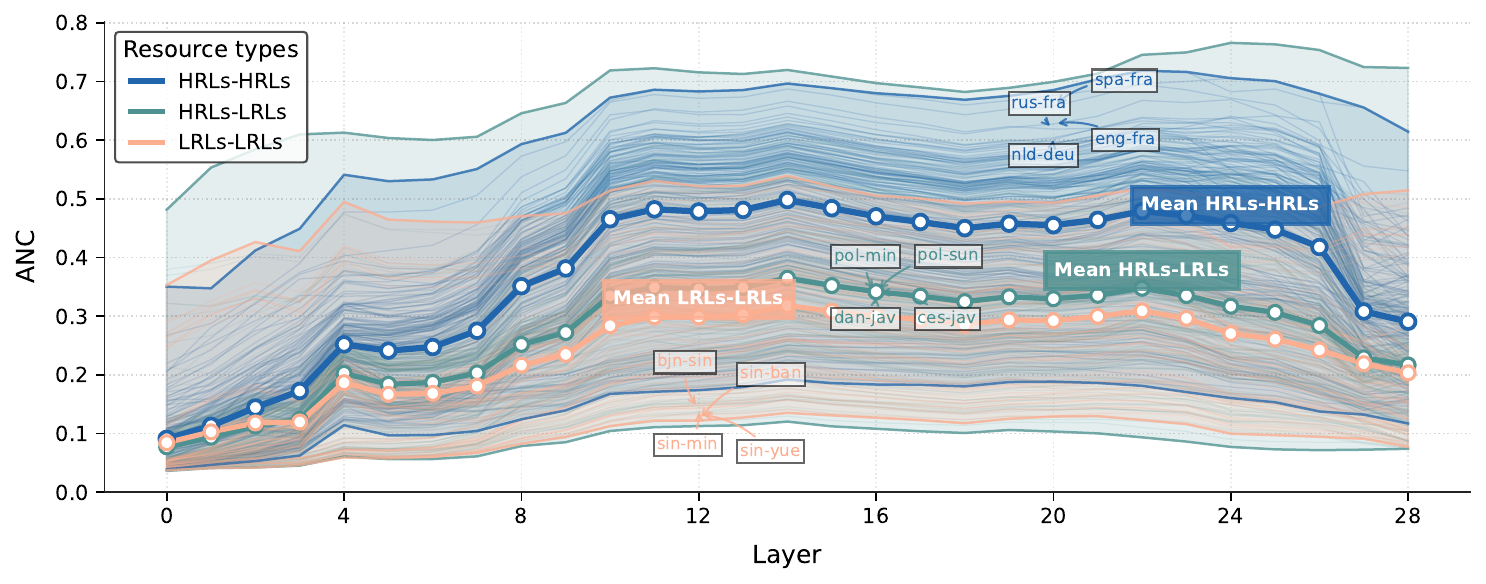}
      \caption{Highlights on pairs w.r.t their resource levels}
      \vspace{5pt}
  \end{subfigure}
  \begin{subfigure}[t]{\linewidth}
      \includegraphics[clip, width=\linewidth]{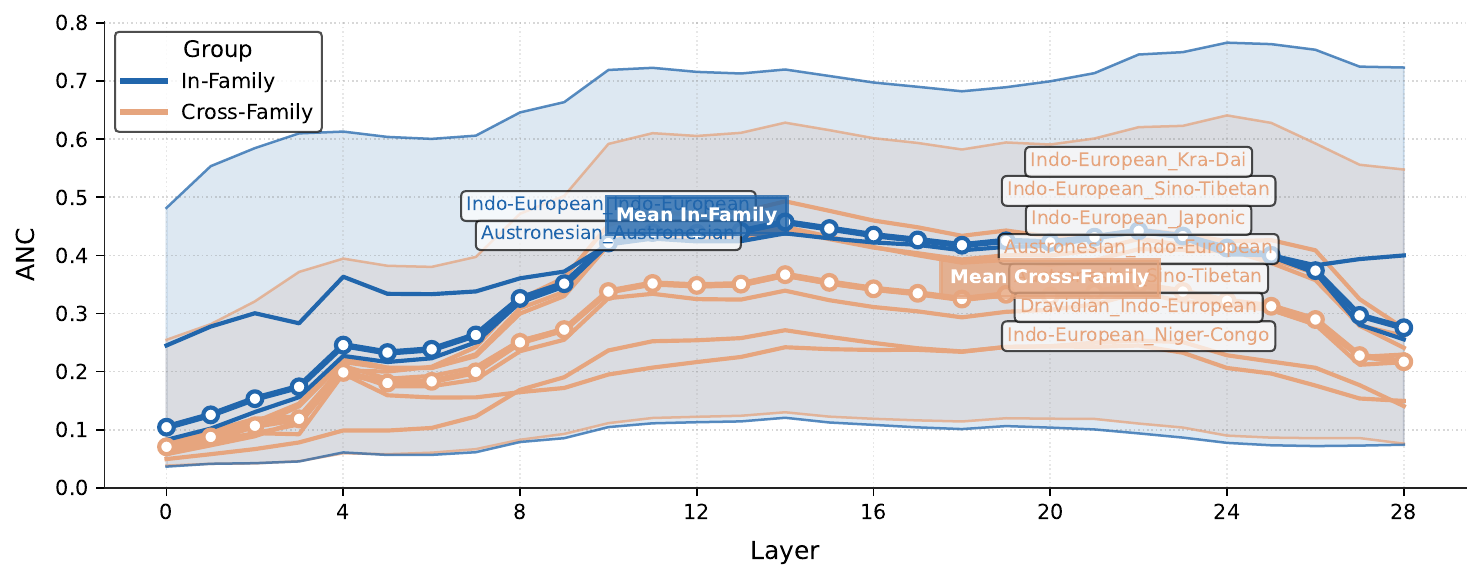}
      \caption{Highlights on pairs w.r.t their linguistic region}
      \vspace{5pt}
  \end{subfigure}
  \begin{subfigure}[t]{\linewidth}
    \centering
      \includegraphics[clip, width=\linewidth]{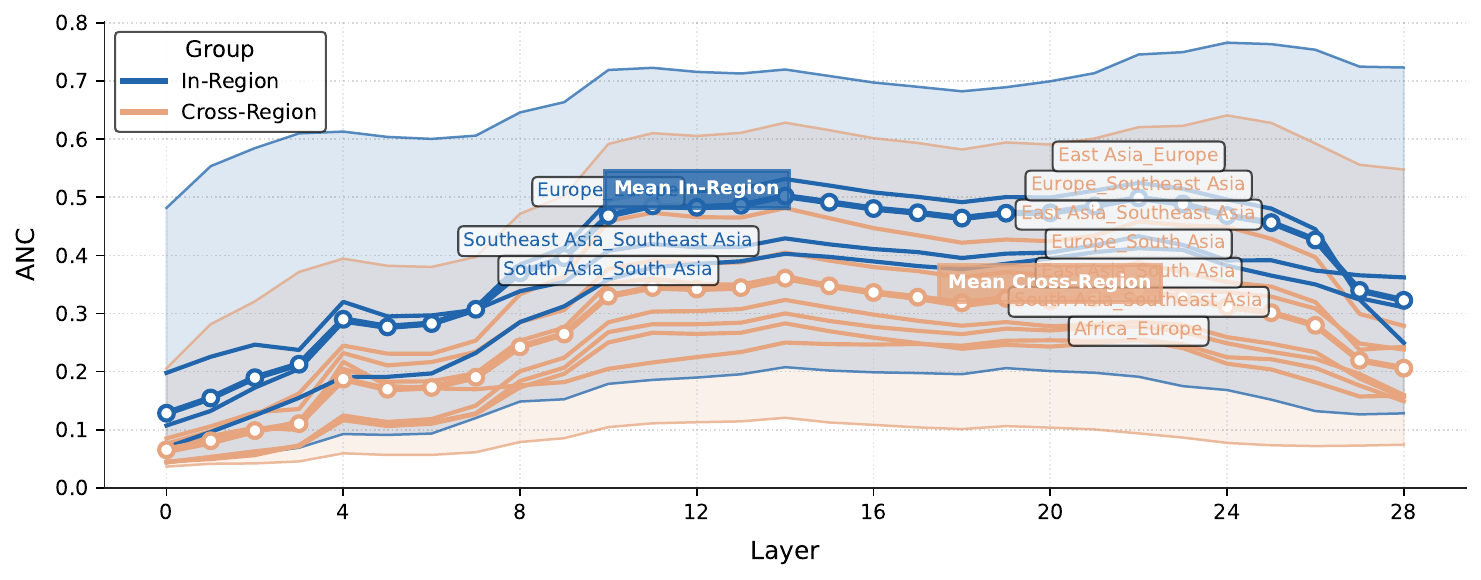}
      \caption{Highlights on pairs w.r.t their linguistic family}
      \vspace{5pt}
  \end{subfigure}
  \caption{Comparisons of per-layer ANC scores on Qwen-2.5 (7B) with highlights on pairs w.r.t their resource levels, linguistic region and family. Consistently stronger alignments are observed between HRLs pairs and within-group mean correlations.}
  \label{fig:anc_qwen_complete}
  \vspace{20pt}
\end{figure*}

\begin{figure}[!t]
  \centering
  \begin{subfigure}[t]{0.85\columnwidth}
      \includegraphics[clip, width=\columnwidth]{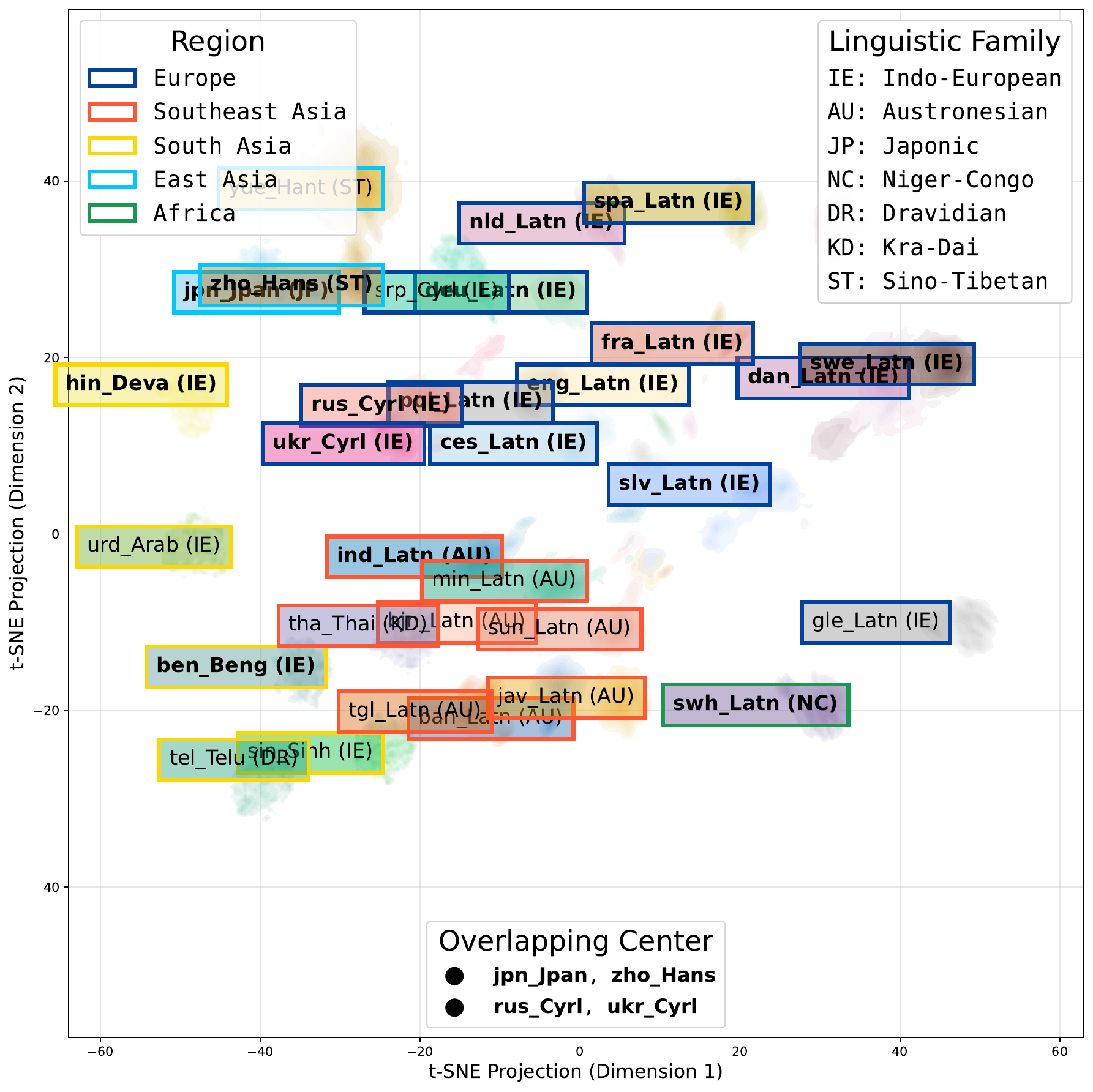}
      \caption{Early (layer 0)}
  \end{subfigure}
  \begin{subfigure}[t]{0.85\columnwidth}
      \includegraphics[clip, width=\columnwidth]{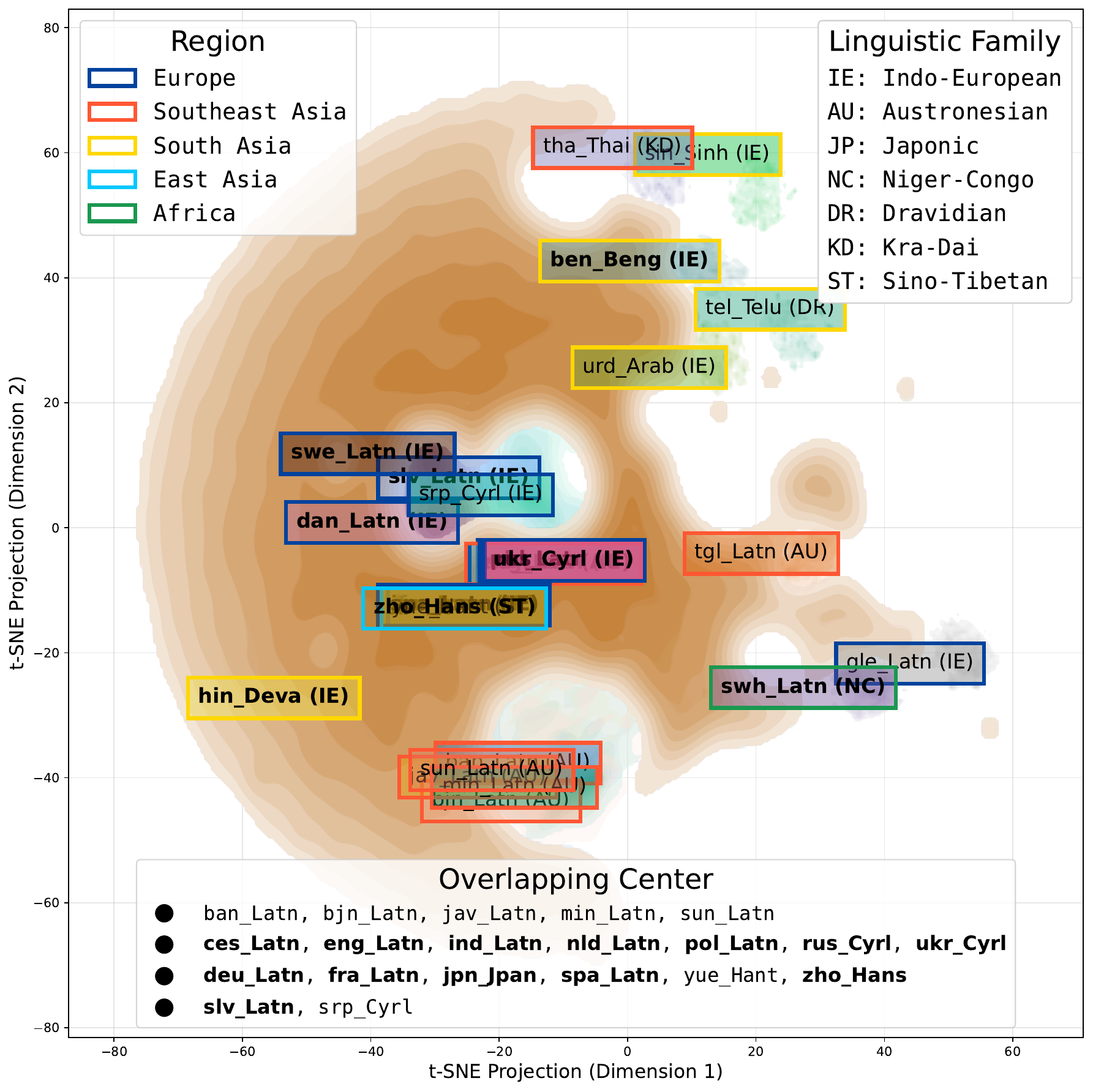}
      \caption{Intermediate (layer 16)}
  \end{subfigure}
  \begin{subfigure}[t]{0.85\columnwidth}
    \centering
      \includegraphics[clip, width=\columnwidth]{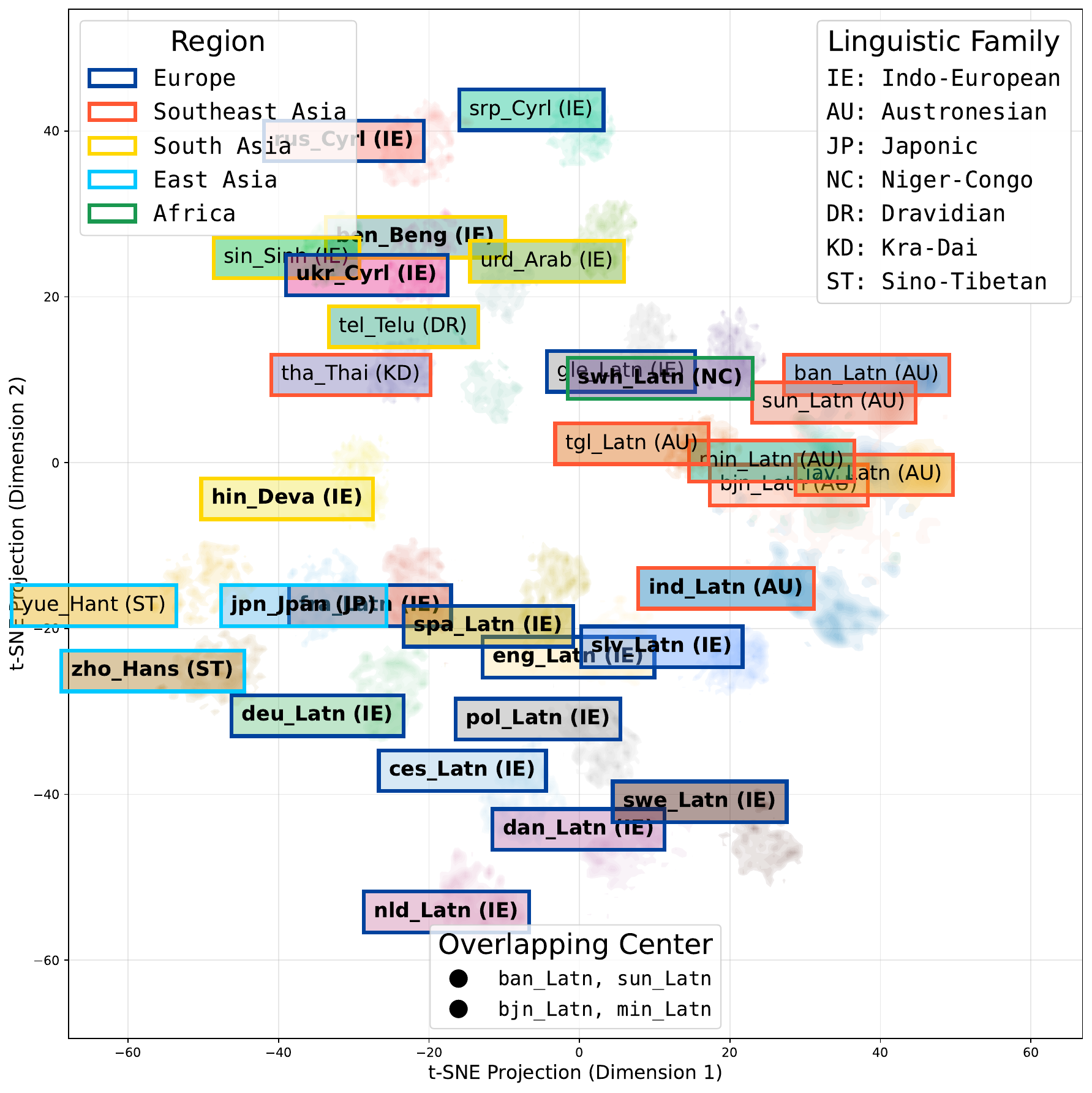}
      \caption{Late (layer 32)}
  \end{subfigure}
  \caption{Hidden-state embeddings of Aya Expanse (8B) projected in t-SNE dimensions, with HRLs in \textbf{bold}. Interlingual overlaps transcending familial and regional boundaries are observed in the intermediate layer representations. In the early and late layers, language representations cluster w.r.t resource levels and linguistic features, with minimal overlap.}
  \label{fig:tsne_aya}
\end{figure}

\begin{figure}[!t]
  \centering
  \begin{subfigure}[t]{0.85\columnwidth}
      \includegraphics[clip, width=\columnwidth]{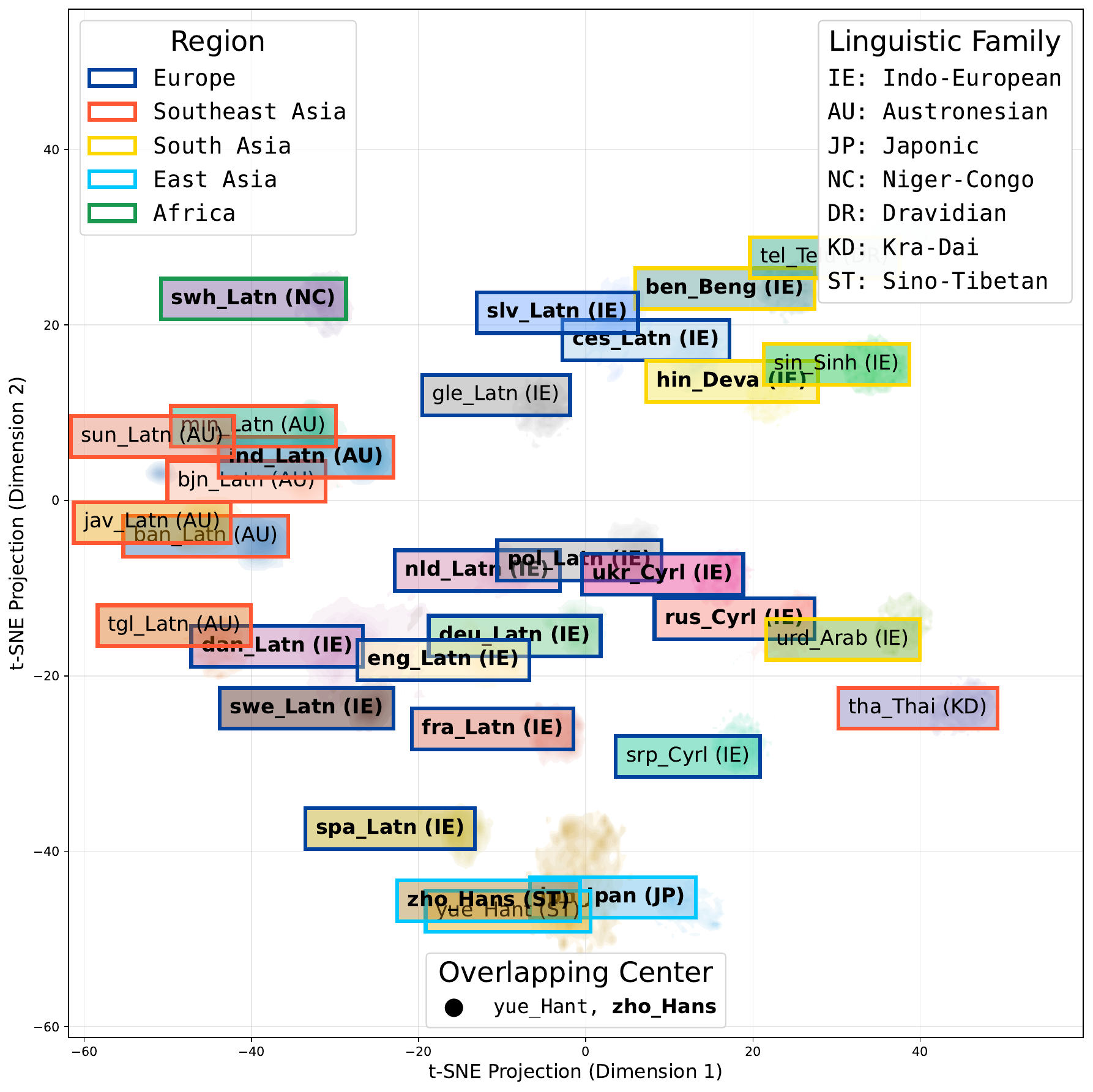}
      \caption{Early (layer 0)}
  \end{subfigure}
  \begin{subfigure}[t]{0.85\columnwidth}
      \includegraphics[clip, width=\columnwidth]{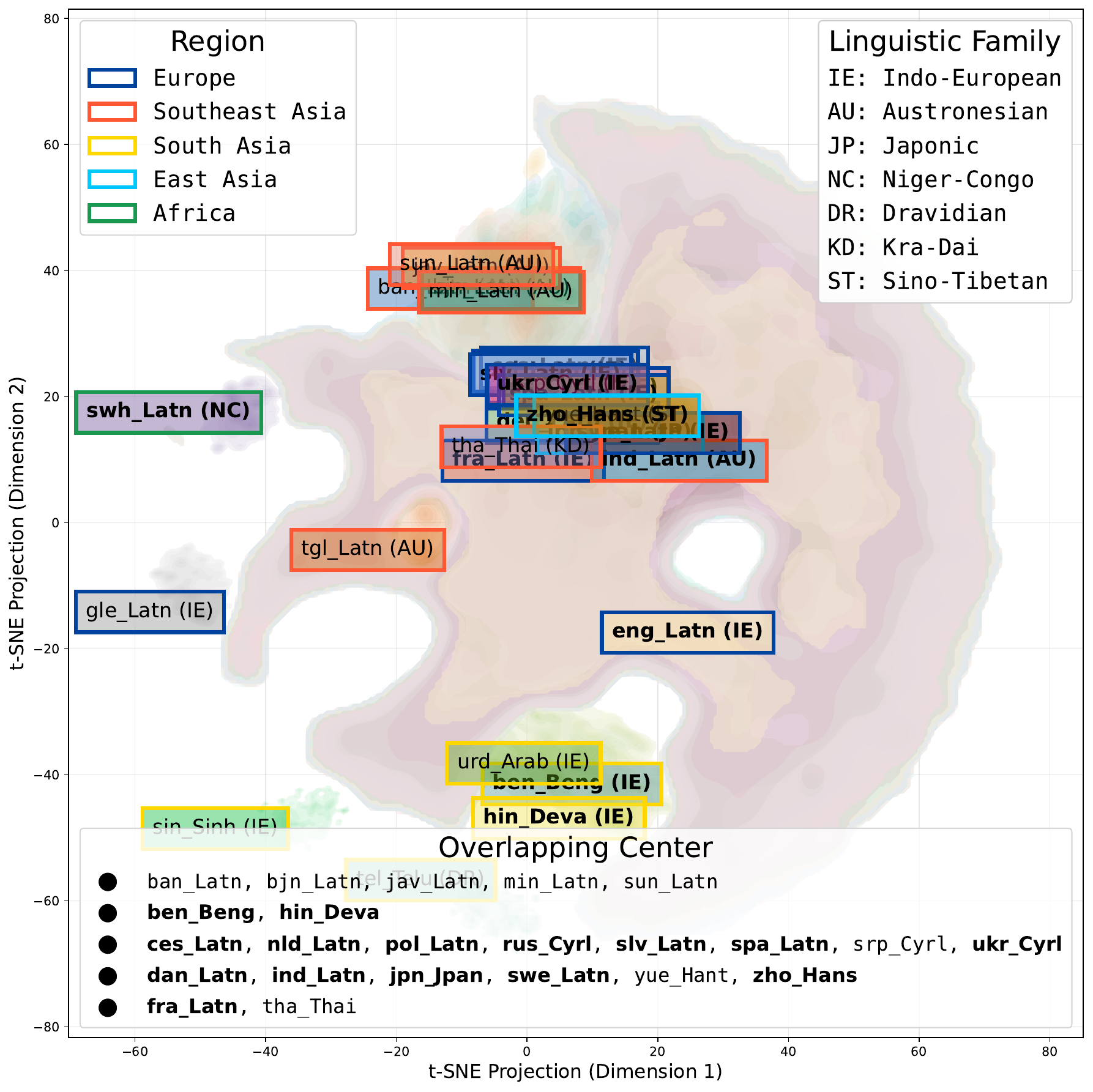}
      \caption{Intermediate (layer 14)}
  \end{subfigure}
  \begin{subfigure}[t]{0.85\columnwidth}
    \centering
      \includegraphics[clip, width=\columnwidth]{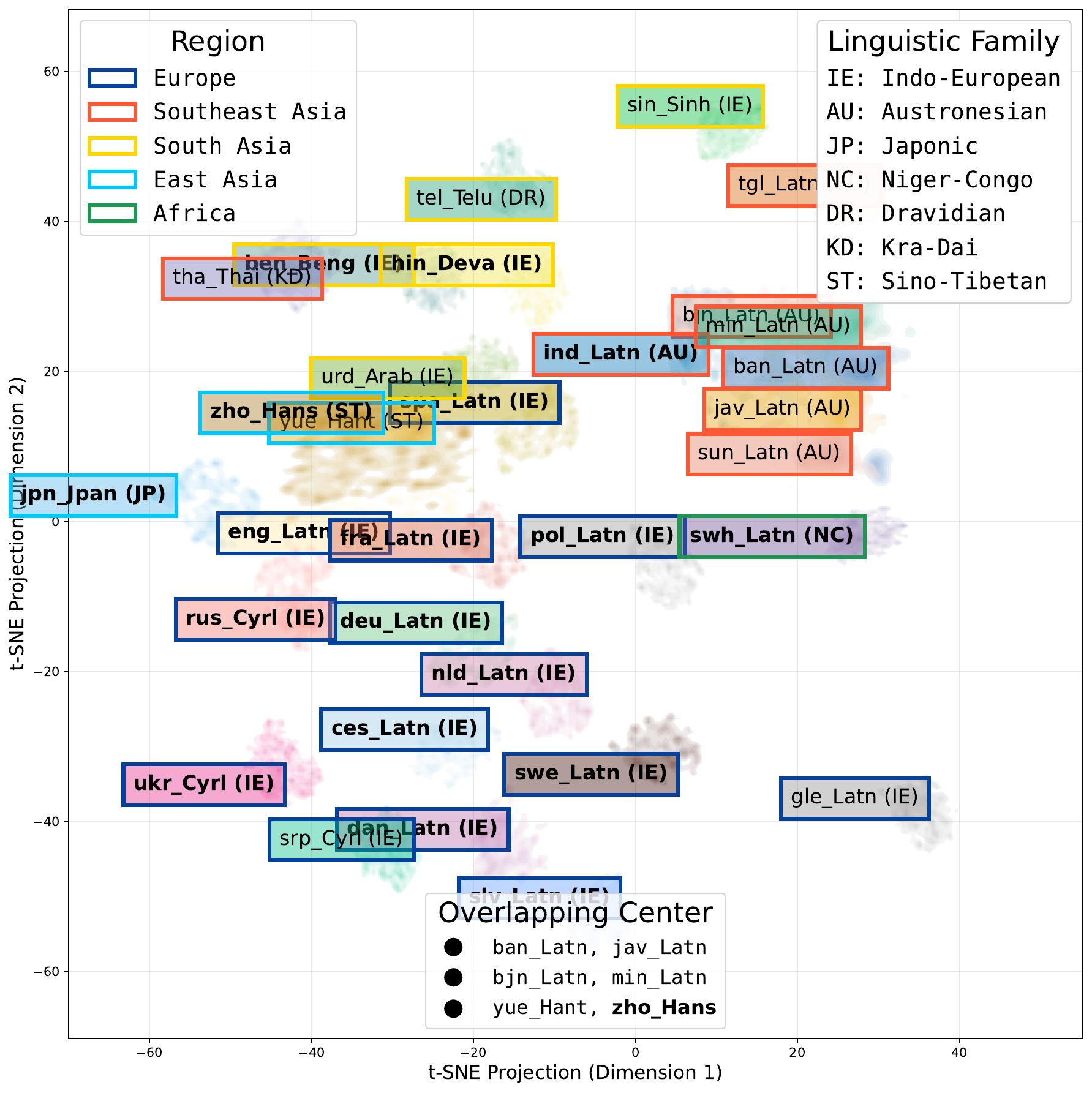}
      \caption{Late (layer 28)}
  \end{subfigure}
  \caption{Hidden-state embeddings of Qwen-2.5 (7B) projected in t-SNE dimensions, with HRLs in \textbf{bold}. Interlingual overlaps transcending familial and regional boundaries are observed in the intermediate layer representations. In the early and late layers, language representations cluster w.r.t resource levels and linguistic features, with minimal overlap.}
  \label{fig:tsne_qwen}
\end{figure}

\begin{figure}[!t]
  \centering
  \begin{subfigure}[t]{0.85\columnwidth}
      \includegraphics[clip, width=\columnwidth]{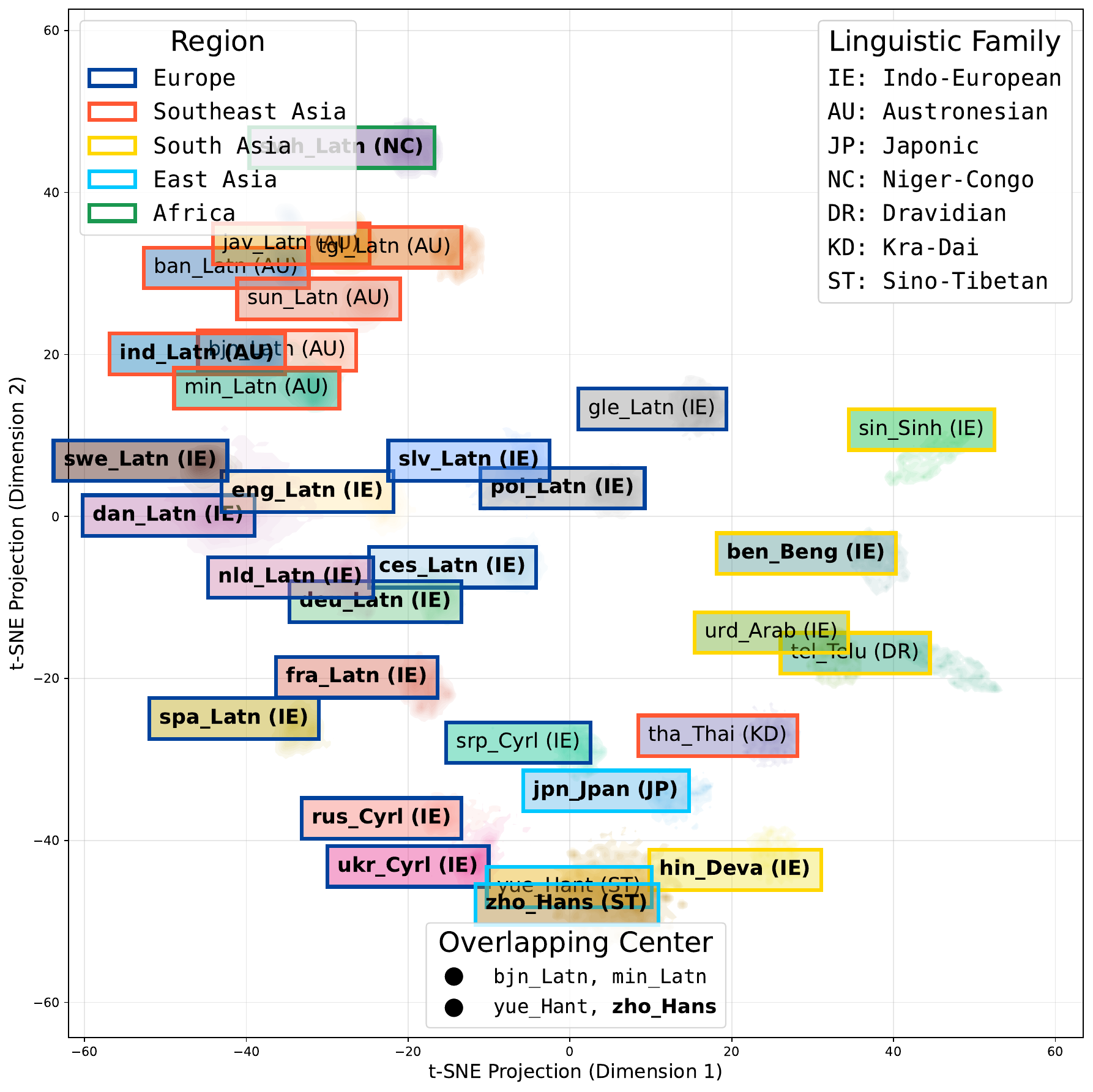}
      \caption{Early (layer 0)}
  \end{subfigure}
  \begin{subfigure}[t]{0.85\columnwidth}
      \includegraphics[clip, width=\columnwidth]{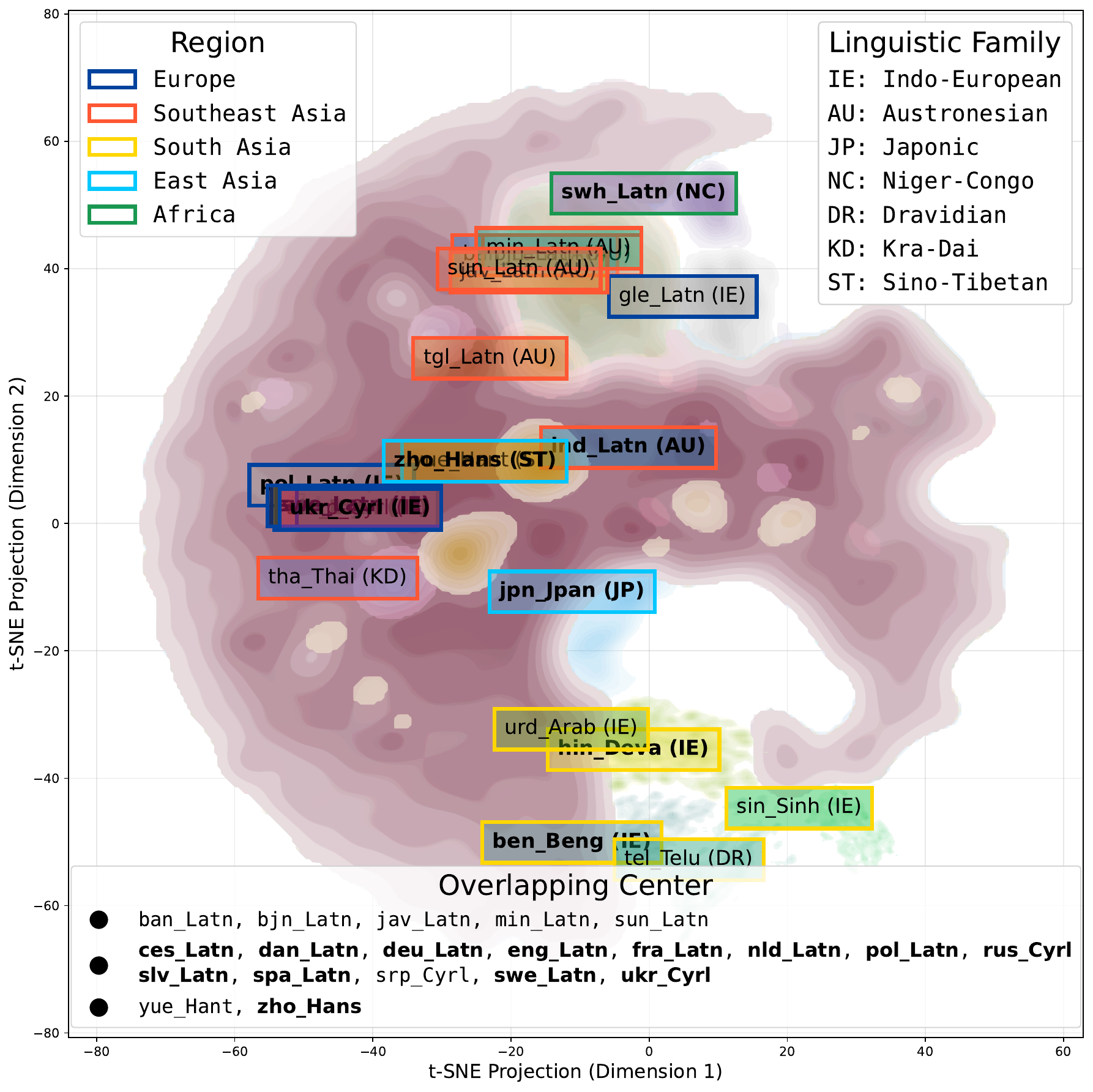}
      \caption{Intermediate (layer 16)}
  \end{subfigure}
  \begin{subfigure}[t]{0.85\columnwidth}
    \centering
      \includegraphics[clip, width=\columnwidth]{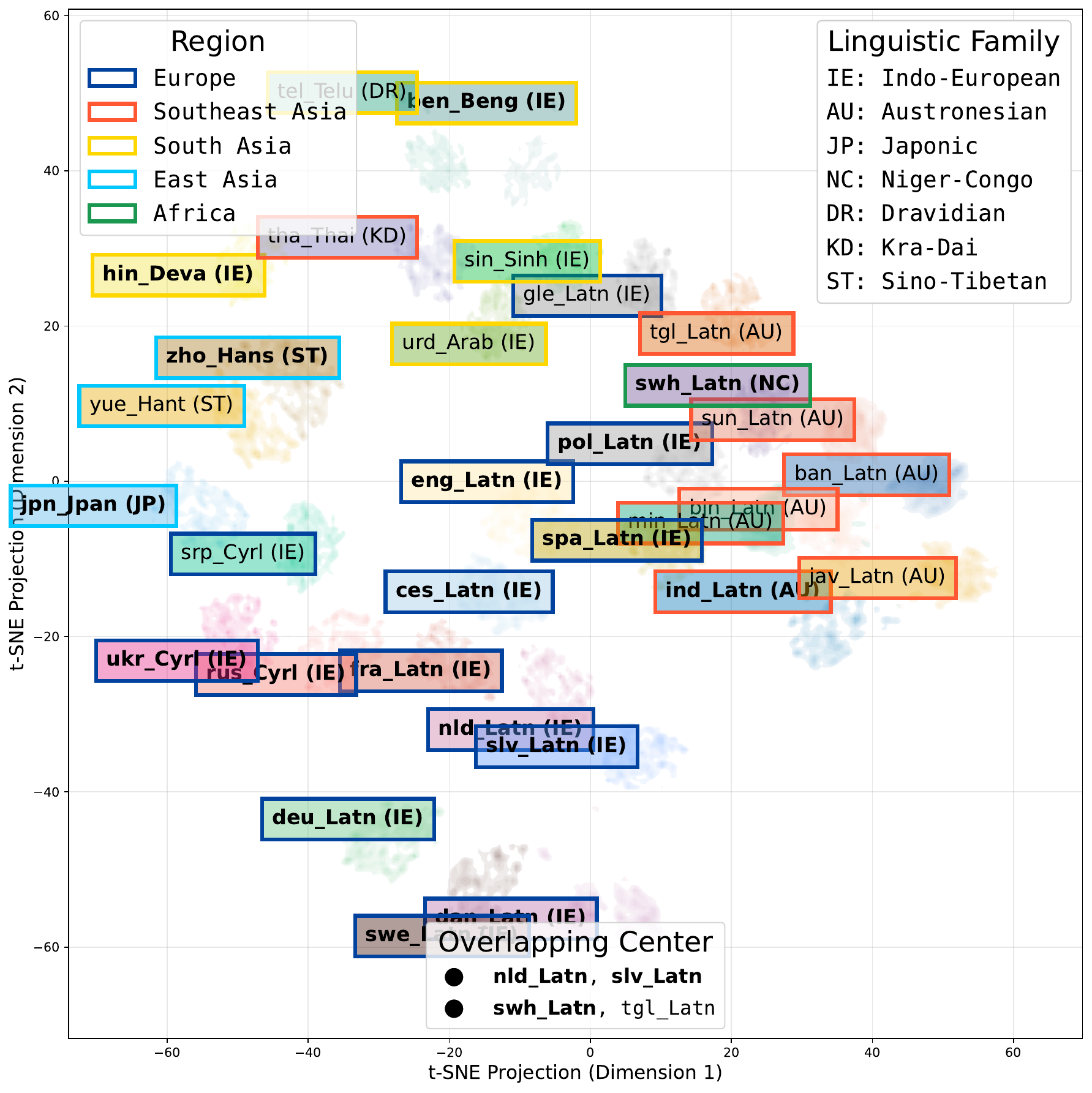}
      \caption{Late (layer 32)}
  \end{subfigure}
  \caption{Hidden-state embeddings of Llama-3.1 (8B) projected in t-SNE dimensions, with HRLs in \textbf{bold}. Interlingual overlaps transcending familial and regional boundaries are observed in the intermediate layer representations. In the early and late layers, language representations cluster w.r.t resource levels and linguistic features, with minimal overlap.}
  \label{fig:tsne_llama31}
\end{figure}

\begin{figure}[!t]
  \centering
  \begin{subfigure}[t]{0.85\columnwidth}
      \includegraphics[clip, width=\columnwidth]{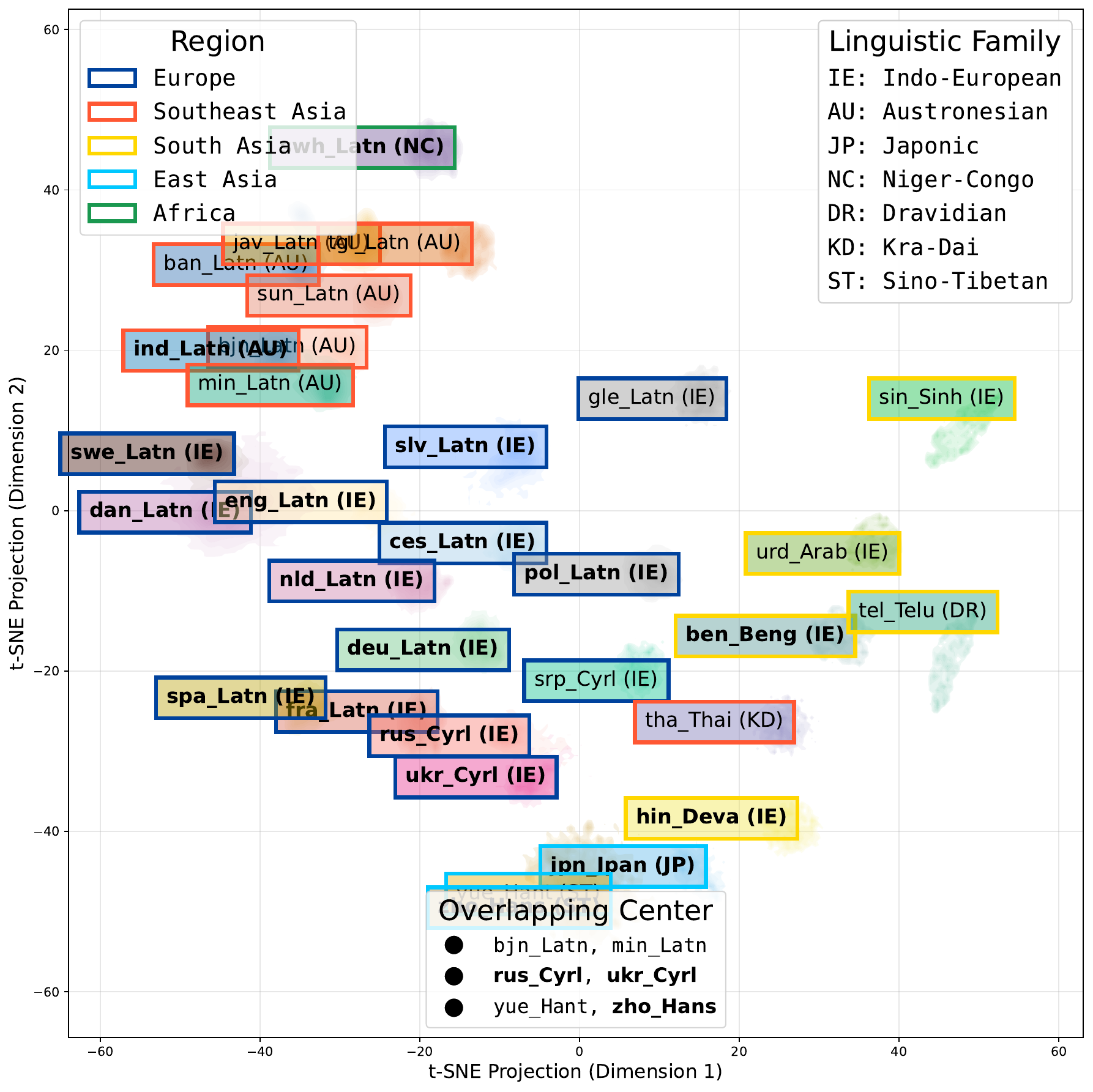}
      \caption{Early (layer 0)}
  \end{subfigure}
  \begin{subfigure}[t]{0.85\columnwidth}
      \includegraphics[clip, width=\columnwidth]{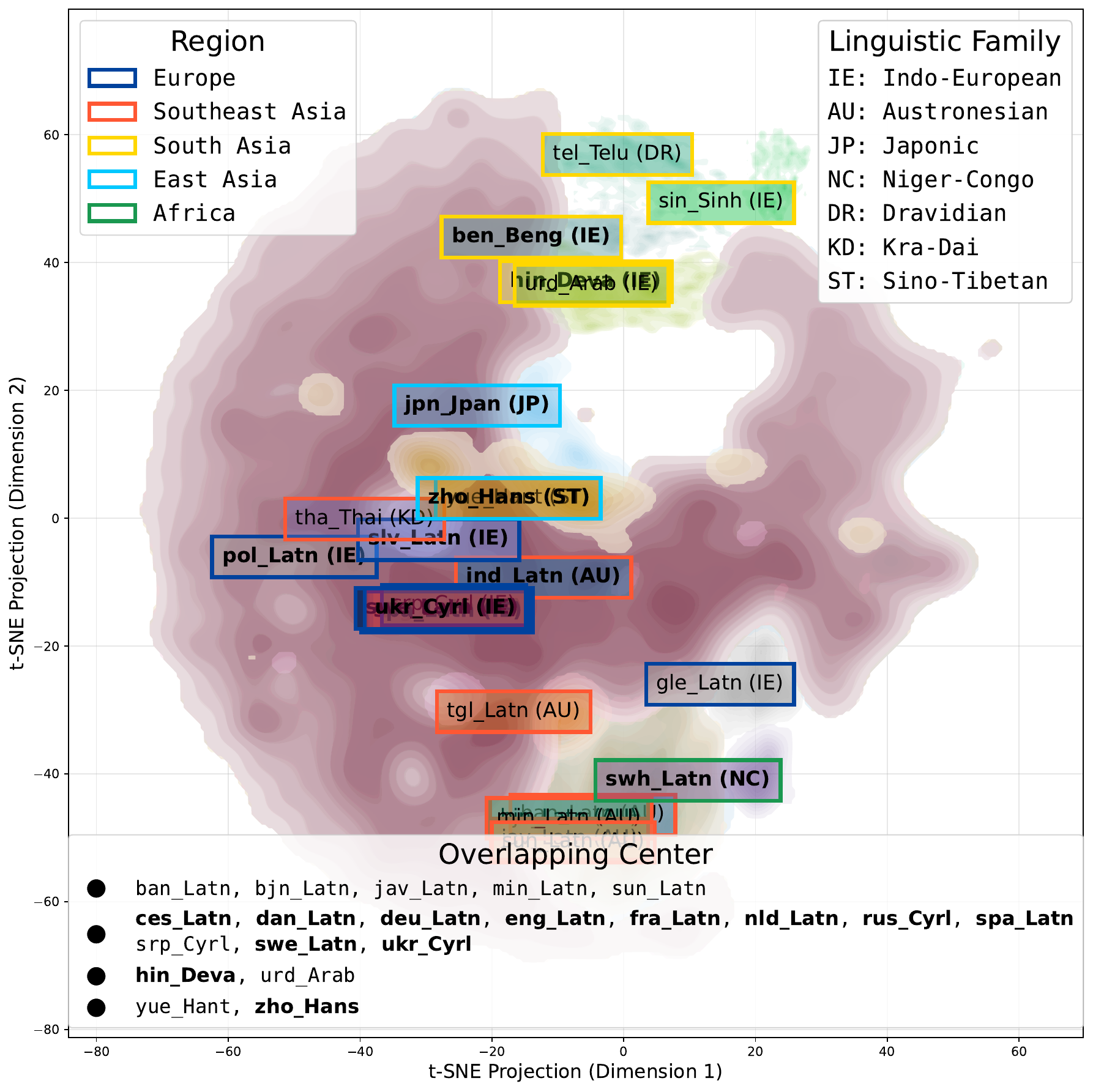}
      \caption{Intermediate (layer 16)}
  \end{subfigure}
  \begin{subfigure}[t]{0.85\columnwidth}
    \centering
      \includegraphics[clip, width=\columnwidth]{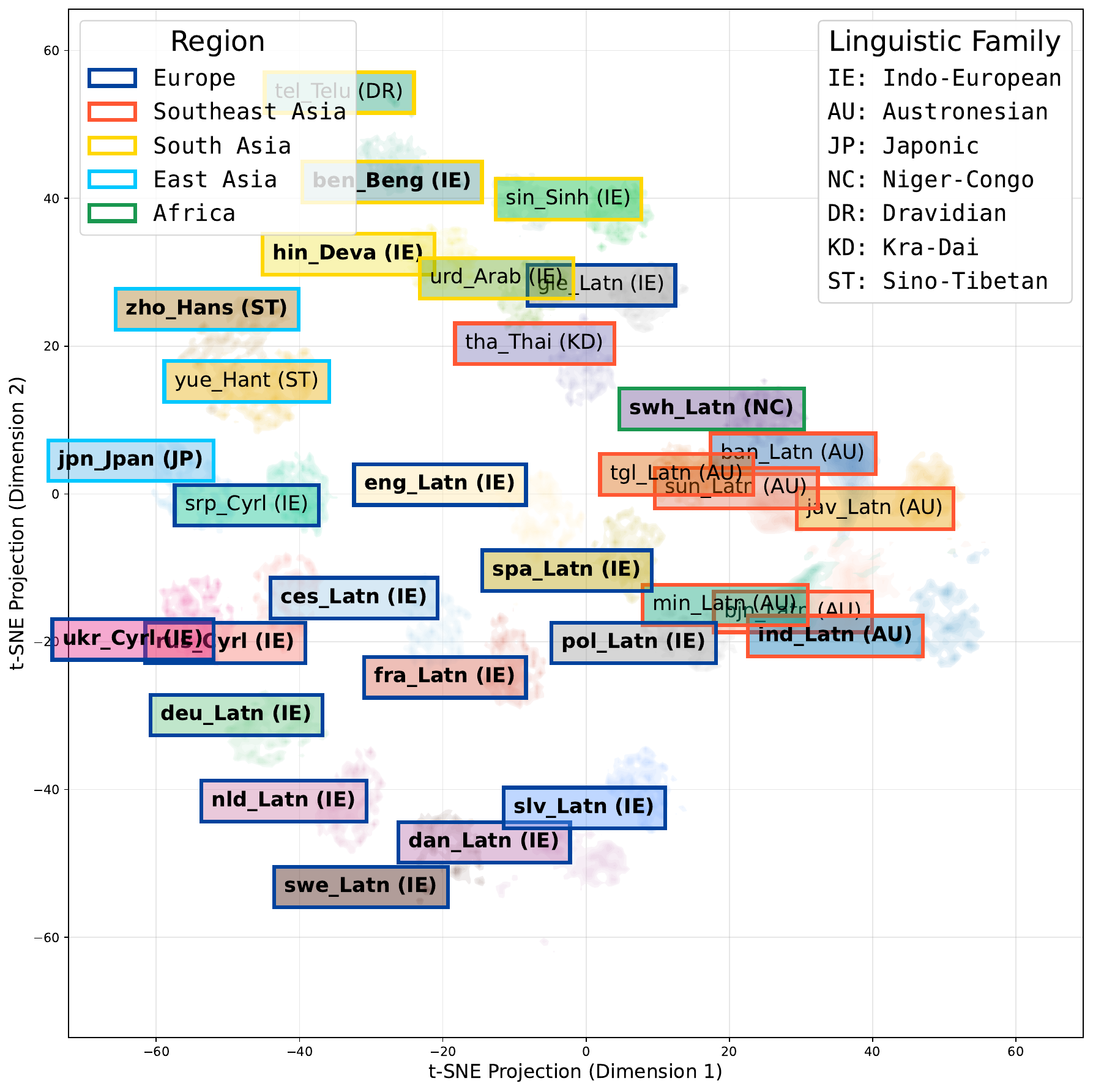}
      \caption{Late (layer 32)}
  \end{subfigure}
  \caption{Hidden-state embeddings of Llama-3.1-Instruct (8B) projected in t-SNE dimensions, with HRLs in \textbf{bold}. Interlingual overlaps transcending familial and regional boundaries are observed in the intermediate layer representations. In the early and late layers, language representations cluster w.r.t resource levels and linguistic features, with minimal overlap.}
  \label{fig:tsne_llama31instruct}
\end{figure}

\begin{figure}[!t]
  \centering
  \begin{subfigure}[t]{0.85\columnwidth}
      \includegraphics[clip, width=\columnwidth]{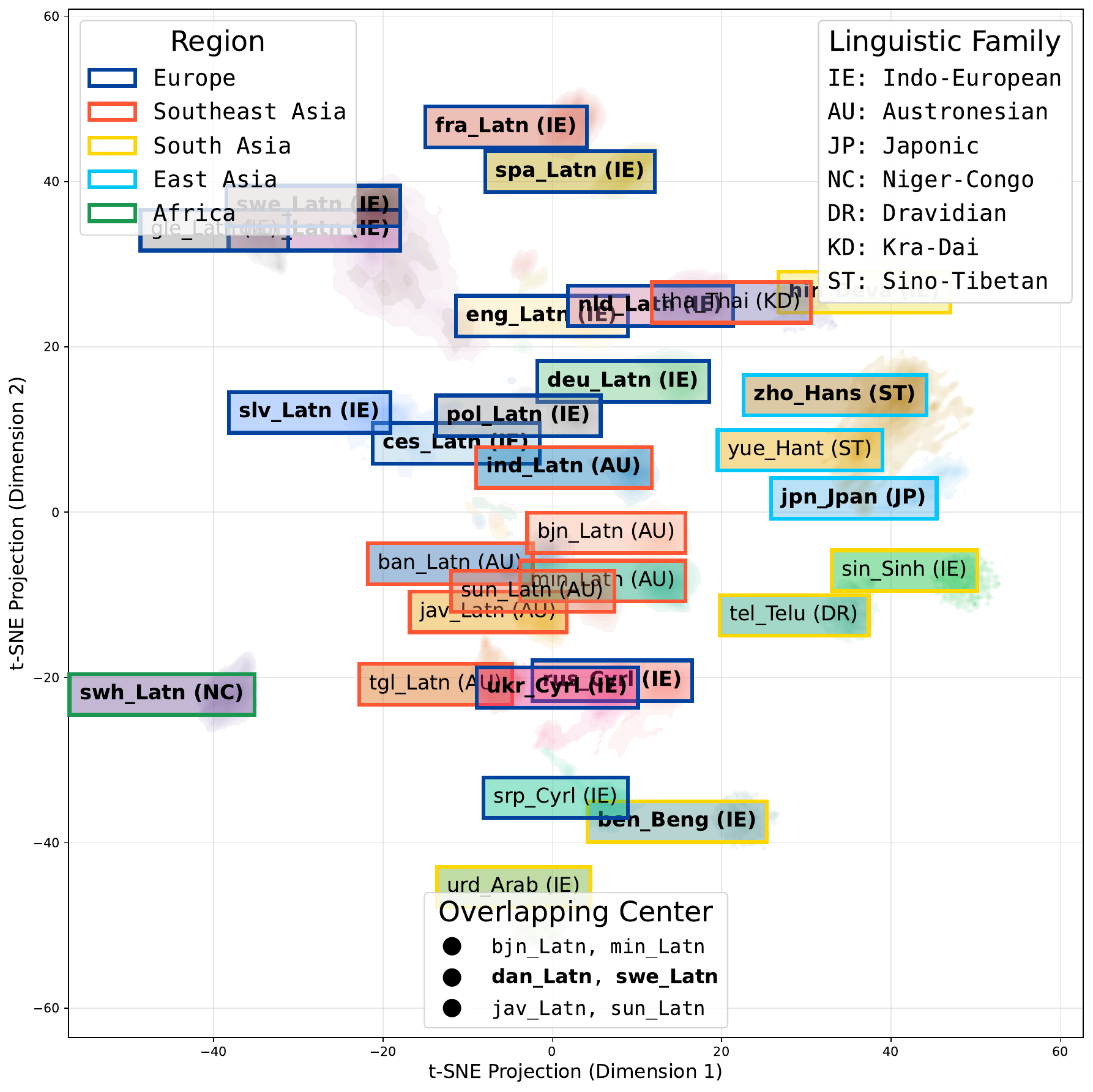}
      \caption{Early (layer 0)}
  \end{subfigure}
  \begin{subfigure}[t]{0.85\columnwidth}
      \includegraphics[clip, width=\columnwidth]{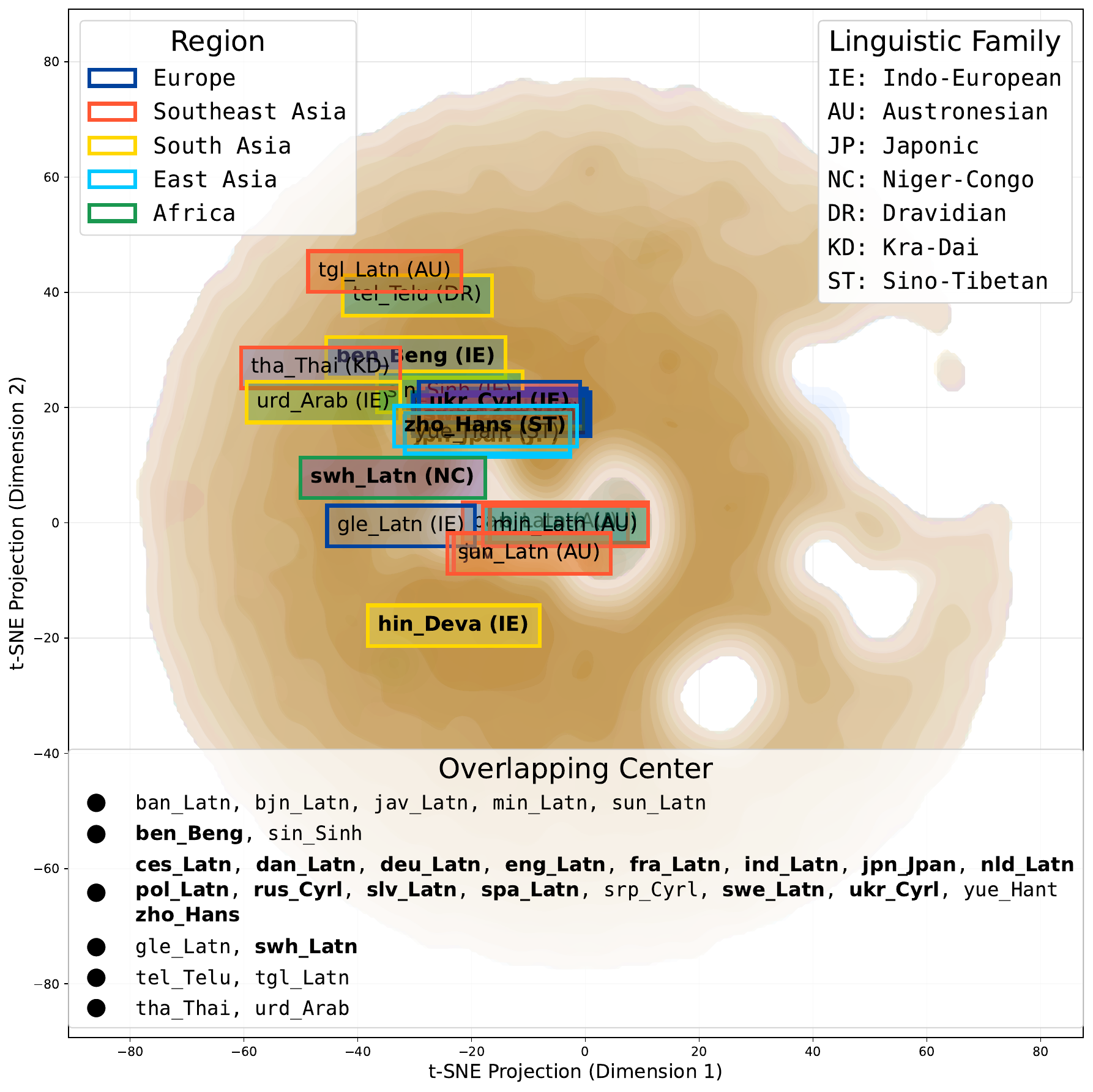}
      \caption{Intermediate (layer 21)}
  \end{subfigure}
  \begin{subfigure}[t]{0.85\columnwidth}
    \centering
      \includegraphics[clip, width=\columnwidth]{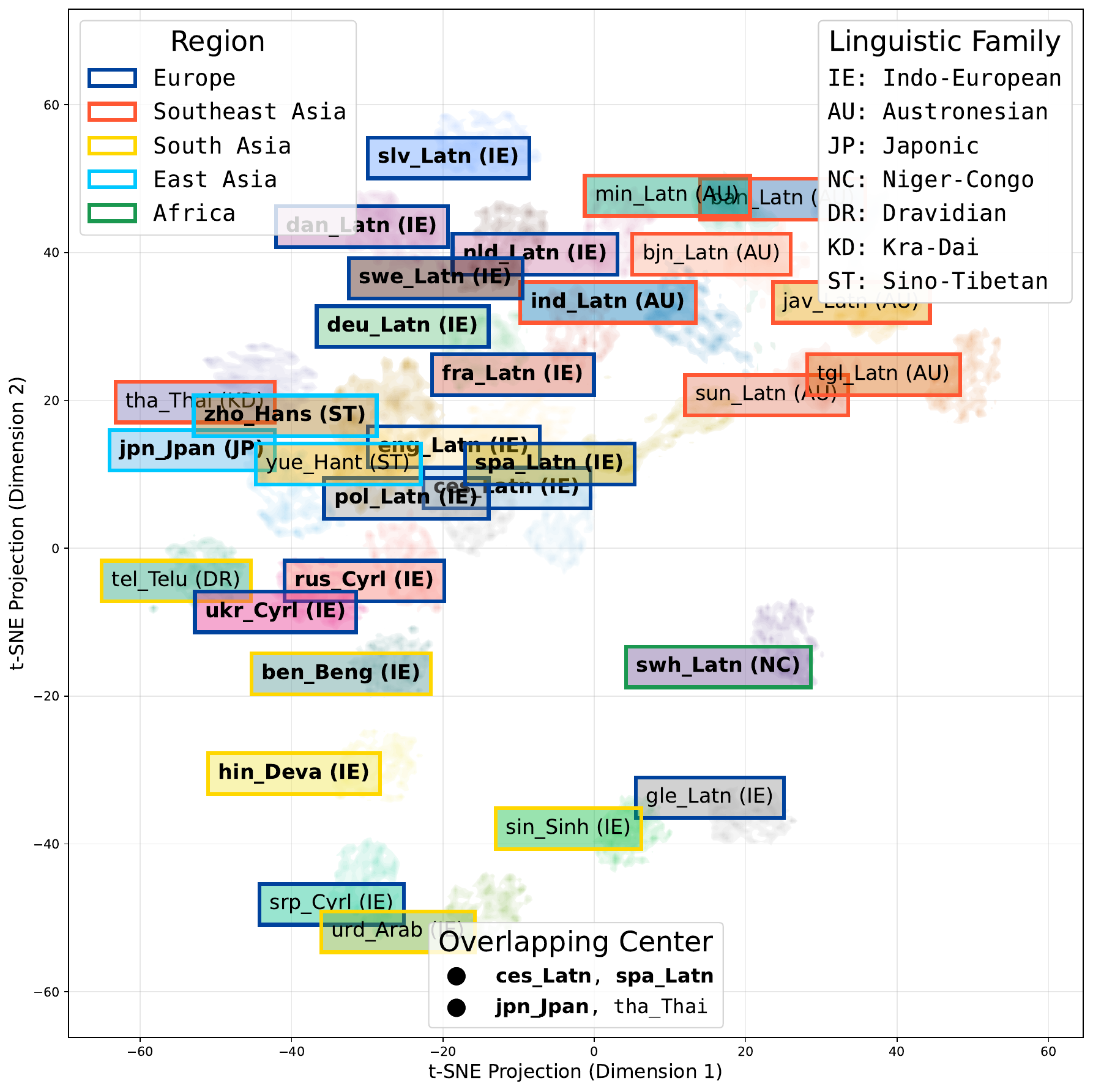}
      \caption{Late (layer 42)}
  \end{subfigure}
  \caption{Hidden-state embeddings of Gemma-2 (9B) projected in t-SNE dimensions, with HRLs in \textbf{bold}. Interlingual overlaps transcending familial and regional boundaries are observed in the intermediate layer representations. In the early and late layers, language representations cluster w.r.t resource levels and linguistic features, with minimal overlap.}
  \label{fig:tsne_gemma2}
\end{figure}

\begin{figure}[]
  \centering
  \begin{subfigure}[t]{0.85\columnwidth}
      \includegraphics[clip, width=\columnwidth]{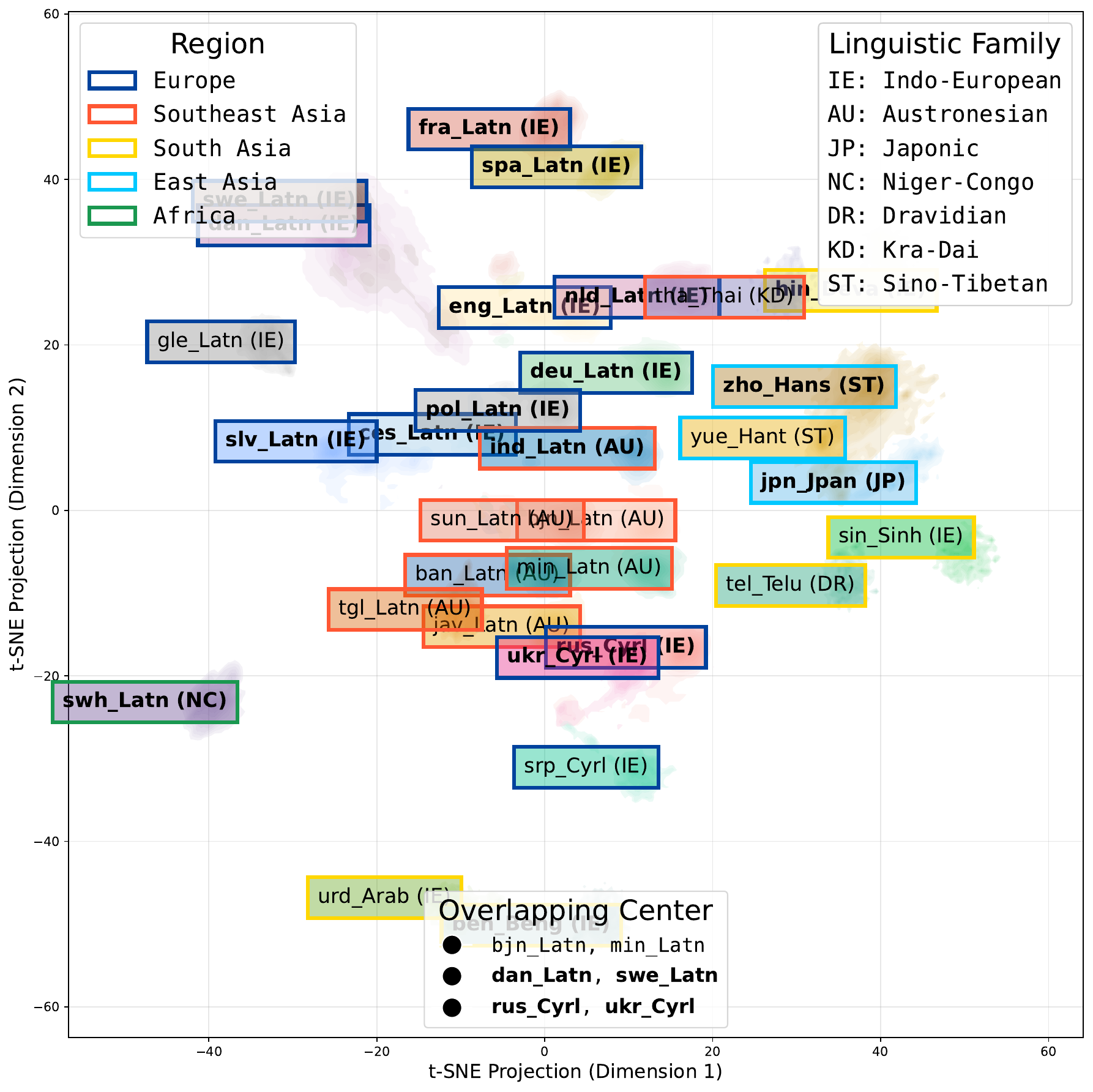}
      \caption{Early (layer 0)}
  \end{subfigure}
  \begin{subfigure}[t]{0.85\columnwidth}
      \includegraphics[clip, width=\columnwidth]{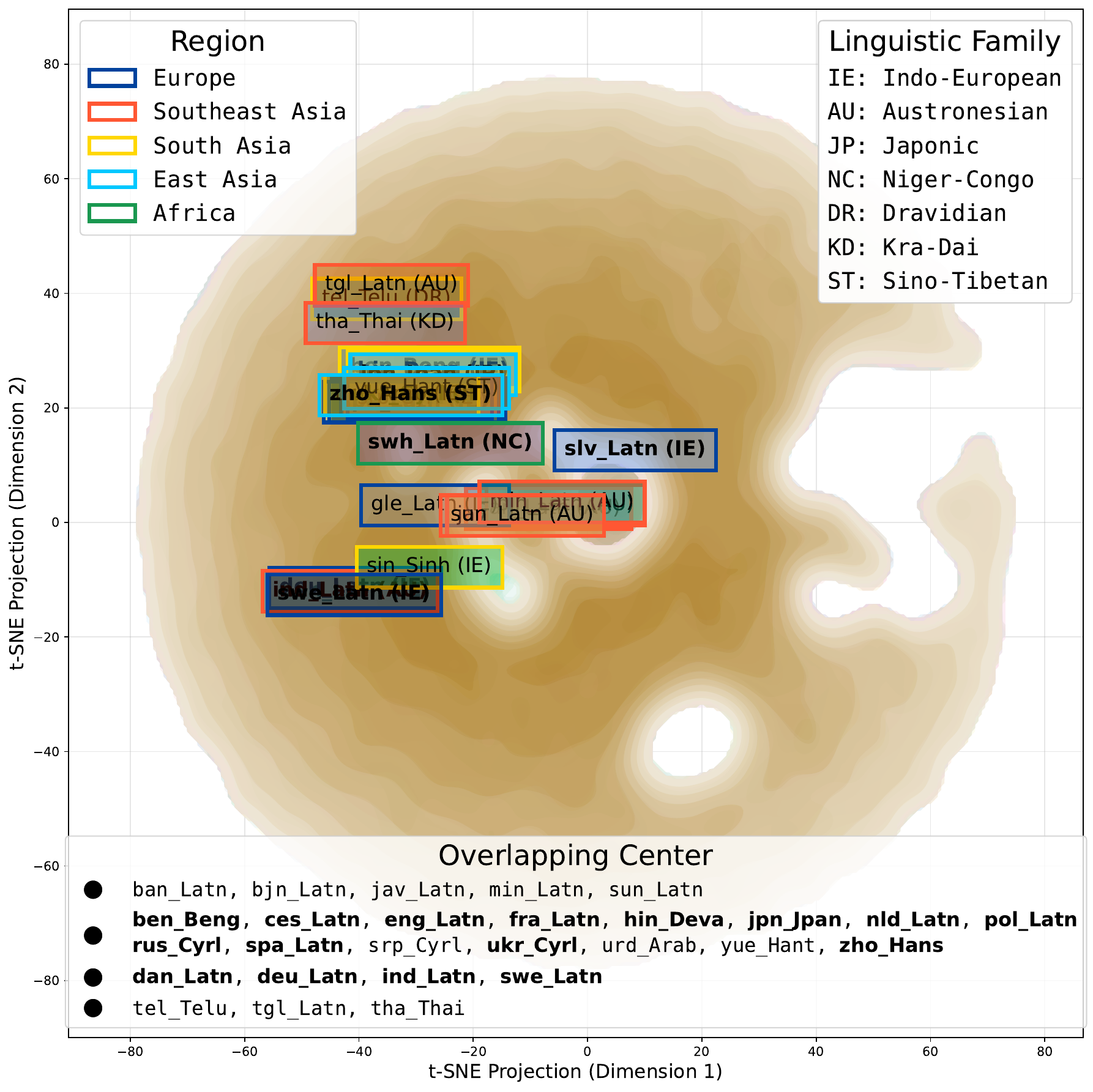}
      \caption{Intermediate (layer 21)}
  \end{subfigure}
  \begin{subfigure}[t]{0.85\columnwidth}
    \centering
      \includegraphics[clip, width=\columnwidth]{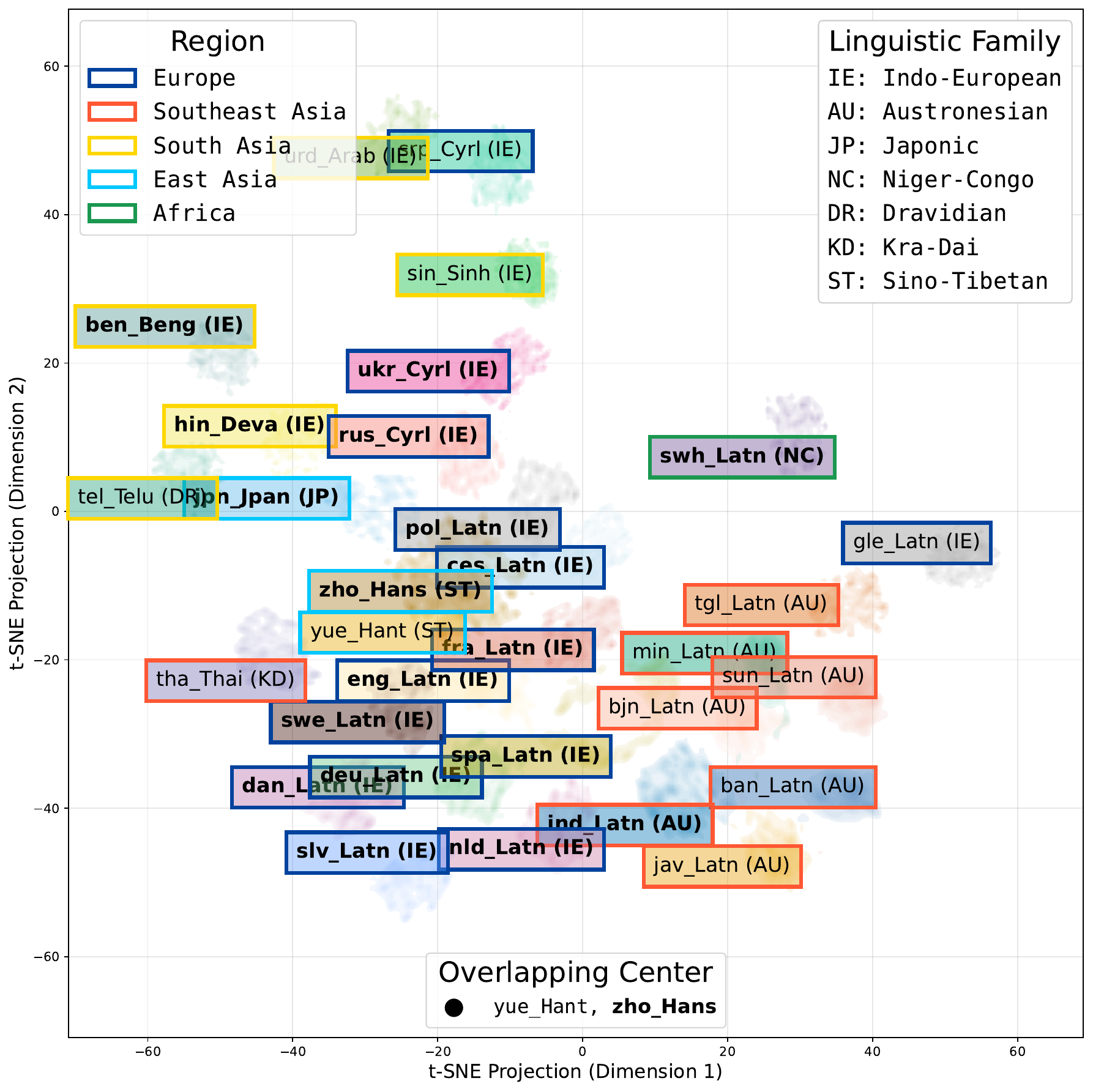}
      \caption{Late (layer 42)}
  \end{subfigure}
  \caption{Hidden-state embeddings of Gemma-2-Instruct (9B) projected in t-SNE dimensions, with HRLs in \textbf{bold}. Interlingual overlaps transcending familial and regional boundaries are observed in the intermediate layer representations. In the early and late layers, language representations cluster w.r.t resource levels and linguistic features, with minimal overlap.}
  \label{fig:tsne_gemma2instruct}
\end{figure}

For example, the t-SNE visualization of LLMs intermediate layers in Figures \ref{fig:tsne_gemma2} and \ref{fig:tsne_aya} shows that Gemma-2 (9B) exhibits more overlapping and closer clustering of language centers compared to Aya Expanse (8B). This observation is further supported by our neuron-wise correlation analysis, showcased in Figures \ref{fig:anc_gemma2_complete} and \ref{fig:anc_aya_complete}, where the intermediate layers of Gemma-2 consistently show mean cross-lingual correlations exceeding 0.5, whereas in the intermediate layers of Aya Expanse, only the mean HRLs-HRLs and in-region records the correlations above 0.5. We conjecture that these variatons on alignment levels stem from the differences in the model architecture and training details of the LLMs.

\section{Observation of Interlingual Alignment Preservation in T-SNE Projections}
\label{app:ilo_in_tsne}

\begin{figure}[!t]
  \centering
  \begin{subfigure}[t]{0.85\columnwidth}
      \includegraphics[clip, width=\columnwidth]{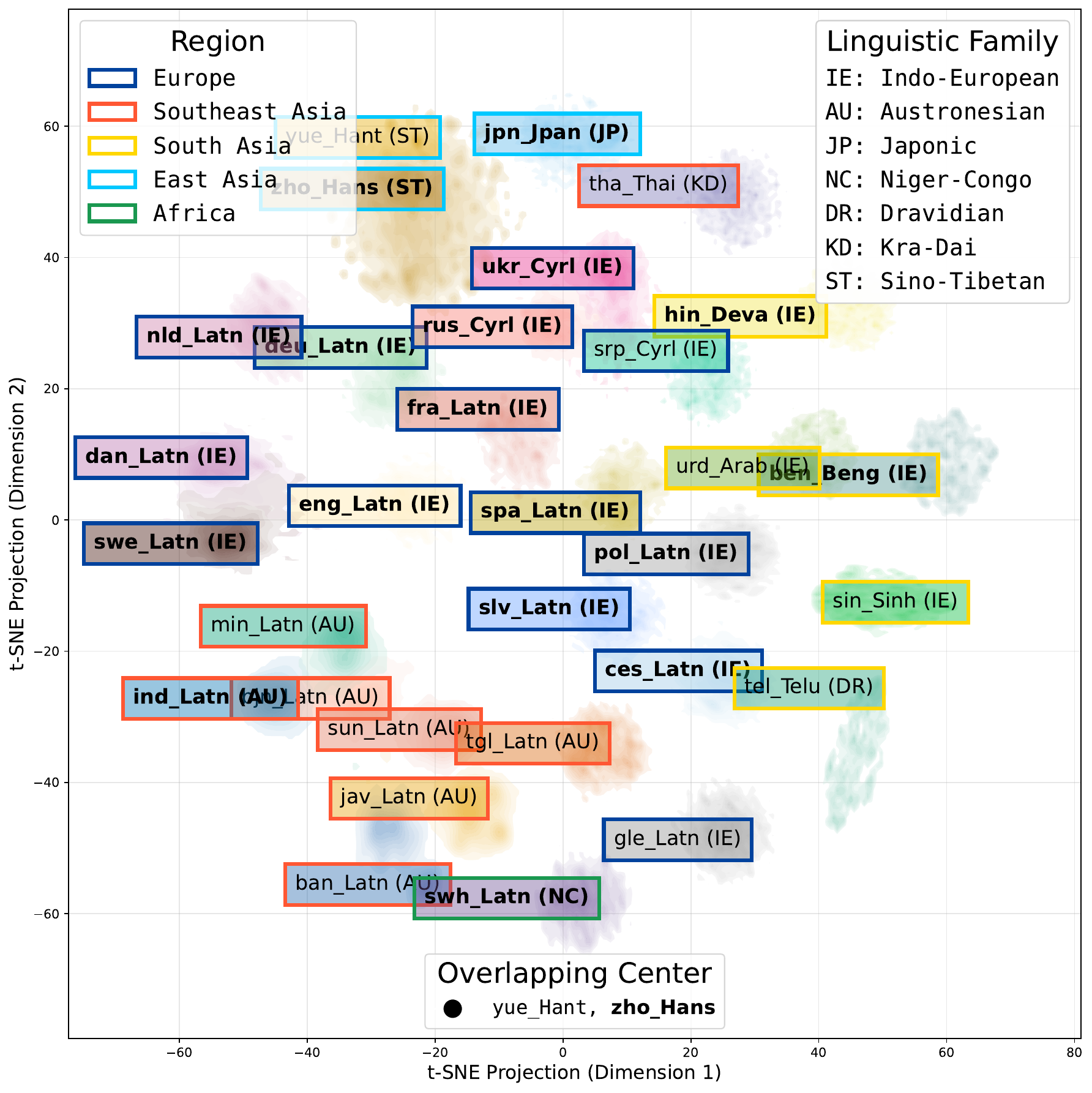}
      \caption{Early (layer 0)}
  \end{subfigure}
  \begin{subfigure}[t]{0.85\columnwidth}
      \includegraphics[clip, width=\columnwidth]{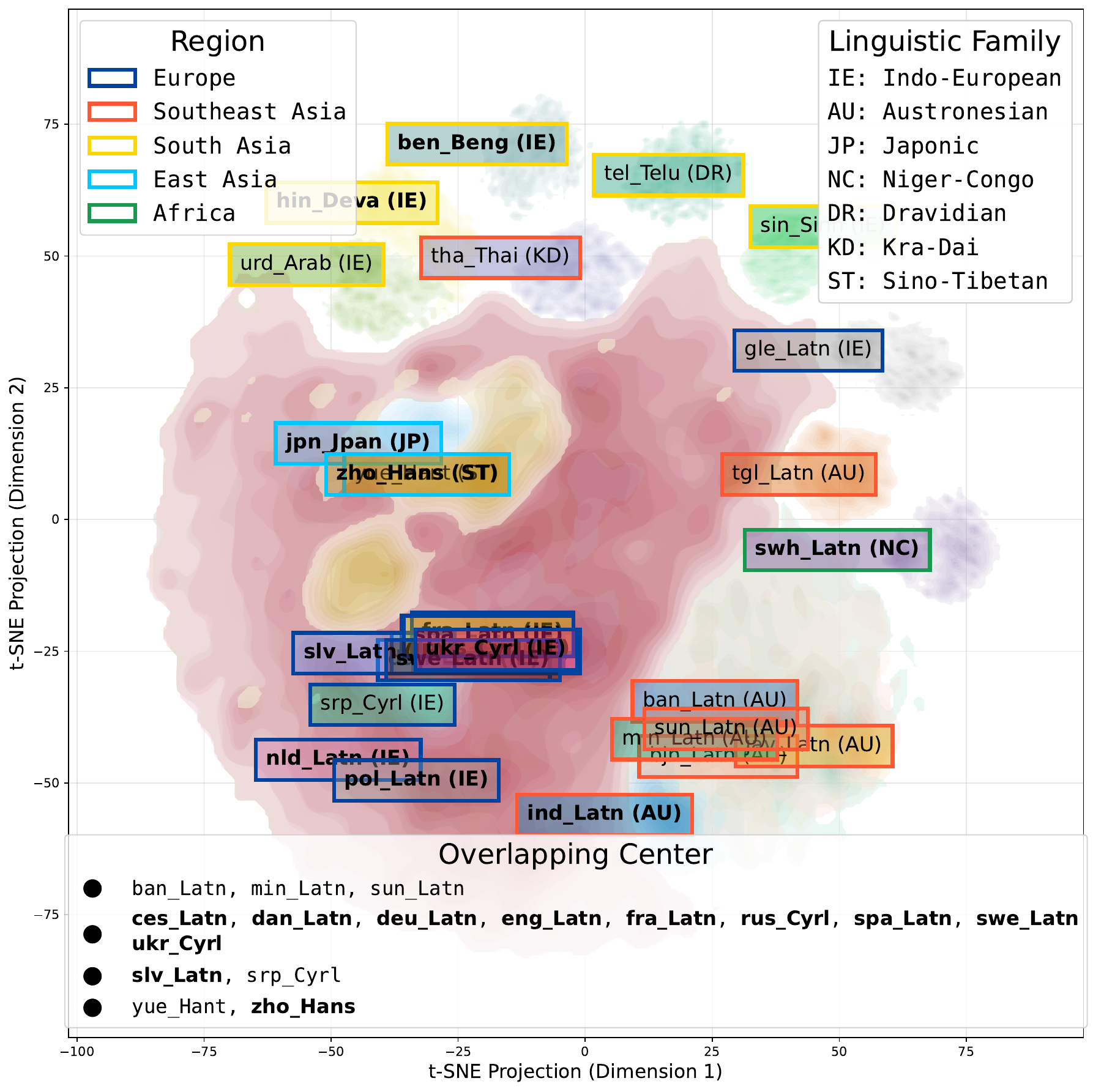}
      \caption{Intermediate (layer 16)}
  \end{subfigure}
  \begin{subfigure}[t]{0.85\columnwidth}
    \centering
      \includegraphics[clip, width=\columnwidth]{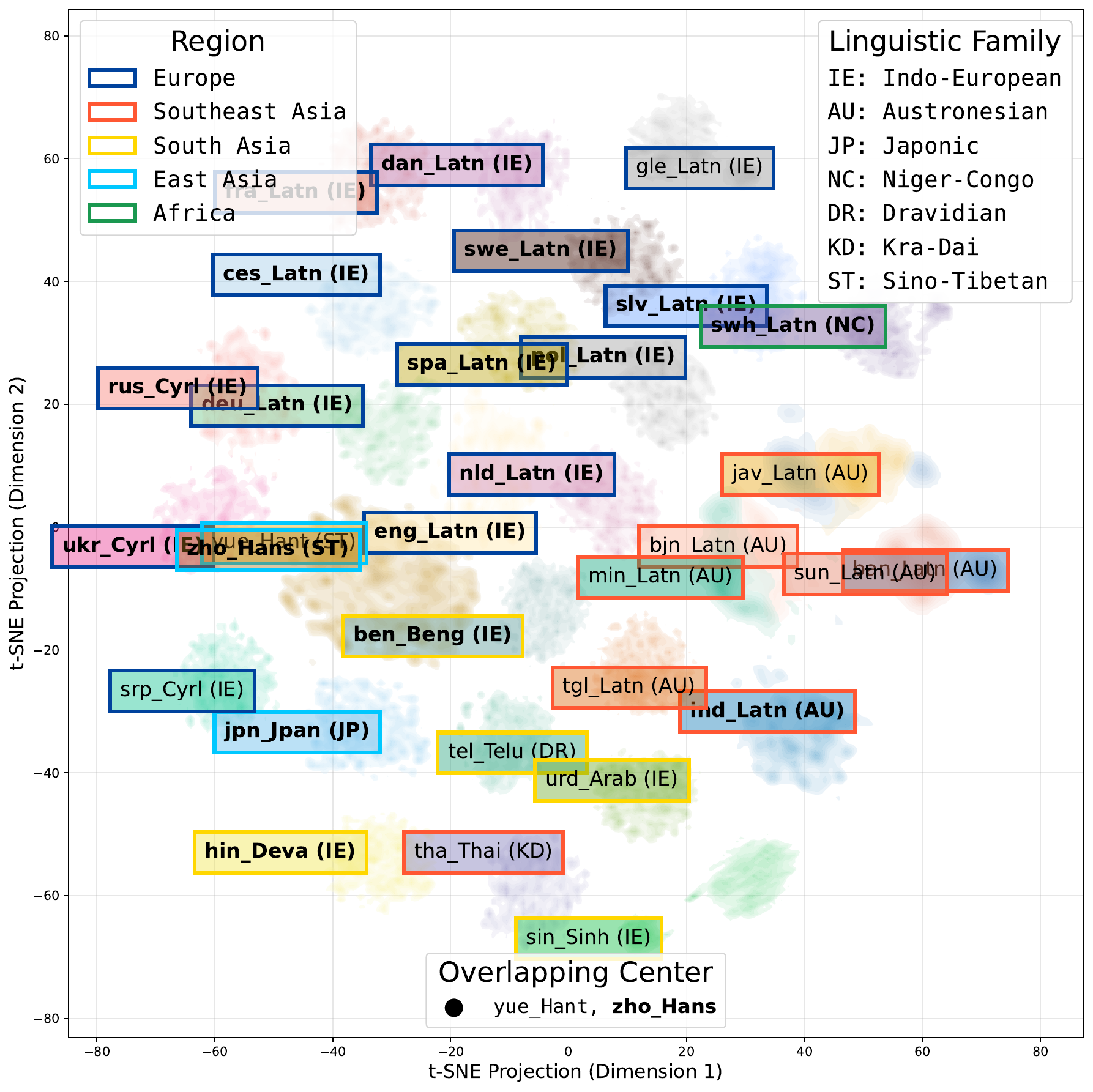}
      \caption{Late (layer 32)}
  \end{subfigure}
  \caption{Hidden-state embeddings of Llama-31 (8B) \textbf{fine-tuned} on single-language dataset on English, projected in t-SNE dimensions, with HRLs in \textbf{bold}. The decline in interlingual semantic alignment is evident from the reduced interlingual overlaps in the projected embeddings within the model's intermediate layer, compared to the observations in Figure~\ref{fig:tsne_llama31}.
  }
  \label{fig:tsne_llama31_ft}
\end{figure}

\begin{figure}[!t]
  \centering
  \begin{subfigure}[t]{0.85\columnwidth}
      \includegraphics[clip, width=\columnwidth]{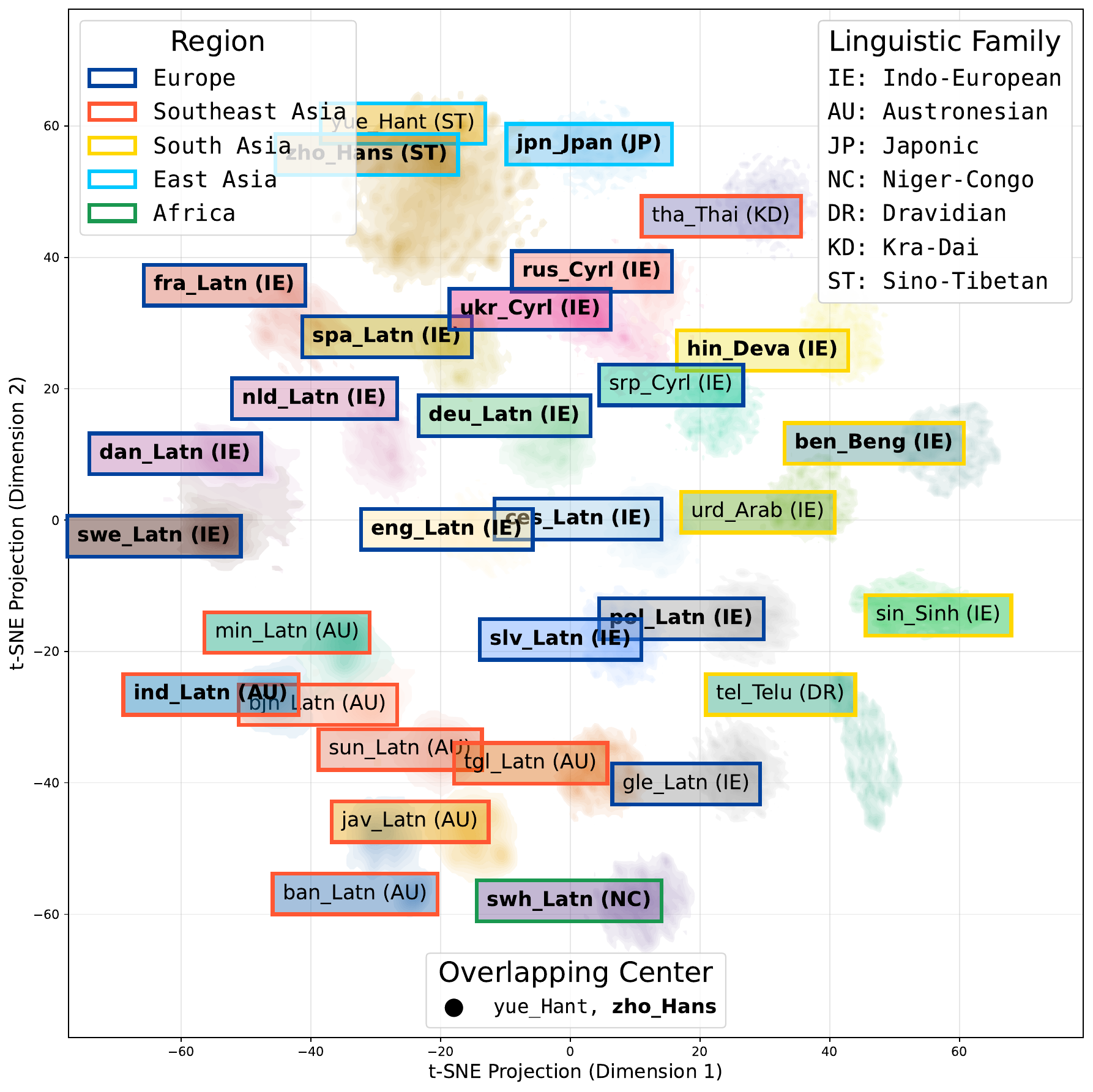}
      \caption{Early (layer 0)}
  \end{subfigure}
  \begin{subfigure}[t]{0.85\columnwidth}
      \includegraphics[clip, width=\columnwidth]{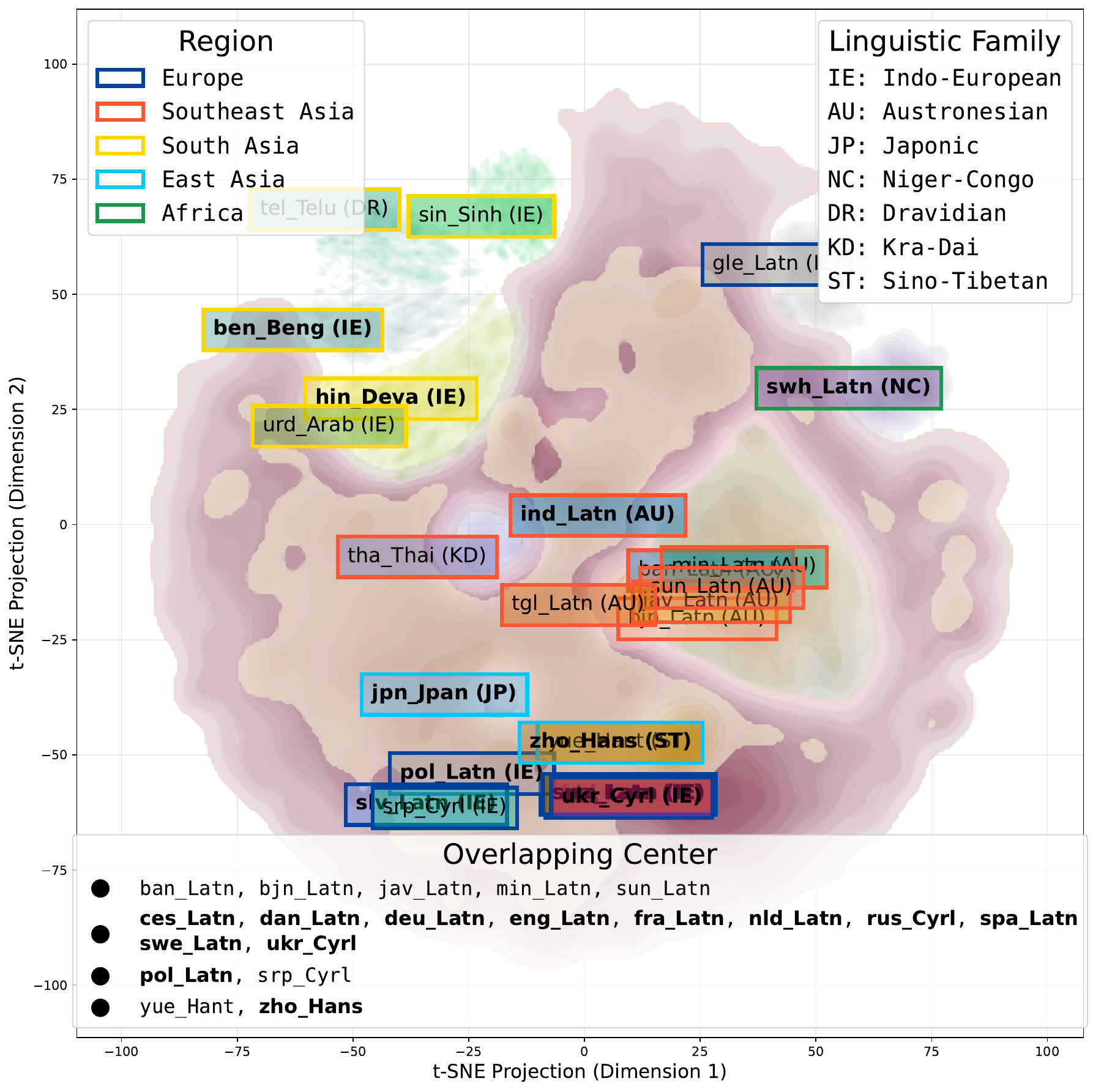}
      \caption{Intermediate (layer 16)}
  \end{subfigure}
  \begin{subfigure}[t]{0.85\columnwidth}
    \centering
      \includegraphics[clip, width=\columnwidth]{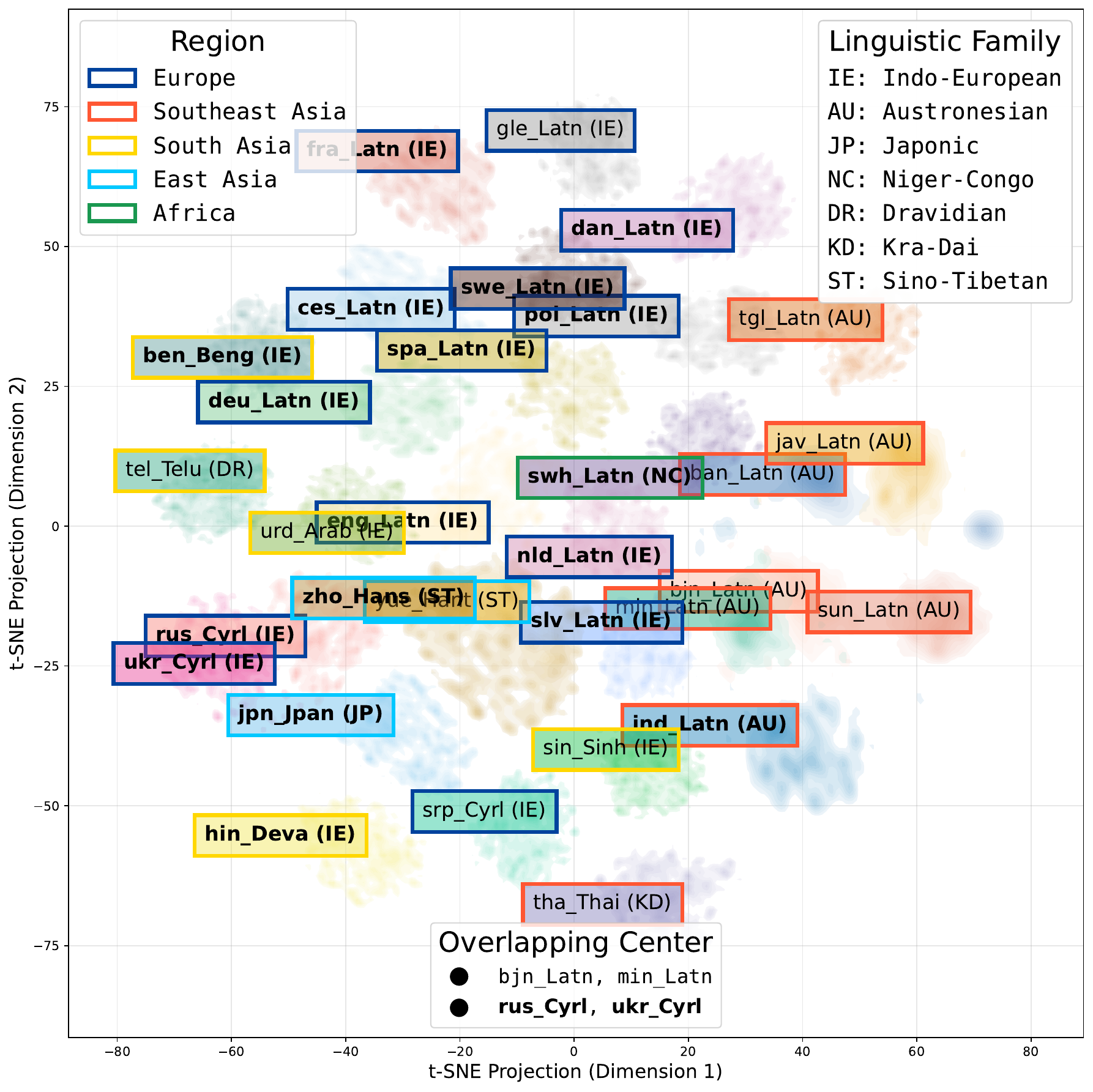}
      \caption{Late (layer 32)}
  \end{subfigure}
  \caption{Hidden-state embeddings of Llama-31 (8B) fine-tuned on single-language dataset on English, with \textbf{selective freezing} strategy, projected in t-SNE dimensions, with HRLs in \textbf{bold}. This approach preserved interlingual alignment, as indicated by high \abbrvmetric{} scores that correlate with observed preservation of interlingual overlaps.}
  \label{fig:tsne_llama31_freeze}
\end{figure}

\begin{figure}[!t]
  \centering
  \begin{subfigure}[t]{0.85\columnwidth}
      \includegraphics[clip, width=\columnwidth]{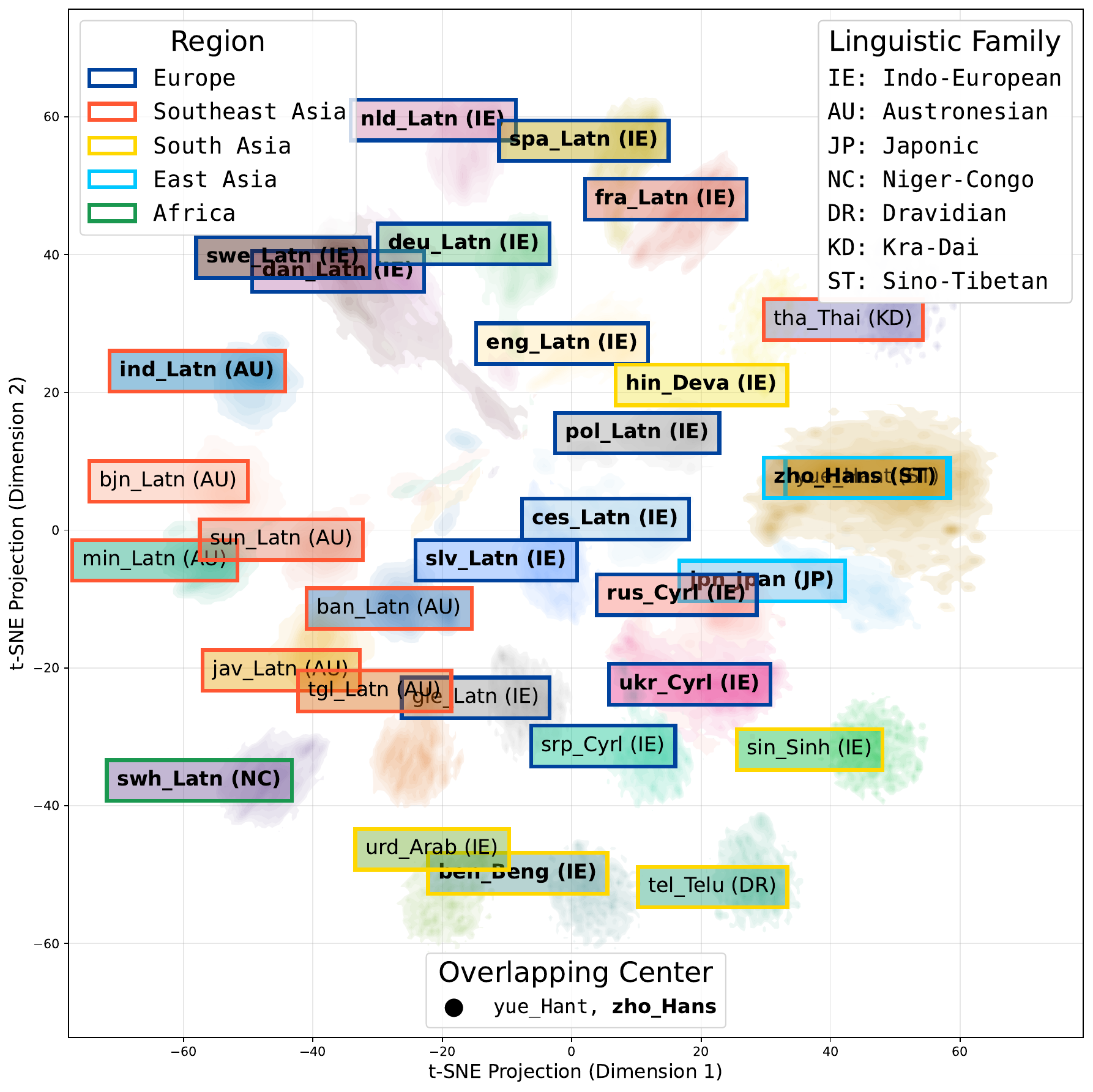}
      \caption{Early (layer 0)}
  \end{subfigure}
  \begin{subfigure}[t]{0.85\columnwidth}
      \includegraphics[clip, width=\columnwidth]{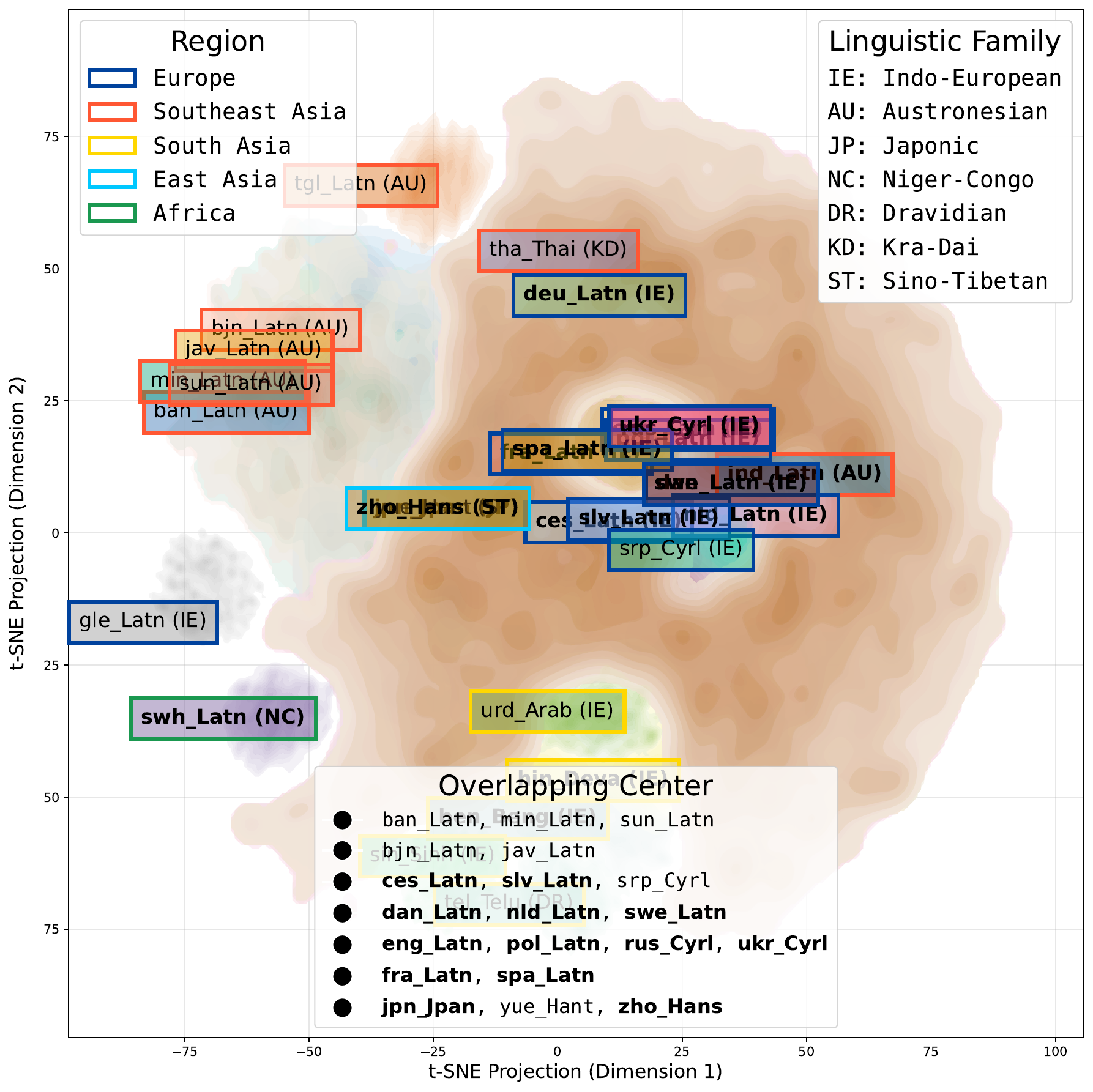}
      \caption{Intermediate (layer 21)}
  \end{subfigure}
  \begin{subfigure}[t]{0.85\columnwidth}
    \centering
      \includegraphics[clip, width=\columnwidth]{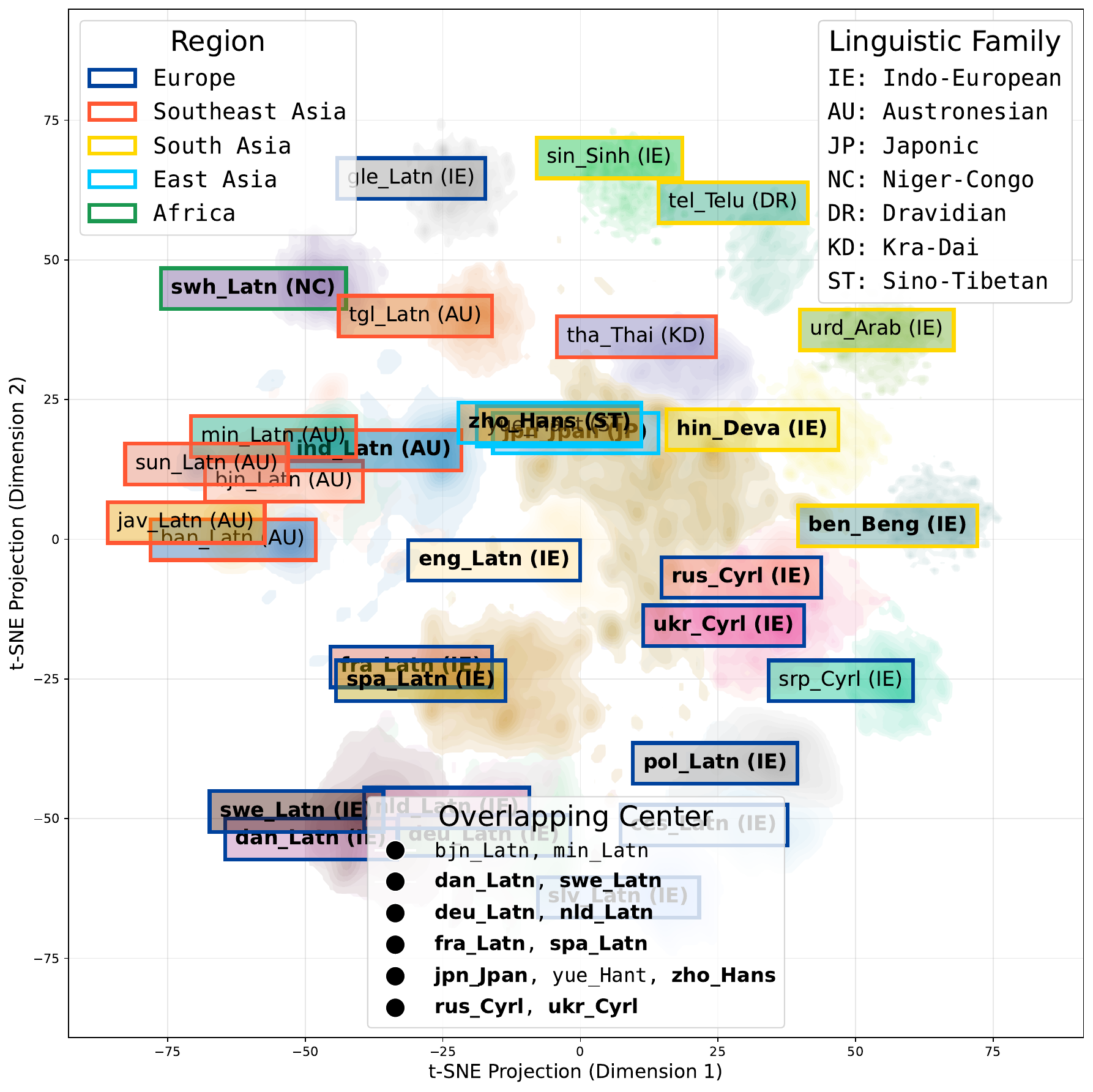}
      \caption{Late (layer 42)}
  \end{subfigure}
  \caption{Hidden-state embeddings of Gemma-2 (9B) \textbf{fine-tuned} on single-language dataset on English, projected in t-SNE dimensions, with HRLs in \textbf{bold}. The decline in interlingual semantic alignment is evident from the reduced interlingual overlaps in the projected embeddings within the model's intermediate layer, compared to the observations in Figure~\ref{fig:tsne_gemma2}.}
  \label{fig:tsne_gemma2_ft}
\end{figure}

\begin{figure}[!t]
  \centering
  \begin{subfigure}[t]{0.85\columnwidth}
      \includegraphics[clip, width=\columnwidth]{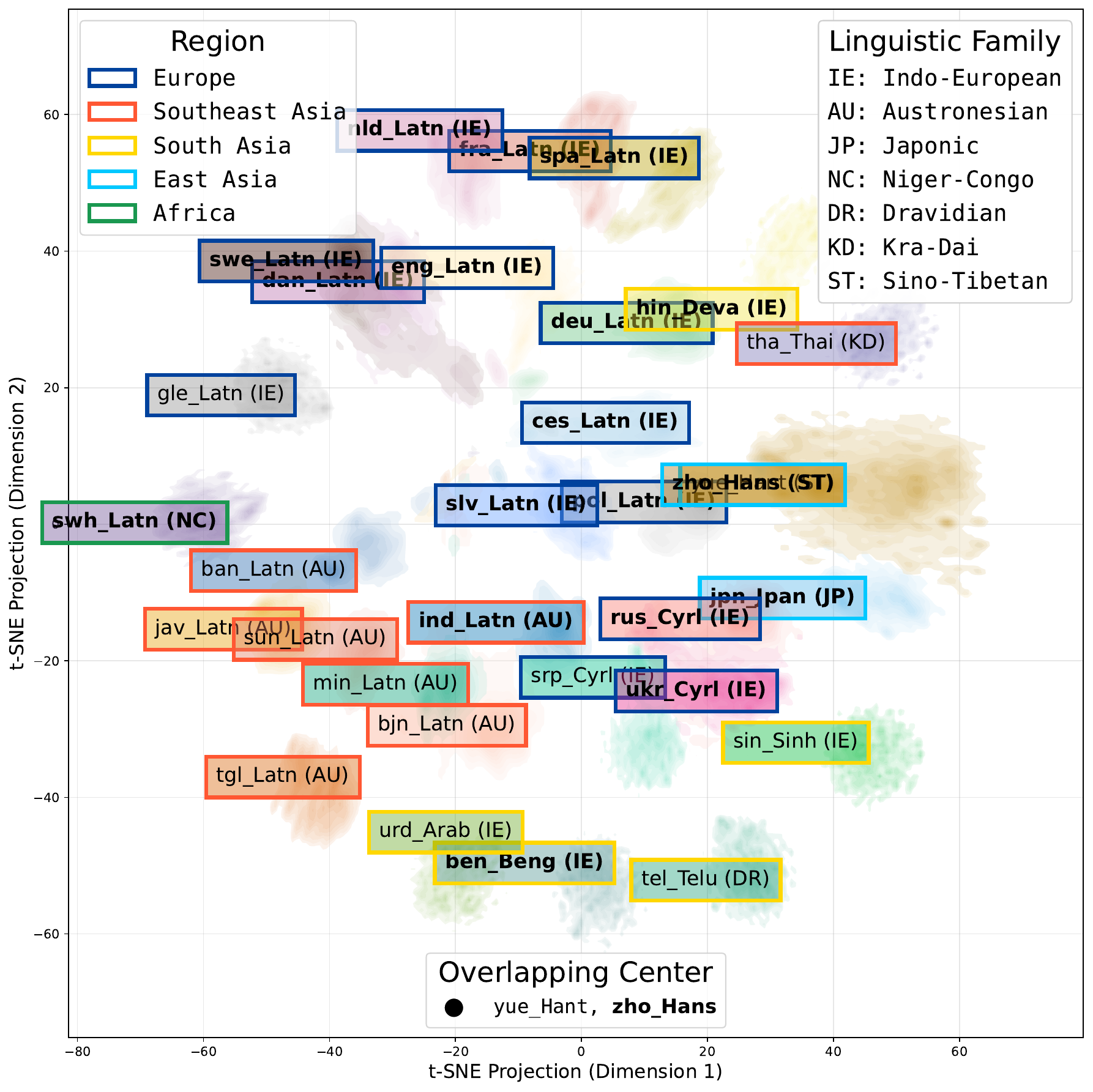}
      \caption{Early (layer 0)}
  \end{subfigure}
  \begin{subfigure}[t]{0.85\columnwidth}
      \includegraphics[clip, width=\columnwidth]{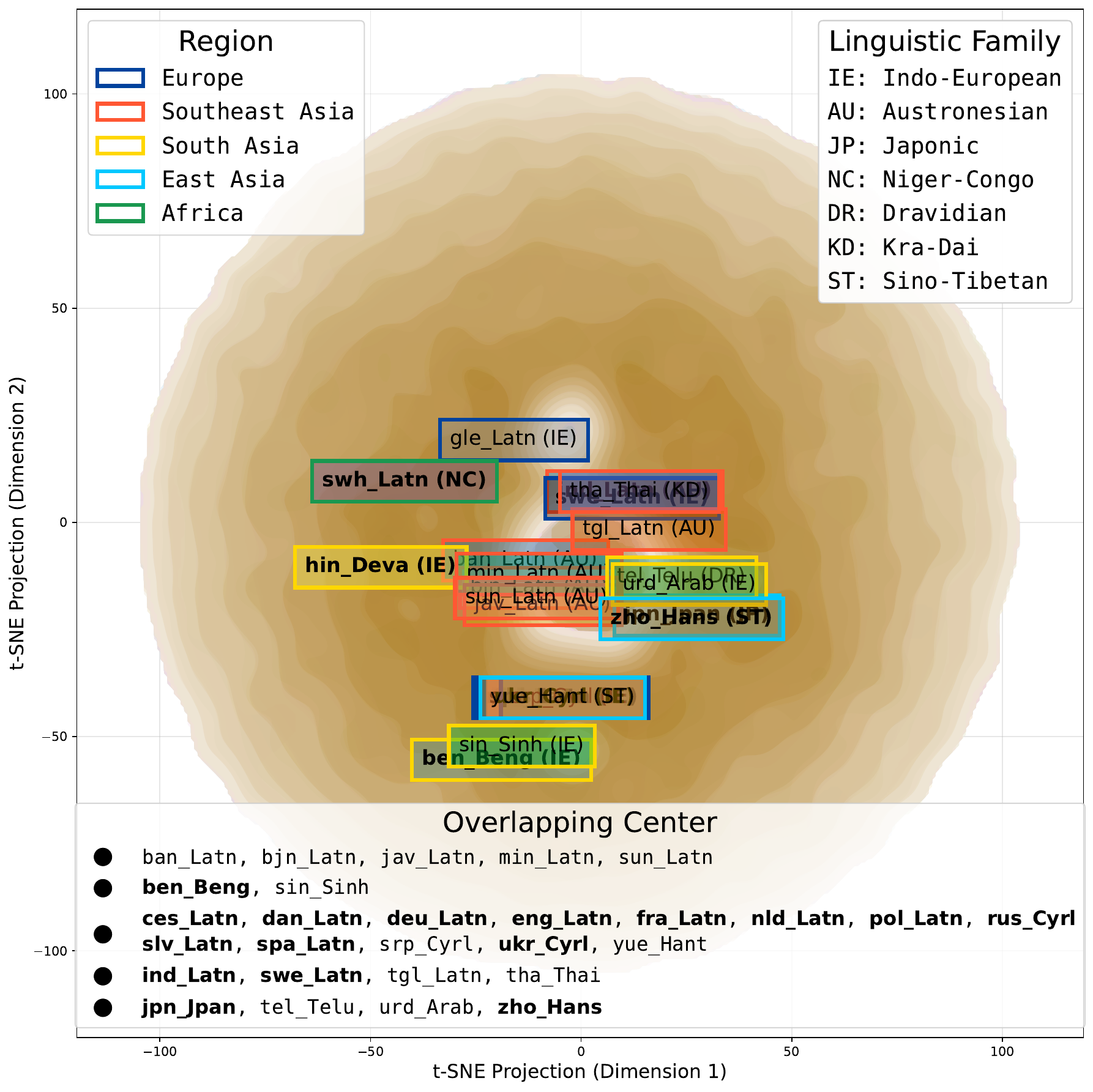}
      \caption{Intermediate (layer 21)}
  \end{subfigure}
  \begin{subfigure}[t]{0.85\columnwidth}
    \centering
      \includegraphics[clip, width=\columnwidth]{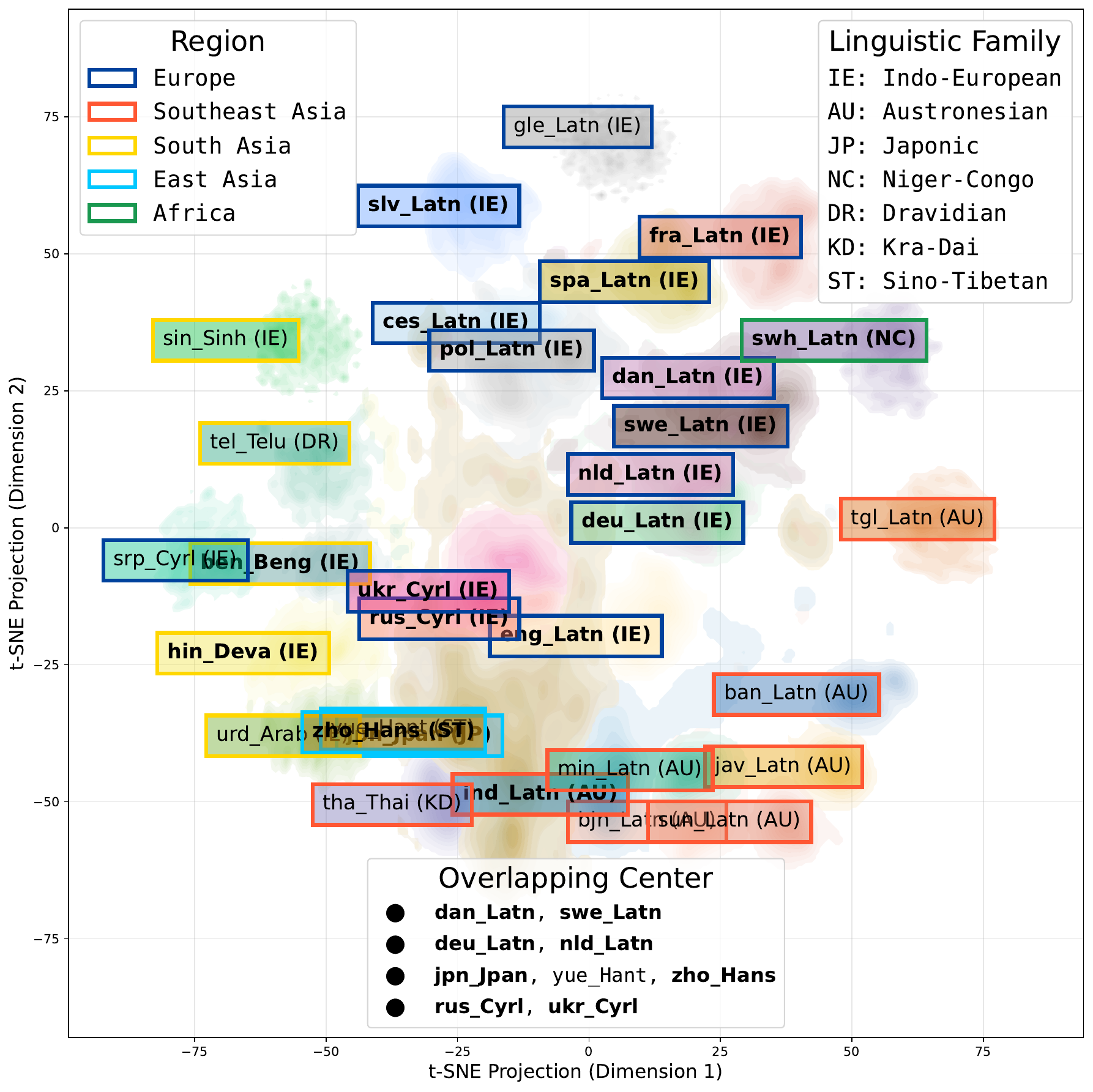}
      \caption{Late (layer 42)}
  \end{subfigure}
  \caption{Hidden-state embeddings of Gemma-2 (9B) fine-tuned on single-language dataset on English, with \textbf{selective freezing} strategy, projected in t-SNE dimensions, with HRLs in \textbf{bold}. This approach preserved interlingual alignment, as indicated by high \abbrvmetric{} scores that correlate with observed preservation of interlingual overlaps.}
  \label{fig:tsne_gemma2_freeze}
\end{figure}

Through our single-language training experiments in the multilingual mathematical reasoning task, we observe that the visual projections using t-SNE, also support that \abbrvmetric{} score effectively captures the same interlingual alignment phenomenon, albeit in a projected lower-dimensional dimensions. In other words, layers with high \abbrvmetric{} scores consistently exhibits interlingual overlaps in the t-SNE dimensions that hints at strong interlingual alignment, whereas those with lower scores tend to be more fragmented. This correspondence validates \abbrvmetric{} as a robust quantitative measure that reflects the local structure of the multilingual shared embedding space. We attach the complete t-SNE visualization projected from the hidden-states of the models underwent single-language training on English in the \textbf{fine-tuning} vs \textbf{selective freezing} modes, frozen on their first 8 layers, the token embedding, final layer normalization, and the language modeling head (output projection layers), of Llama-3.1 (8B) and Gemma-2 (9B) respectively in Figures~\ref{fig:tsne_llama31_ft} vs ~\ref{fig:tsne_llama31_freeze}, and ~\ref{fig:tsne_gemma2_ft} vs ~\ref{fig:tsne_gemma2_freeze}.

\begin{table*}[!ht]
\centering
\resizebox{0.94\linewidth}{!}{%
\begin{tabular}{l|c|rrrrrrrrrrr|rr}
\toprule
& & \multicolumn{11}{c}{\textbf{Accuracy}} & \multicolumn{2}{|c}{\textbf{Average}} \\
\cmidrule(lr){3-13}\cmidrule(lr){14-15}
\multirow{-2.3}{*}{\textbf{Method}} & \multirow{-2.2}{*}{\shortstack{\textbf{Training} \\ \textbf{languages}}} & \multicolumn{1}{c}{\textbf{ben}} & \multicolumn{1}{c}{\textbf{tha*}} & \multicolumn{1}{c}{\textbf{swh}} & \multicolumn{1}{c}{\textbf{tel*}} & \multicolumn{1}{c}{\textbf{jpn}} & \multicolumn{1}{c}{\textbf{zho}} & \multicolumn{1}{c}{\textbf{deu}} & \multicolumn{1}{c}{\textbf{fra}} & \multicolumn{1}{c}{\textbf{rus}} & \multicolumn{1}{c}{\textbf{spa}} & \multicolumn{1}{c}{\textbf{eng}} & \multicolumn{1}{|c}{\textbf{All}} & \multicolumn{1}{c}{\textbf{XL}} \\ \midrule
Pre-trained & mixed & 13.2\% & 12.0\% & 9.2\% & 16.0\% & 10.0\% & 17.6\% & 16.8\% & 16.8\% & 10.8\% & 15.2\% & 17.6\% & 11.2\% & \multicolumn{1}{c}{-} \\ \midrule
\multirow{9.7}{*}{Fine-tuning} & ben & \cellcolor[HTML]{00FFFF}\textbf{27.6\%} & \cellcolor[HTML]{EA9999}4.4\% & \cellcolor[HTML]{EA9999}2.0\% & \cellcolor[HTML]{EA9999}4.4\% & \cellcolor[HTML]{EA9999}11.6\% & \cellcolor[HTML]{EA9999}12.8\% & \cellcolor[HTML]{EA9999}6.8\% & \cellcolor[HTML]{EA9999}10.4\% & \cellcolor[HTML]{EA9999}10.0\% & \cellcolor[HTML]{EA9999}14.4\% & \cellcolor[HTML]{EA9999}18.4\% & 11.2\% & 9.5\% \\
 & tha* & \cellcolor[HTML]{EA9999}5.6\% & \cellcolor[HTML]{00FFFF}\textbf{32.4\%} & \cellcolor[HTML]{EA9999}6.0\% & \cellcolor[HTML]{EA9999}2.8\% & 10.4\% & \cellcolor[HTML]{EA9999}14.4\% & \cellcolor[HTML]{EA9999}14.8\% & 16.8\% & 12.0\% & 20.0\% & 26.0\% & 14.7\% & 12.9\% \\
 & swh & \cellcolor[HTML]{EA9999}5.6\% & \cellcolor[HTML]{EA9999}5.6\% & \cellcolor[HTML]{00FFFF}{\ul \textbf{32.4\%}} & \cellcolor[HTML]{EA9999}0.8\% & 10.4\% & \cellcolor[HTML]{EA9999}9.6\% & \cellcolor[HTML]{EA9999}15.6\% & \cellcolor[HTML]{EA9999}14.8\% & 10.8\% & 21.2\% & 26.4\% & 13.9\% & 12.1\% \\
 & jpn & \cellcolor[HTML]{EA9999}2.4\% & \cellcolor[HTML]{EA9999}6.0\% & \cellcolor[HTML]{EA9999}2.8\% & \cellcolor[HTML]{EA9999}2.4\% & \cellcolor[HTML]{00FFFF}\textbf{26.8\%} & 19.6\% & \cellcolor[HTML]{EA9999}13.2\% & \cellcolor[HTML]{EA9999}10.8\% & 14.4\% & 18.0\% & 26.0\% & 12.9\% & 11.6\% \\
 & zho & \cellcolor[HTML]{EA9999}2.0\% & \cellcolor[HTML]{EA9999}6.4\% & \cellcolor[HTML]{EA9999}1.6\% & \cellcolor[HTML]{EA9999}0.8\% & 16.8\% & \cellcolor[HTML]{00FFFF}\textbf{32.0\%} & 17.6\% & \cellcolor[HTML]{EA9999}10.4\% & 16.4\% & 18.0\% & 28.0\% & 13.6\% & 11.8\% \\
 & deu & \cellcolor[HTML]{EA9999}4.4\% & \cellcolor[HTML]{EA9999}9.2\% & \cellcolor[HTML]{EA9999}5.2\% & \cellcolor[HTML]{EA9999}\textbf{6.8\%} & 16.0\% & 18.4\% & \cellcolor[HTML]{00FFFF}\textbf{32.8\%} & 23.6\% & 23.2\% & 26.4\% & 34.4\% & 18.2\% & 16.8\% \\
 & fra & \cellcolor[HTML]{EA9999}5.6\% & \cellcolor[HTML]{EA9999}10.8\% & \cellcolor[HTML]{EA9999}6.0\% & \cellcolor[HTML]{EA9999}0.8\% & 17.6\% & 18.8\% & 29.2\% & \cellcolor[HTML]{00FFFF}\textbf{30.8\%} & 21.6\% & 29.6\% & 31.6\% & 18.4\% & 17.2\% \\
 & rus & \cellcolor[HTML]{EA9999}4.8\% & \cellcolor[HTML]{EA9999}4.8\% & \cellcolor[HTML]{EA9999}5.2\% & \cellcolor[HTML]{EA9999}1.2\% & 13.2\% & \cellcolor[HTML]{EA9999}16.8\% & 30.0\% & 24.4\% & \cellcolor[HTML]{00FFFF}\textbf{32.8\%} & 29.2\% & 29.2\% & 17.4\% & 15.9\% \\
 & spa & \cellcolor[HTML]{EA9999}7.2\% & \cellcolor[HTML]{EA9999}7.6\% & \cellcolor[HTML]{EA9999}4.8\% & \cellcolor[HTML]{EA9999}4.4\% & 17.6\% & 22.0\% & 26.8\% & 27.6\% & 28.4\% & \cellcolor[HTML]{00FFFF}\textbf{33.2\%} & 37.6\% & \textbf{19.7\%} & \textbf{18.4\%} \\
 & eng & \cellcolor[HTML]{EA9999}8.0\% & \cellcolor[HTML]{EA9999}10.4\% & \cellcolor[HTML]{EA9999}8.0\% & \cellcolor[HTML]{EA9999}6.0\% & 17.6\% & 20.8\% & 28.0\% & 24.4\% & 25.2\% & 29.6\% & \cellcolor[HTML]{00FFFF}\textbf{39.2\%} & \textbf{19.7\%} & 17.8\% \\ \midrule
\multirow{9.7}{*}{\shortstack[l]{Selective \\ Freezing}} & ben & \cellcolor[HTML]{00FFFF}{\ul \textbf{36.0\%}} & 13.2\% & 17.2\% & 20.0\% & 22.8\% & 19.6\% & 19.6\% & 22.0\% & 21.2\% & 18.0\% & 26.8\% & 21.5\% & 20.0\% \\
 & tha* & 14.4\% & \cellcolor[HTML]{00FFFF}{\ul \textbf{34.4\%}} & 14.0\% & \cellcolor[HTML]{EA9999}13.6\% & 16.8\% & 21.6\% & 20.0\% & 22.8\% & 21.2\% & 24.8\% & 27.2\% & 21.0\% & 19.6\% \\
 & swh & 13.2\% & 14.4\% & \cellcolor[HTML]{00FFFF}{\ul \textbf{30.4\%}} & \cellcolor[HTML]{EA9999}11.2\% & 15.2\% & 20.4\% & 26.8\% & 25.2\% & 20.8\% & 29.6\% & 29.6\% & 21.5\% & 20.6\% \\
 & jpn & \cellcolor[HTML]{EA9999}12.8\% & 14.8\% & 19.2\% & \cellcolor[HTML]{EA9999}13.2\% & \cellcolor[HTML]{00FFFF}{\ul \textbf{27.6\%}} & 26.8\% & 22.0\% & 21.6\% & 23.6\% & 21.6\% & 26.4\% & 20.9\% & 20.2\% \\
 & zho & \cellcolor[HTML]{EA9999}12.8\% & 19.2\% & 15.6\% & \cellcolor[HTML]{EA9999}13.6\% & 22.0\% & \cellcolor[HTML]{00FFFF}{\ul \textbf{34.8\%}} & 26.4\% & 27.2\% & 22.4\% & 24.8\% & 31.2\% & 22.7\% & 21.5\% \\
 & deu & \cellcolor[HTML]{EA9999}11.2\% & 17.6\% & 18.8\% & \cellcolor[HTML]{EA9999}14.0\% & 20.0\% & 21.2\% & \cellcolor[HTML]{00FFFF}{33.6\%} & 26.0\% & 26.8\% & 28.0\% & 35.2\% & 22.9\% & 21.9\% \\
 & fra & 20.4\% & 17.6\% & 22.4\% & 20.0\% & 23.6\% & 24.0\% & 30.4\% & \cellcolor[HTML]{00FFFF}{35.6\%} & 28.4\% & 33.2\% & 32.8\% & 26.2\% & 25.3\% \\
 & rus & 15.2\% & 17.6\% & 24.0\% & 17.2\% & 18.4\% & 18.4\% & 28.8\% & 26.0\% & \cellcolor[HTML]{00FFFF}{\ul \textbf{36.4\%}} & 27.6\% & 32.4\% & 23.8\% & 22.6\% \\
 & spa & 18.4\% & 21.2\% & 26.4\% & 18.8\% & 22.0\% & 26.4\% & {\ul \textbf{36.4\%}} & 31.6\% & 29.2\% & \cellcolor[HTML]{00FFFF}35.6\% & 38.8\% & 27.7\% & 26.9\% \\
 & eng & 22.4\% & 25.6\% & 26.8\% & {\ul \textbf{22.4\%}} & 24.8\% & 26.0\% & 34.4\% & {\ul \textbf{36.0\%}} & 34.0\% & {\ul \textbf{39.2\%}} & \cellcolor[HTML]{00FFFF}{\ul \textbf{41.6\%}} & {\ul \textbf{30.3\%}} & {\ul \textbf{29.2\%}} \\ \bottomrule
\end{tabular}}
\caption{Cross-lingual transfer performance on MGSM for Gemma-2 (9B) w/ and w/o selective freezing. ``XL'' denotes average on languages that were not fine-tuned. Diagonal entries in \sethlcolor{highlightBlue}\hl{blue highlights} correspond to source language performances. \sethlcolor{highlightRed}\hl{Red highlights} indicate decrease from pre-trained baseline. \textbf{Bold} and {\ul underline} respectively denote the best within group and within column. The (*) marks languages classified as low-resource in Flores-200.}
\label{tab:xlt_frz_gemma}
\end{table*}

\section{Ablation Studies}

Here we provide comprehensive ablations to all of the hyperparameters in our study and thoroughly analyzes the impact on each of them.

\subsection{t-SNE perplexity}
\label{sec:appendix_tsne_ppl}

\begin{figure}[!t]
  \centering
  \begin{subfigure}[t]{0.85\columnwidth}
      \includegraphics[clip, width=\columnwidth]{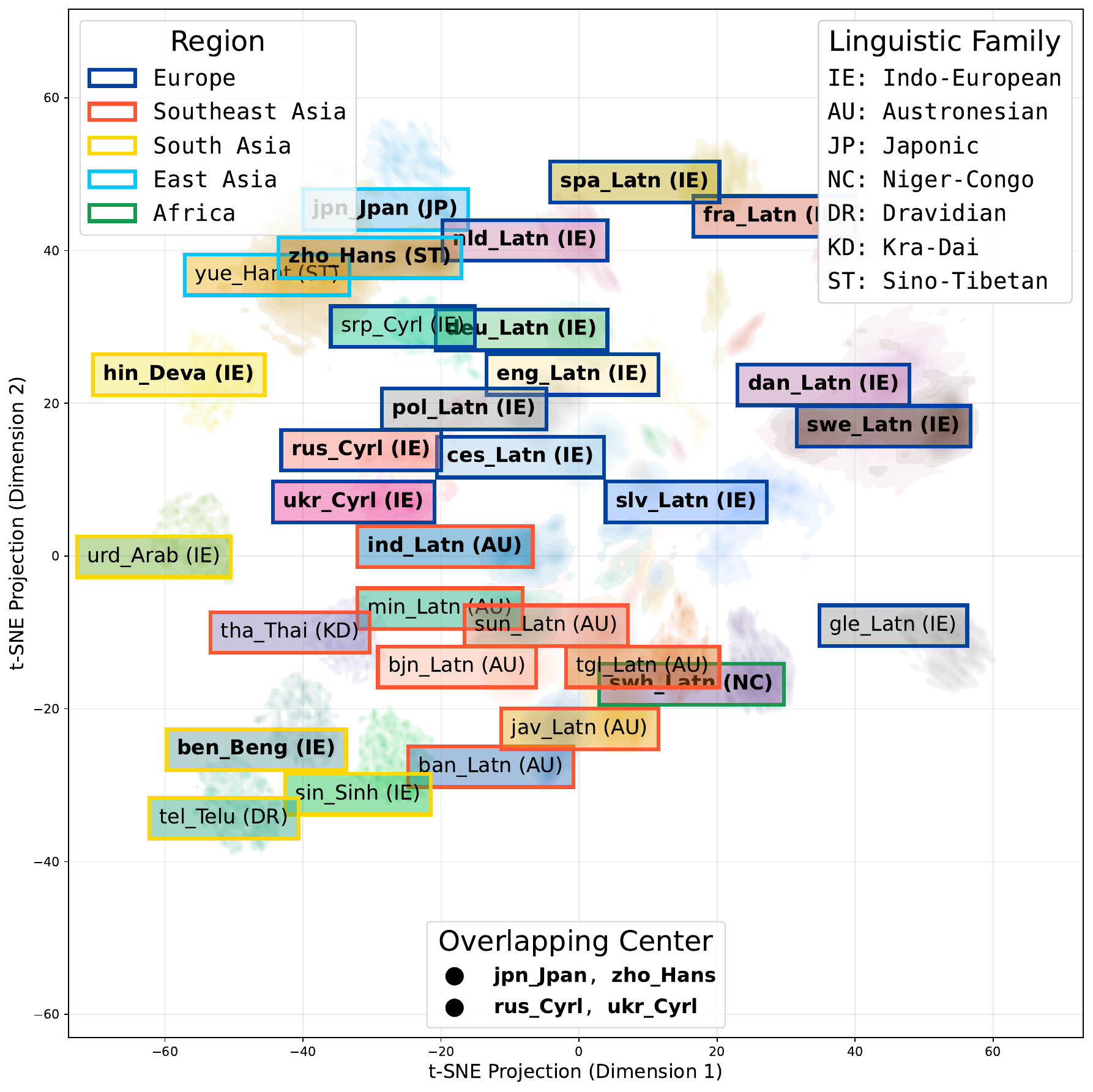}
      \caption{Early (layer 0), perplexity = 5}
  \end{subfigure}
  \begin{subfigure}[t]{0.85\columnwidth}
      \includegraphics[clip, width=\columnwidth]{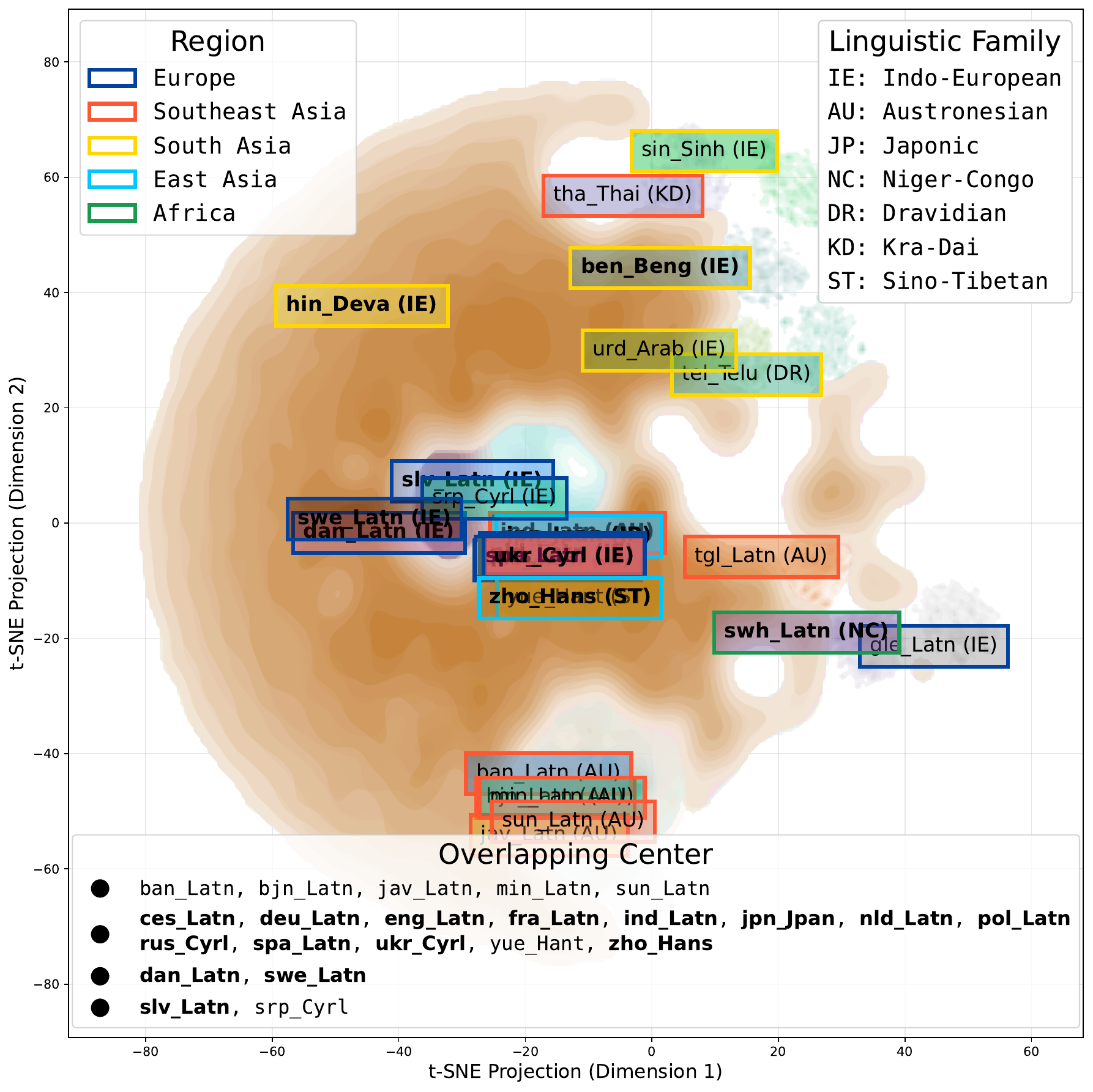}
      \caption{Intermediate (layer 16), perplexity = 5}
  \end{subfigure}
  \begin{subfigure}[t]{0.85\columnwidth}
    \centering
      \includegraphics[clip, width=\columnwidth]{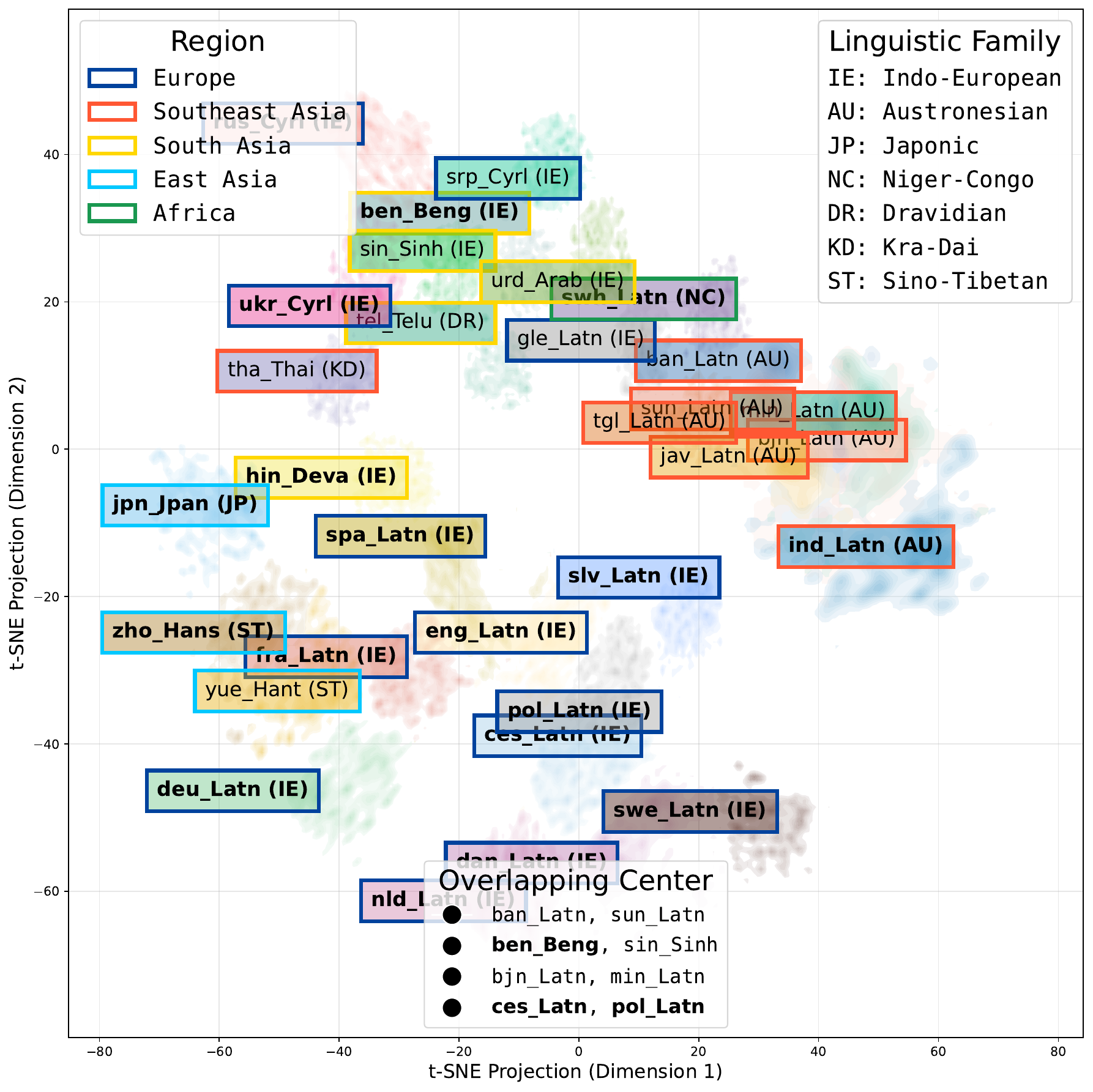}
      \caption{Late (layer 32), perplexity = 5}
  \end{subfigure}
  \caption{Hidden-state embeddings of Aya Expanse (8B) projected in t-SNE dimensions, with HRLs in \textbf{bold}. The t-SNE visualizations are derived using the perplexity value of 5.}
  \label{fig:abl_ppl5}
\end{figure}

\begin{figure}[!t]
  \centering
  \begin{subfigure}[t]{0.85\columnwidth}
      \includegraphics[clip, width=\columnwidth]{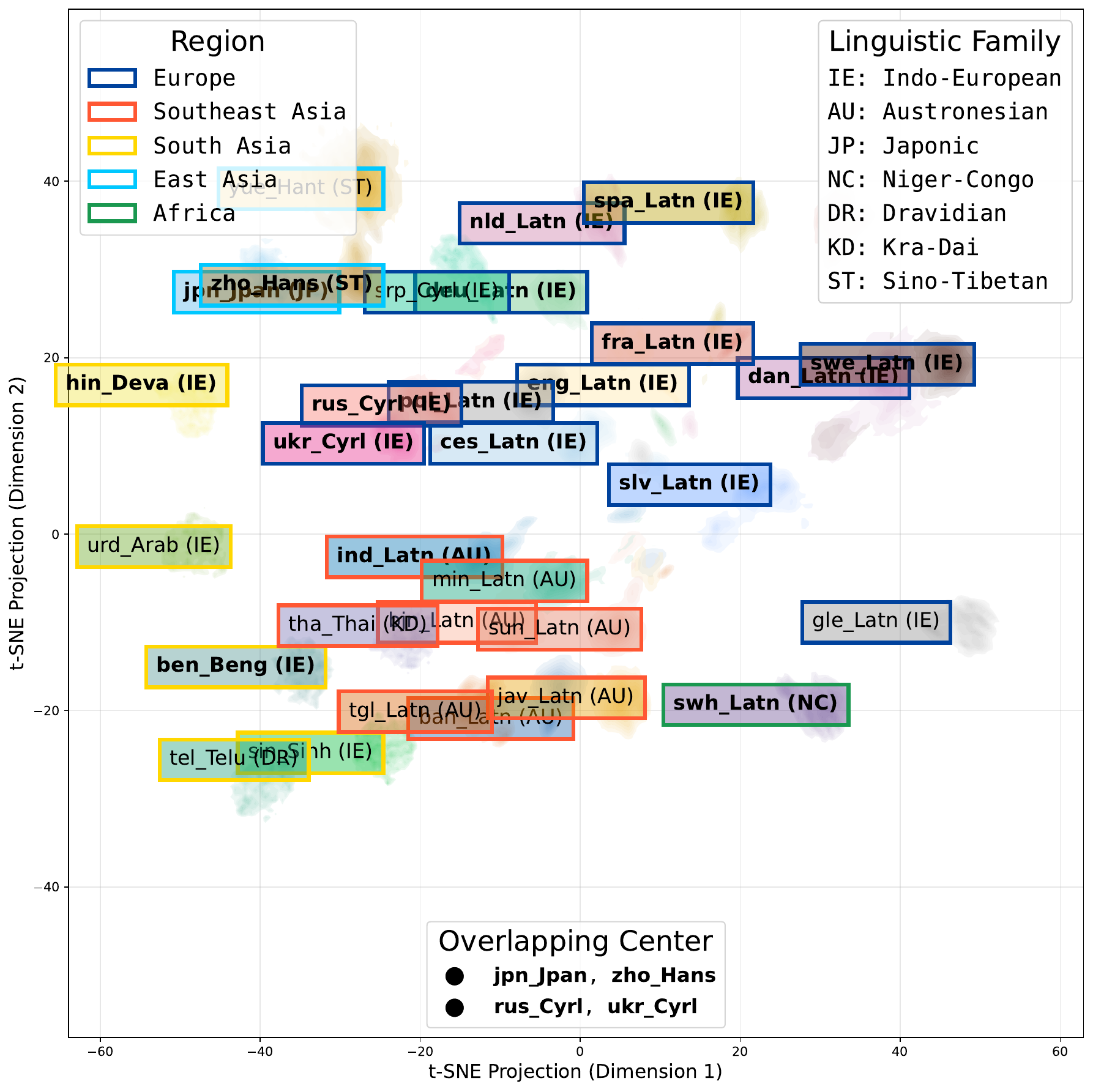}
      \caption{Early (layer 0), perplexity = 15}
  \end{subfigure}
  \begin{subfigure}[t]{0.85\columnwidth}
      \includegraphics[clip, width=\columnwidth]{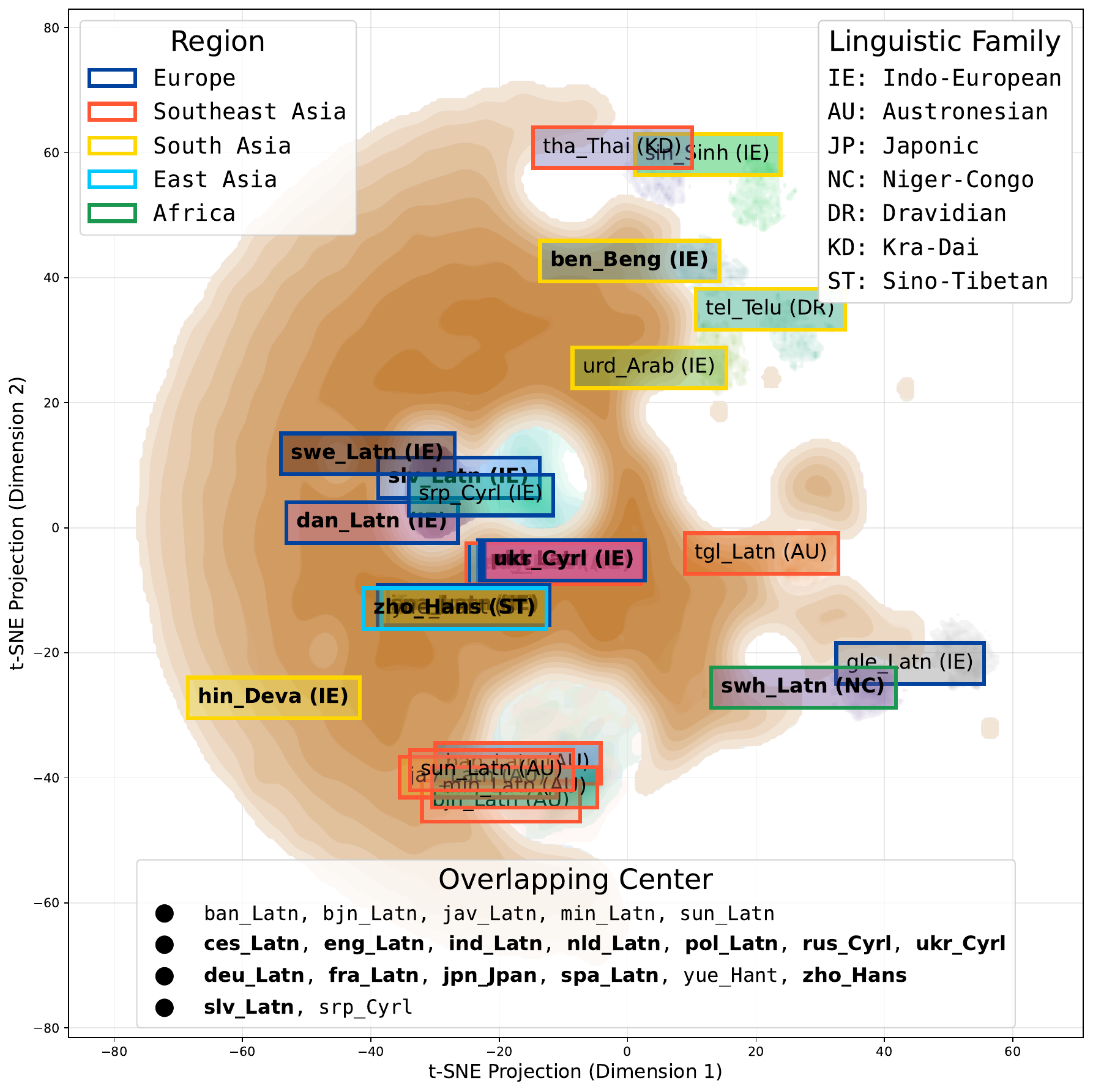}
      \caption{Intermediate (layer 16), perplexity = 15}
  \end{subfigure}
  \begin{subfigure}[t]{0.85\columnwidth}
    \centering
      \includegraphics[clip, width=\columnwidth]{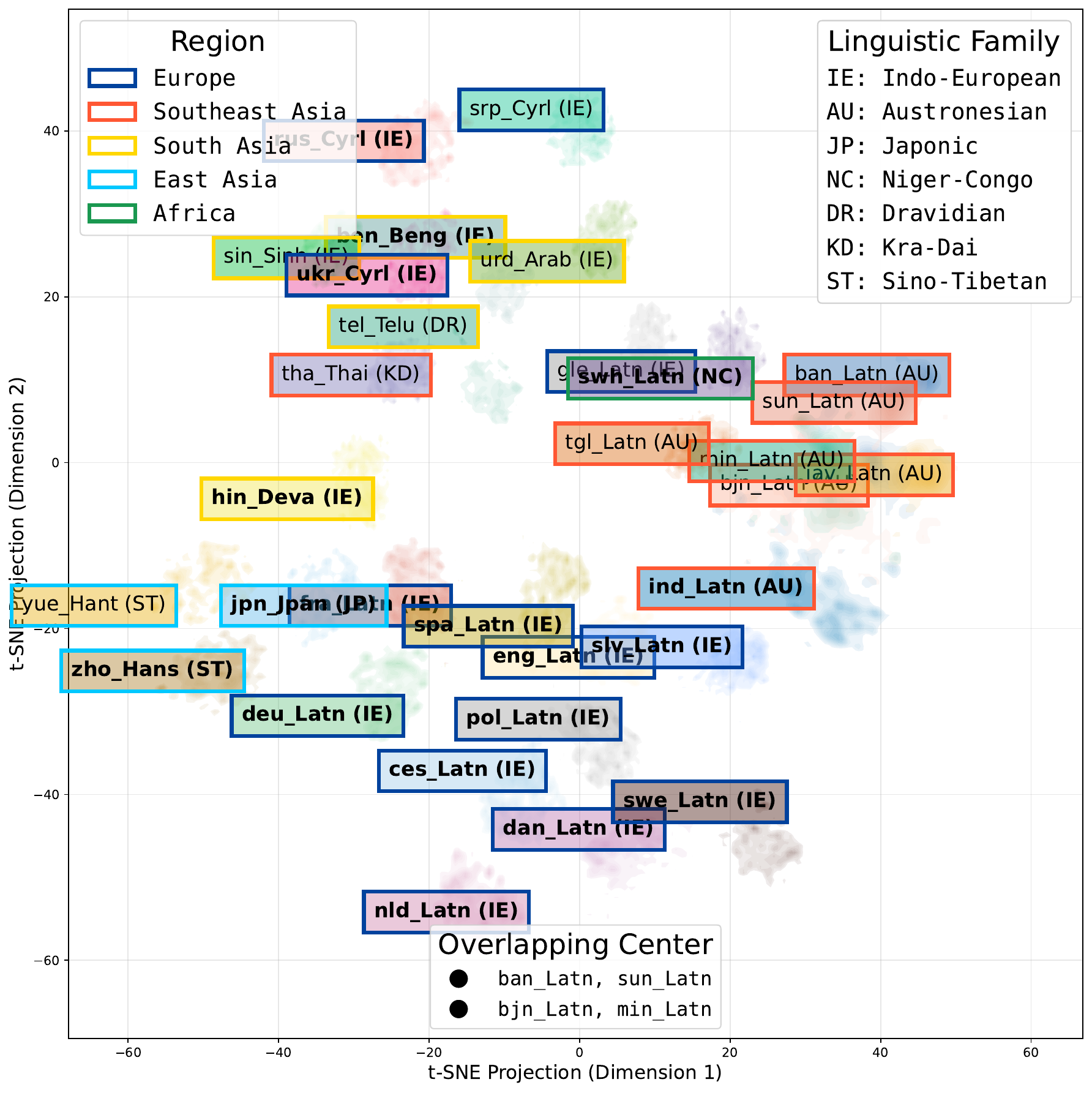}
      \caption{Late (layer 32), perplexity = 15}
  \end{subfigure}
  \caption{Hidden-state embeddings of Aya Expanse (8B) projected in t-SNE dimensions, with HRLs in \textbf{bold}. The t-SNE visualizations are derived using the perplexity value of 15.}
  \label{fig:abl_ppl15}
\end{figure}

\begin{figure}[!t]
  \centering
  \begin{subfigure}[t]{0.85\columnwidth}
      \includegraphics[clip, width=\columnwidth]{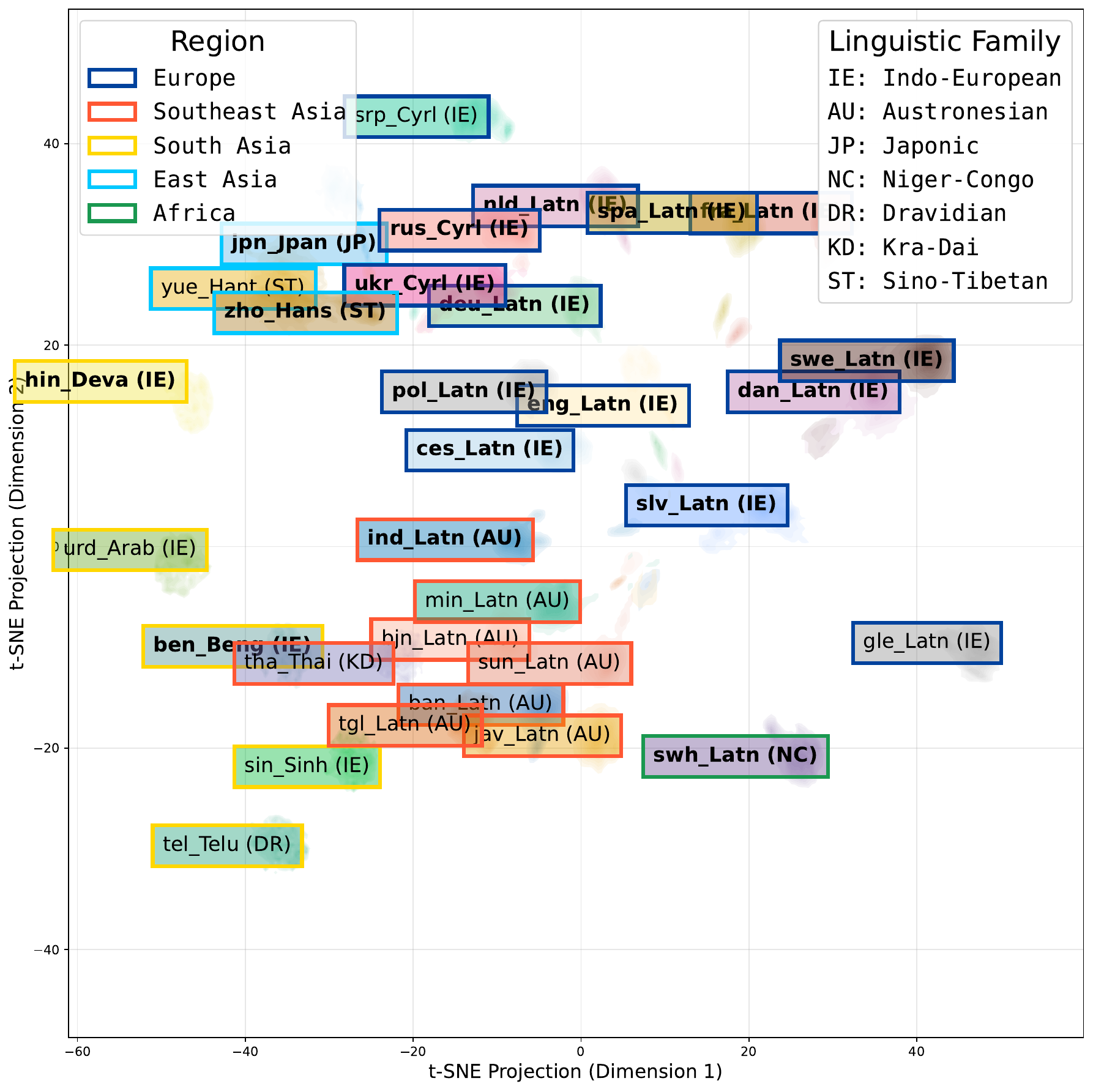}
      \caption{Early (layer 0), perplexity = 30}
  \end{subfigure}
  \begin{subfigure}[t]{0.85\columnwidth}
      \includegraphics[clip, width=\columnwidth]{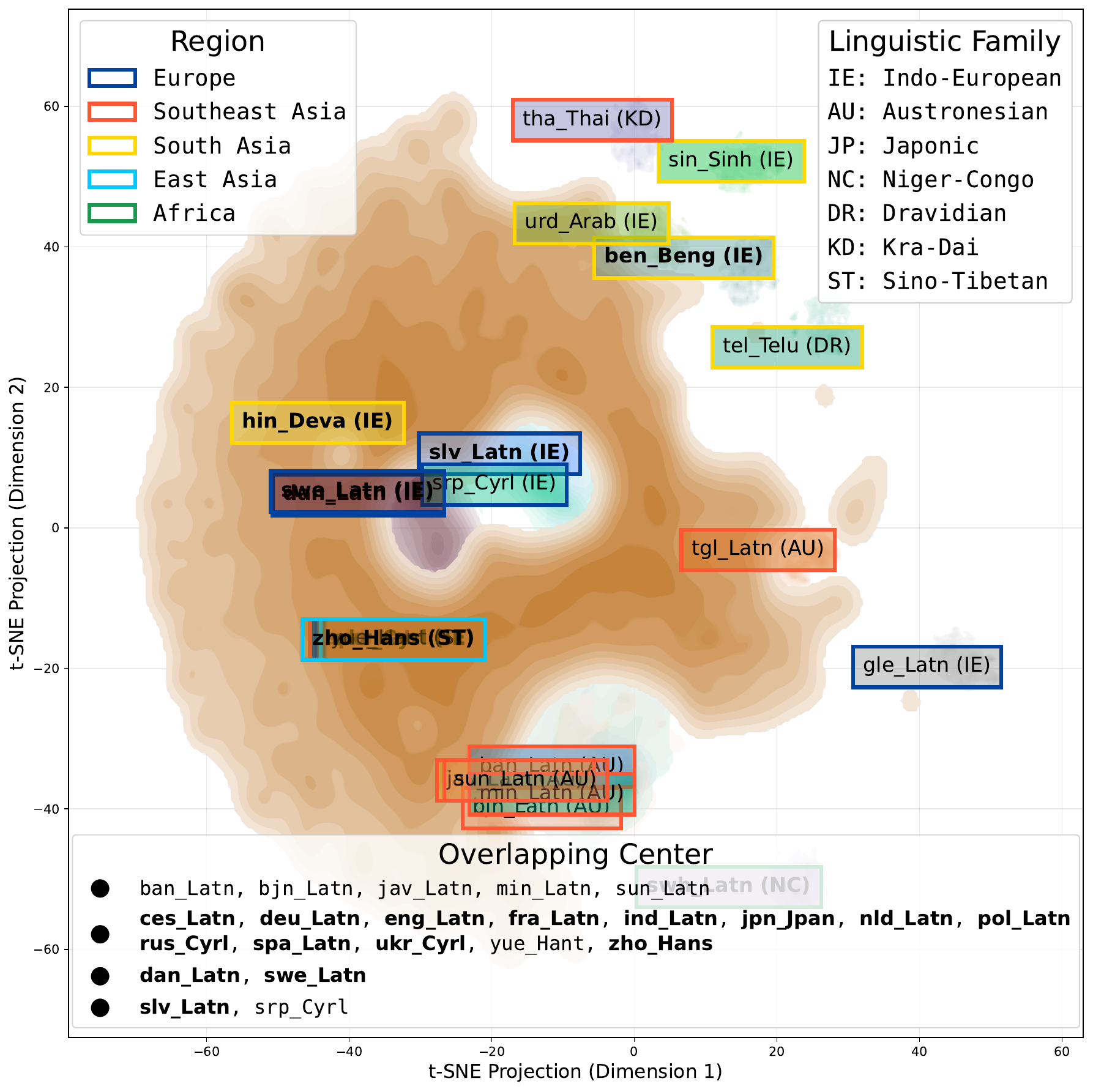}
      \caption{Intermediate (layer 16), perplexity = 30}
  \end{subfigure}
  \begin{subfigure}[t]{0.85\columnwidth}
    \centering
      \includegraphics[clip, width=\columnwidth]{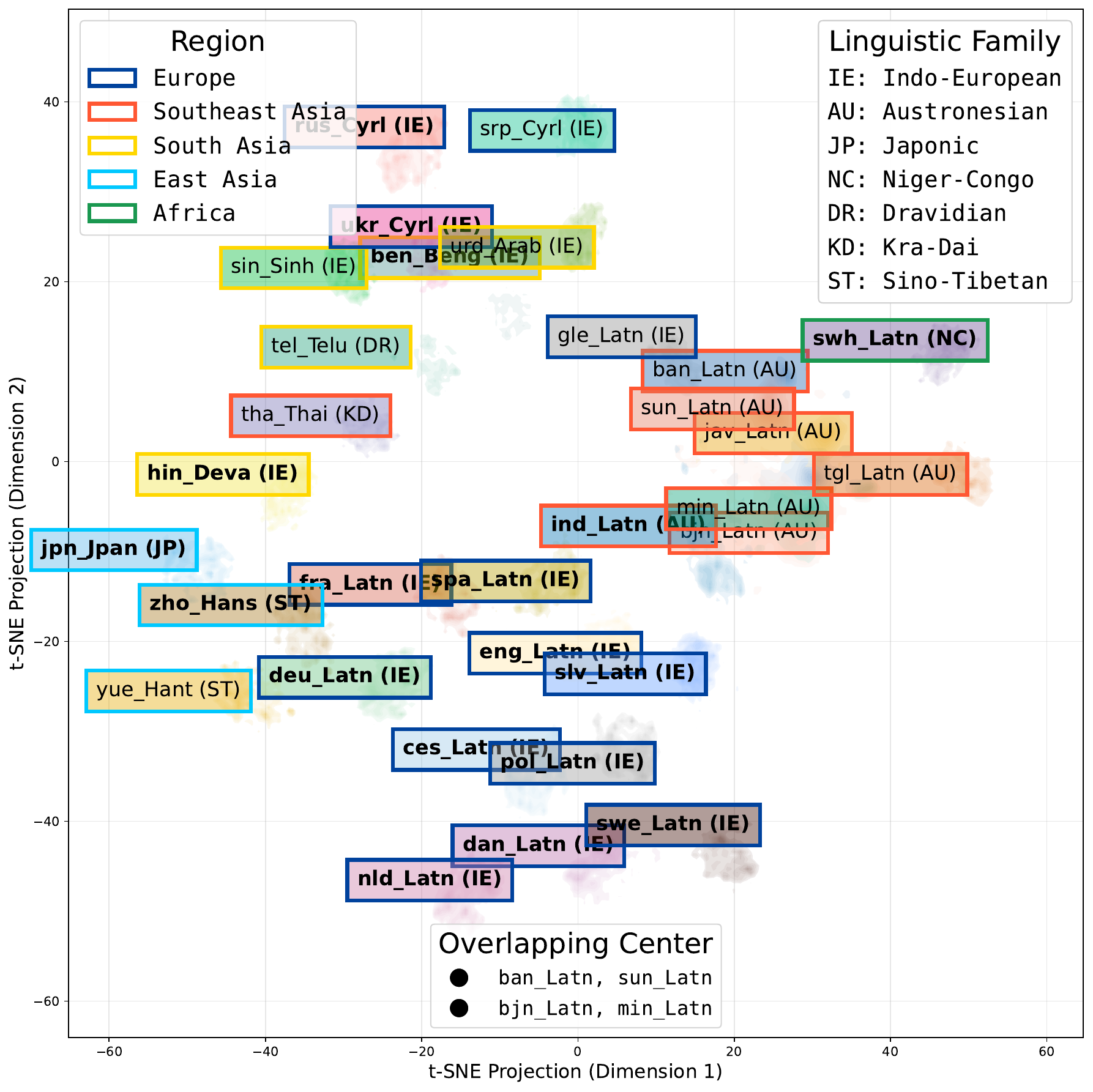}
      \caption{Late (layer 32), perplexity = 30}
  \end{subfigure}
  \caption{Hidden-state embeddings of Aya Expanse (8B) projected in t-SNE dimensions, with HRLs in \textbf{bold}. The t-SNE visualizations are derived using the perplexity value of 30.}
  \label{fig:abl_ppl30}
\end{figure}

\begin{figure}[!t]
  \centering
  \begin{subfigure}[t]{0.85\columnwidth}
      \includegraphics[clip, width=\columnwidth]{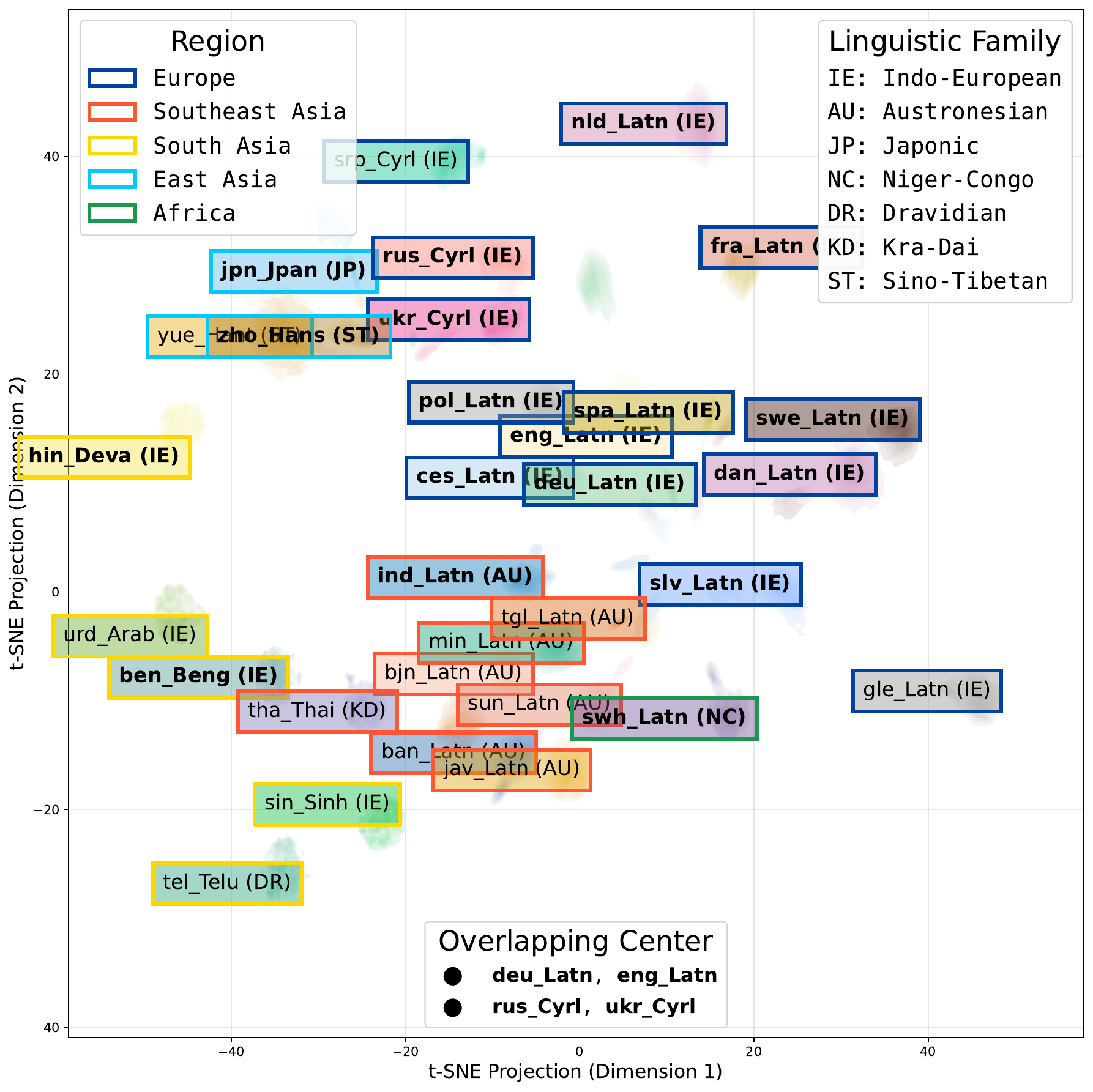}
      \caption{Early (layer 0), perplexity = 50}
  \end{subfigure}
  \begin{subfigure}[t]{0.85\columnwidth}
      \includegraphics[clip, width=\columnwidth]{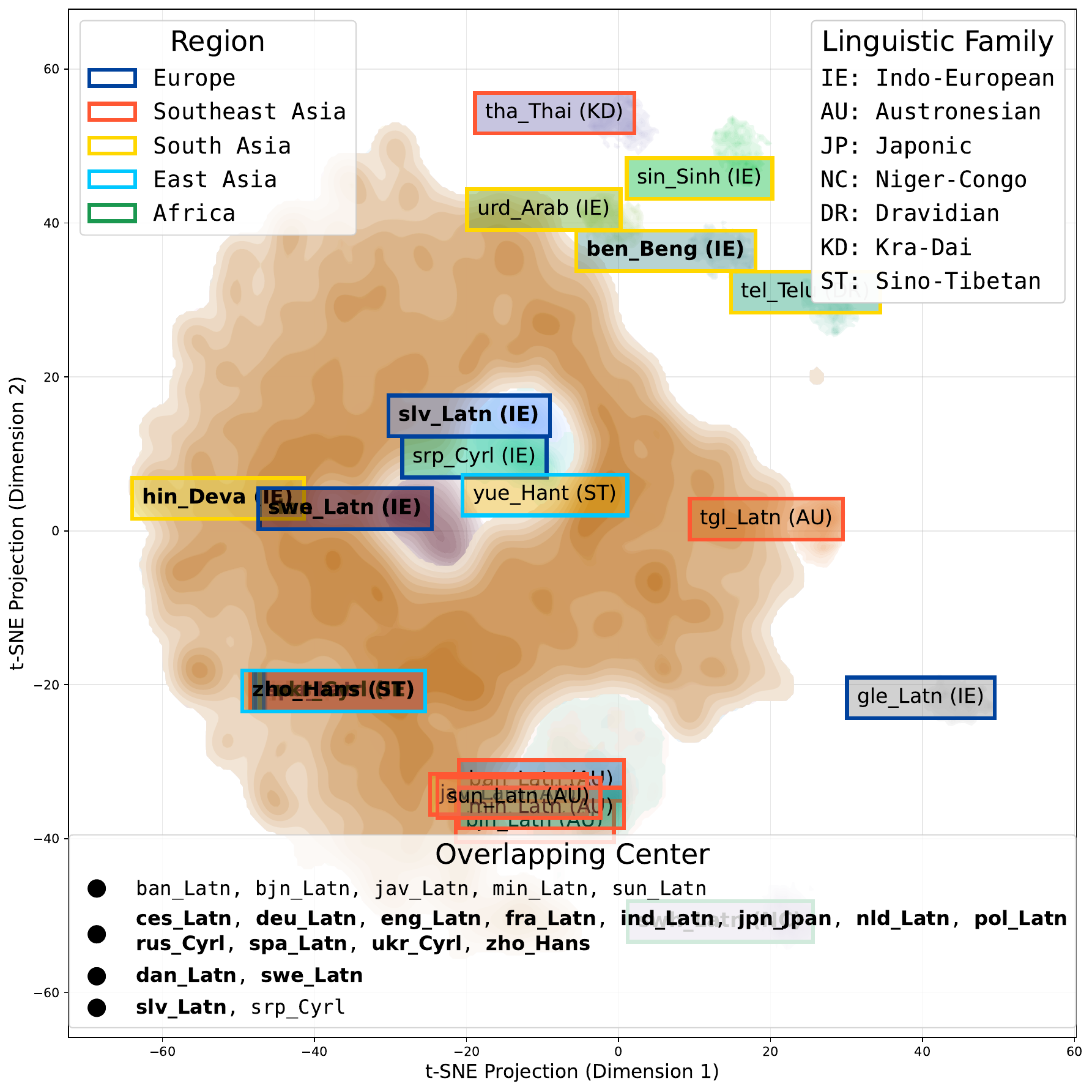}
      \caption{Intermediate (layer 16), perplexity = 50}
  \end{subfigure}
  \begin{subfigure}[t]{0.85\columnwidth}
    \centering
      \includegraphics[clip, width=\columnwidth]{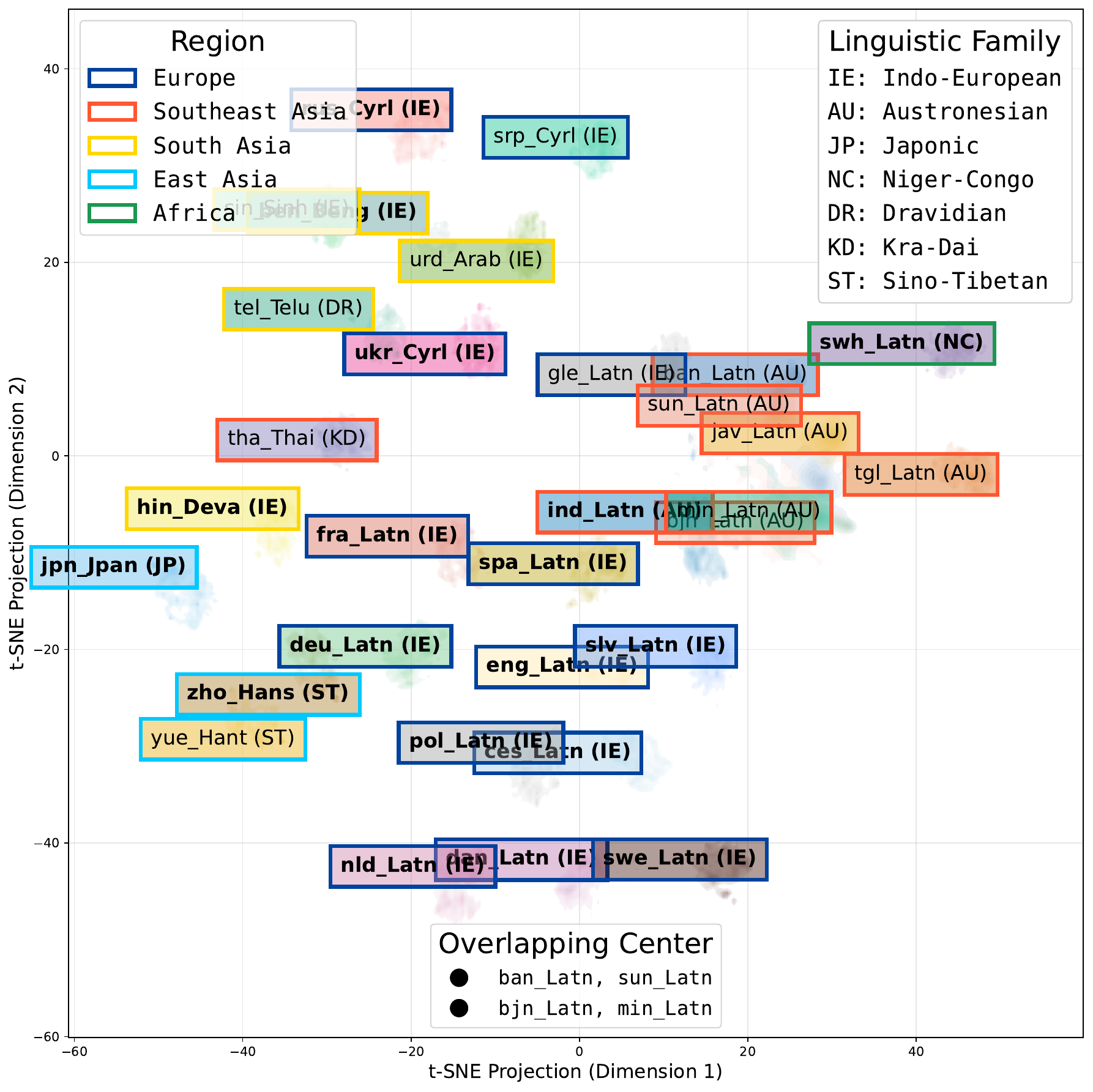}
      \caption{Late (layer 32), perplexity = 50}
  \end{subfigure}
  \caption{Hidden-state embeddings of Aya Expanse (8B) projected in t-SNE dimensions, with HRLs in \textbf{bold}. The t-SNE visualizations are derived using the perplexity value of 50.}
  \label{fig:abl_ppl50}
\end{figure}

We conducted additional t-SNE analysis using perplexity values of 5, 30, and 50, on early, middle, and late layers of Aya Expanse (8B), and visualize them in Figures~\ref{fig:abl_ppl5},~\ref{fig:abl_ppl15},~\ref{fig:abl_ppl30}, and~\ref{fig:abl_ppl50}. Throughout the various perplexity settings, we similarly observe that in the early and late layers, language representations exhibit a minimal overlap, while they cluster according to resource levels and linguistic features. There are different overlaps in the early layer, between Germany and English instead of Japanese and Chinese, when the perplexity is set to 50; additional overlaps between pairs of Bengali, Sinhala, and Czech, Polish in the late layer, with the perplexity set to 5; and no overlap at all in the early layer when the perplexity is set to 30. We also observed similar interlingual overlaps in the intermediate layer that mainly involve high-resource languages with some representations consistently remaining fragmented outside these overlaps, and that low-resource languages overlap due to regional factors. The same set of languages overlaps, with minor differences: the languages of Danish, Swedish, and Ukrainian are added to the overlap with the perplexity set to 5, 30, and 50, and with Yue Chinese missing in the overlaps when the perplexity is set to 50.

These observations substantiate the findings that the interlingual overlapping patterns remain consistent in all cases regardless of the perplexity values used. These additional analyses reinforce the notion that these representational patterns are inherent to the model’s learned structure rather than artifacts of a specific t-SNE configuration.

\subsection{$k$-NN parameters of the ILO score}
\label{sec:appendix_knn}

\begin{figure*}[!t]
  \centering
  \begin{subfigure}[t]{\linewidth}
    \centering
      \includegraphics[clip, width=0.85\linewidth]{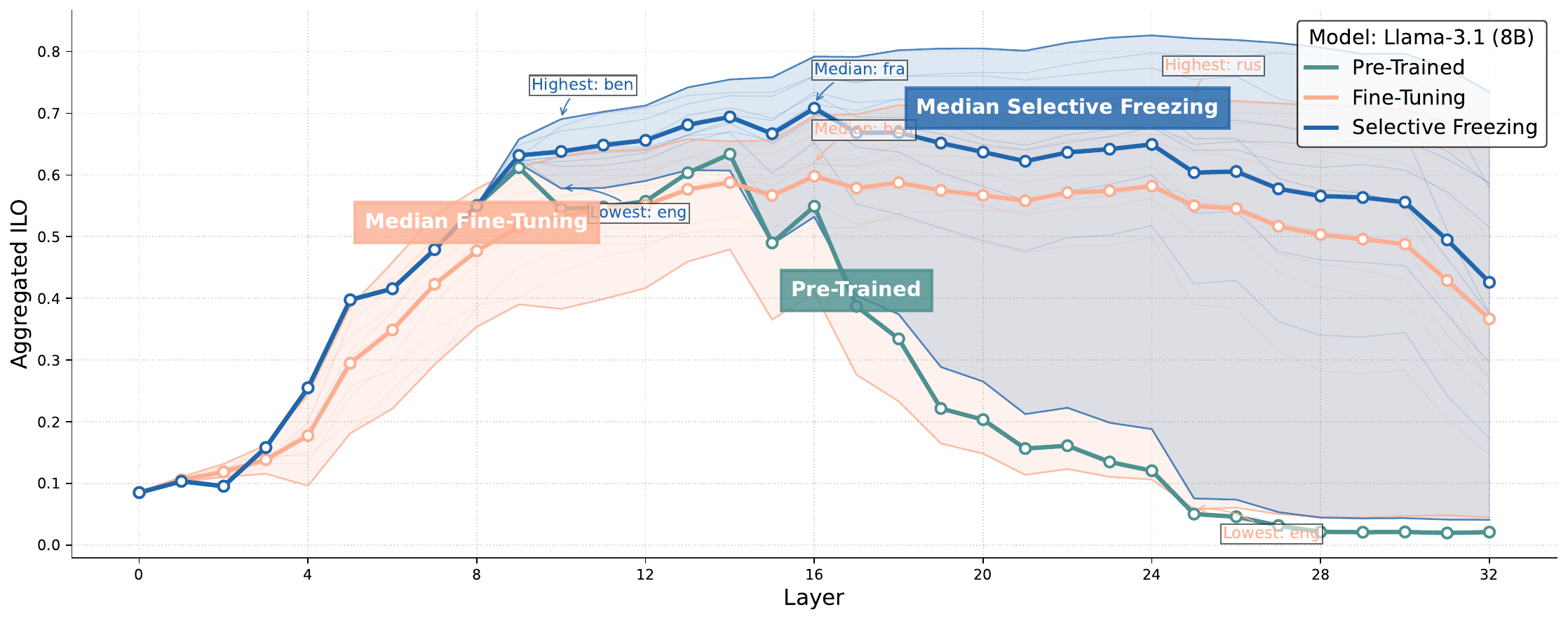}
      \vspace{-5pt}
      \caption{ILO scores are derived using $k=5$, $\tau$ = 3, and cosine distance metric}
      \vspace{-2pt}
  \end{subfigure}
  \begin{subfigure}[t]{\linewidth}
    \centering
      \includegraphics[clip, width=0.85\linewidth]{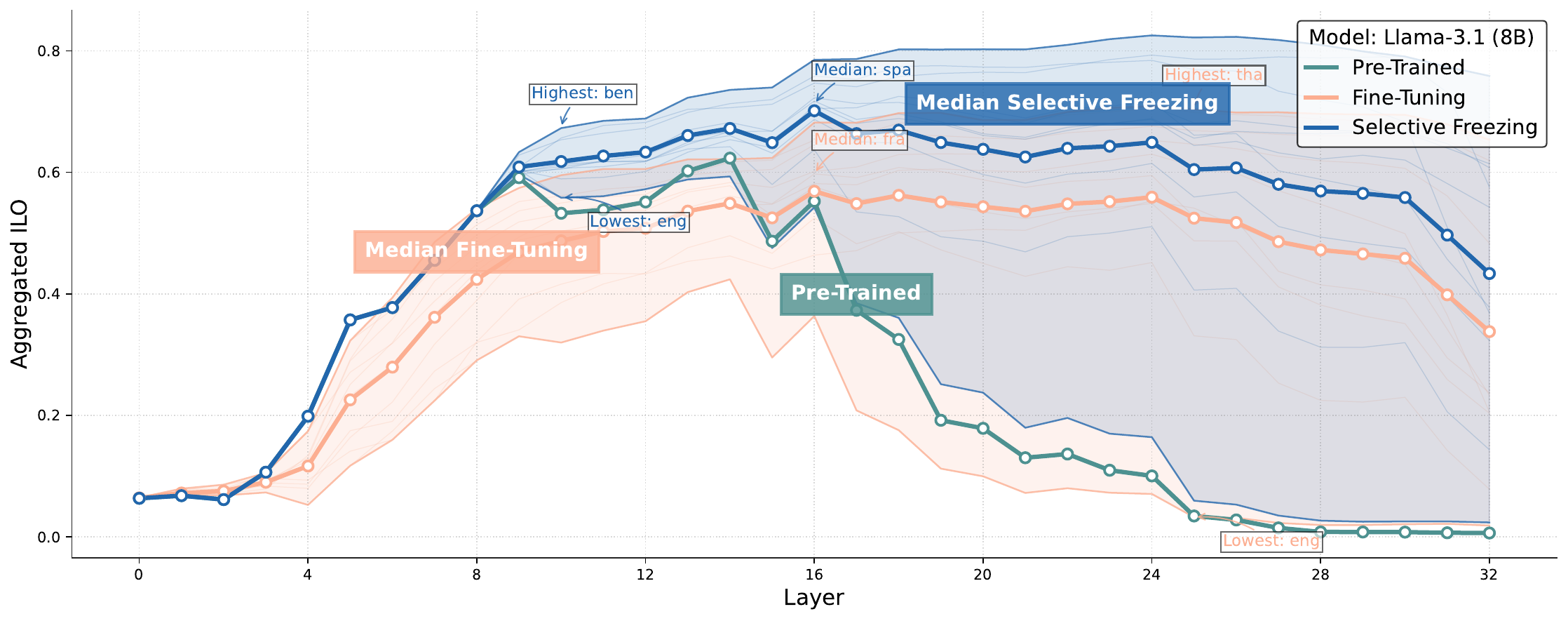}
      \vspace{-5pt}
      \caption{ILO scores are derived using $k=10$, $\tau$ = 5, and cosine distance metric}
      \vspace{-2pt}
  \end{subfigure}
  \begin{subfigure}[t]{\linewidth}
    \centering
      \includegraphics[clip, width=0.85\linewidth]{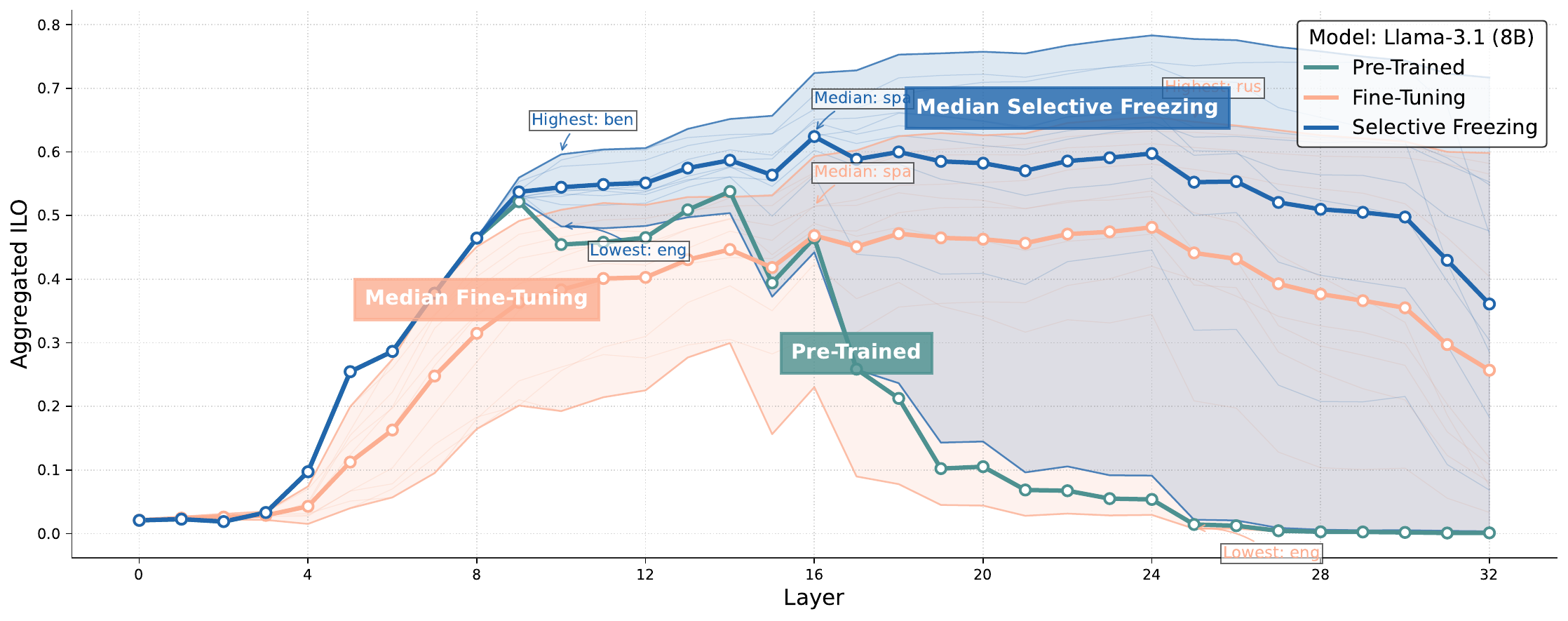}
      \vspace{-5pt}
      \caption{ILO scores are derived using $k=20$, $\tau$ = 10, and cosine distance metric}
      \vspace{-2pt}
  \end{subfigure}
  \vspace{5pt}
  \caption{Layer-wise \(\bar{\operatorname{\abbrvmetric{}}}_{\mathcal{L}}\) scores for all of the source languages in the single-language training on Llama-3.1 (8B) in \textbf{pre-trained}, \textbf{fine-tuning}, and \textbf{selective freezing} modes, with freezing the first 8 layers. Here, the ILO scores derived using cosine distance metric with variations of the $k$-NN parameters.
  }
  \label{fig:abl_knn_cosine}
\end{figure*}

\begin{figure*}[!t]
  \centering
  \begin{subfigure}[t]{\linewidth}
    \centering
      \includegraphics[clip, width=0.85\linewidth]{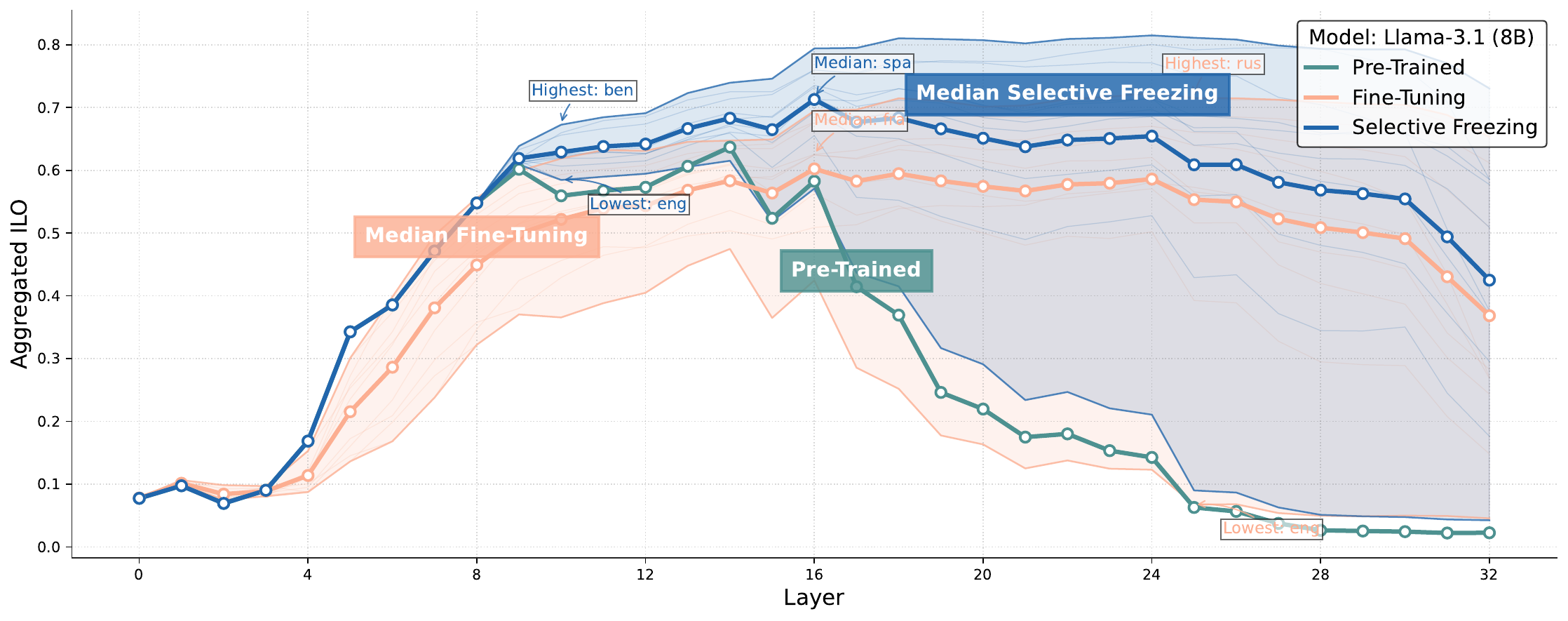}
      \vspace{-5pt}
      \caption{ILO scores are derived using $k=5$, $\tau$ = 3, and Euclidean distance metric}
      \vspace{-2pt}
  \end{subfigure}
  \begin{subfigure}[t]{\linewidth}
    \centering
      \includegraphics[clip, width=0.85\linewidth]{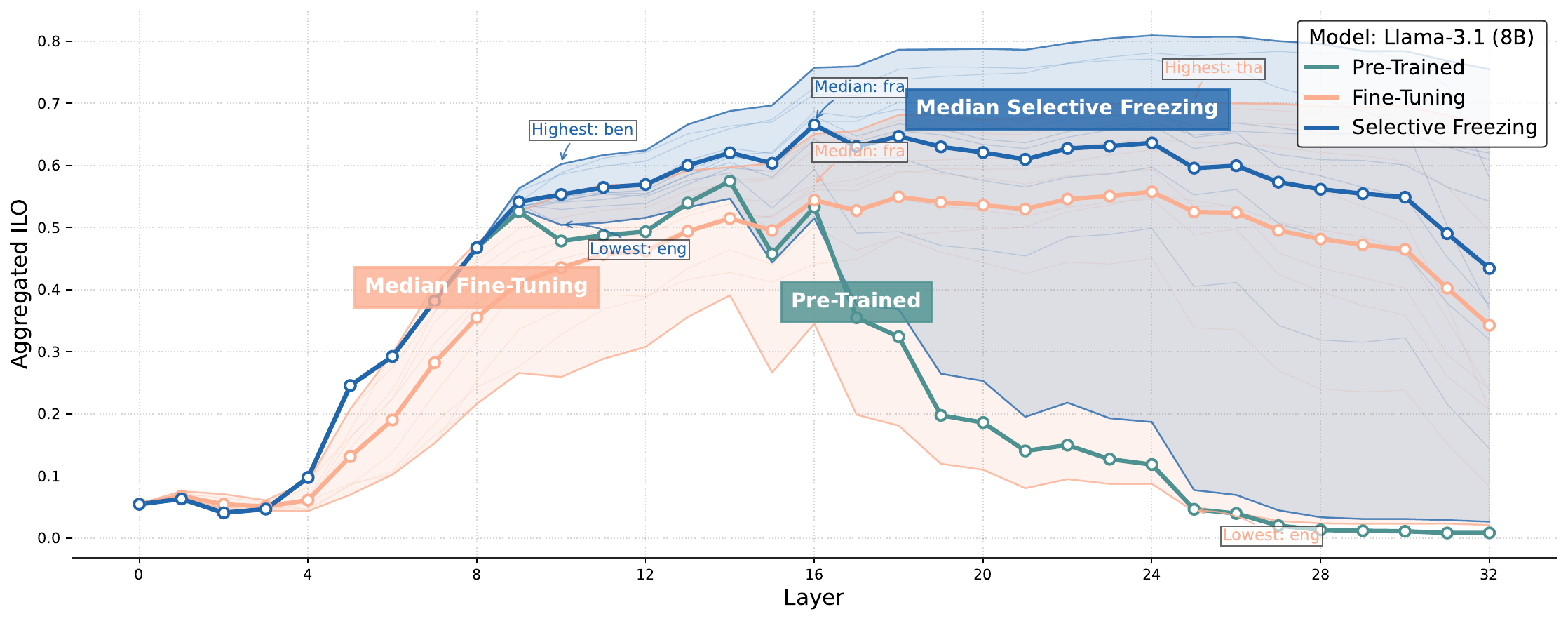}
      \vspace{-5pt}
      \caption{ILO scores are derived using $k=10$, $\tau$ = 5, and Euclidean distance metric}
      \vspace{-2pt}
  \end{subfigure}
  \begin{subfigure}[t]{\linewidth}
    \centering
      \includegraphics[clip, width=0.85\linewidth]{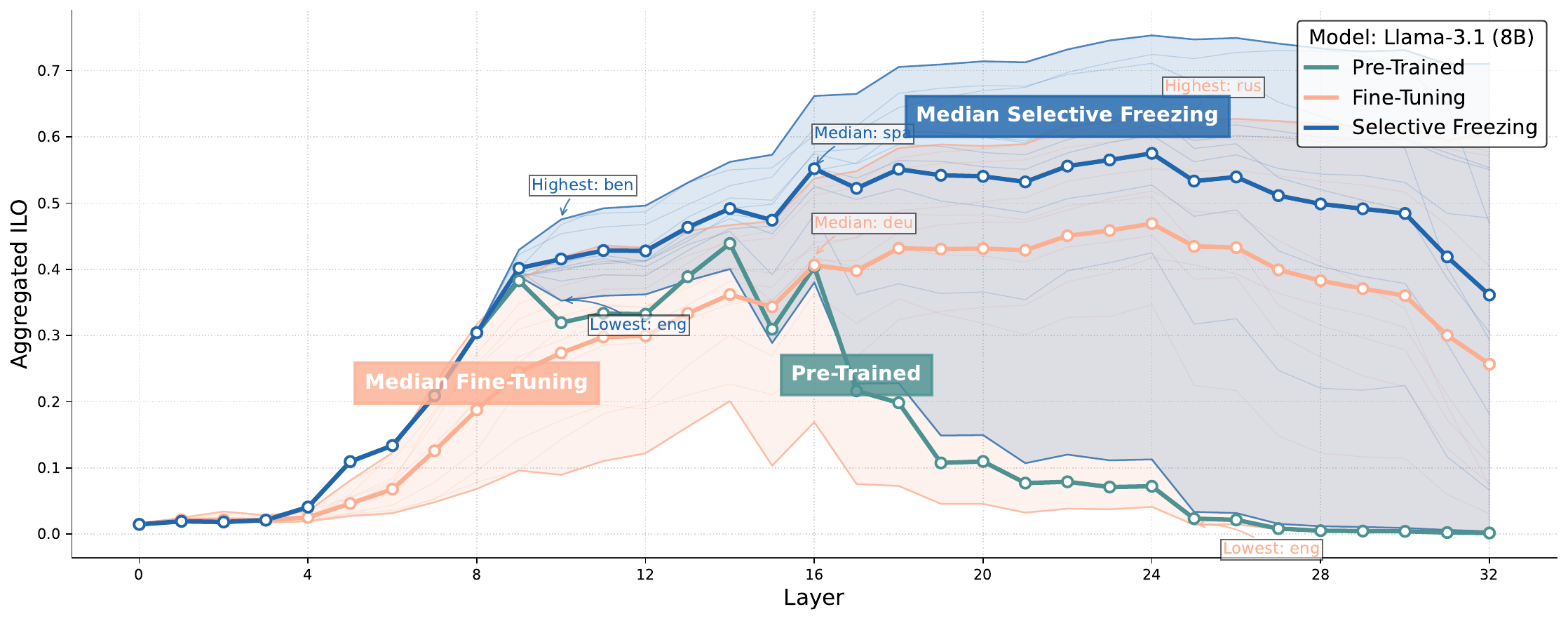}
      \vspace{-5pt}
      \caption{ILO scores are derived using $k=20$, $\tau$ = 10, and Euclidean distance metric}
      \vspace{-2pt}
  \end{subfigure}
  \vspace{5pt}
  \caption{Layer-wise \(\bar{\operatorname{\abbrvmetric{}}}_{\mathcal{L}}\) scores for all of the source languages in the single-language training on Llama-3.1 (8B) in \textbf{pre-trained}, \textbf{fine-tuning}, and \textbf{selective freezing} modes, with freezing the first 8 layers. Here, the ILO scores derived using Euclidean distance metric with variations of the $k$-NN parameters.
  }
  \label{fig:abl_knn_euclidean}
\end{figure*}

We further conducted ablation studies over different settings of $k$ and $\tau$—specifically, [(5,3), (10,5), (20,10)]—using both cosine and Euclidean distances. We report the results in Table~\ref{fig:abl_knn_cosine} and~\ref{fig:abl_knn_euclidean}. Our results indicate that a lower $k$ ($k$ = 5, $\tau$ = 3) leads to a modest increase in the overall aggregated ILO across all layers by about 0.03–0.05, whereas a higher $k$ ($k$ = 20, $\tau$ = 10) results in a reduction of roughly 0.1–0.15 relative to our main illustration in Figure 5. Nonetheless, we find that all the trends remain consistent with our findings. When ablating a different distance metric, i.e, cosine distance, we find that the influence of varying $k$ values is slightly less pronounced, with the aggregated ILO scores remaining within a similar range.

In summary, despite the different selection of the $k$-NN parameters and distance metric, observations using ILO score consistently highlight similar trend on the decrease of alignment degree in the same layers, and that the model trained with the selective-freezing mechanism sustains their prior semantic alignment levels in all layers.

\subsection{Layer selection for selective freezing}
\label{sec:appendix_layer}

\begin{figure*}[!t]
  \centering
  \begin{subfigure}[t]{\linewidth}
    \centering
      \includegraphics[clip, width=0.82\linewidth]{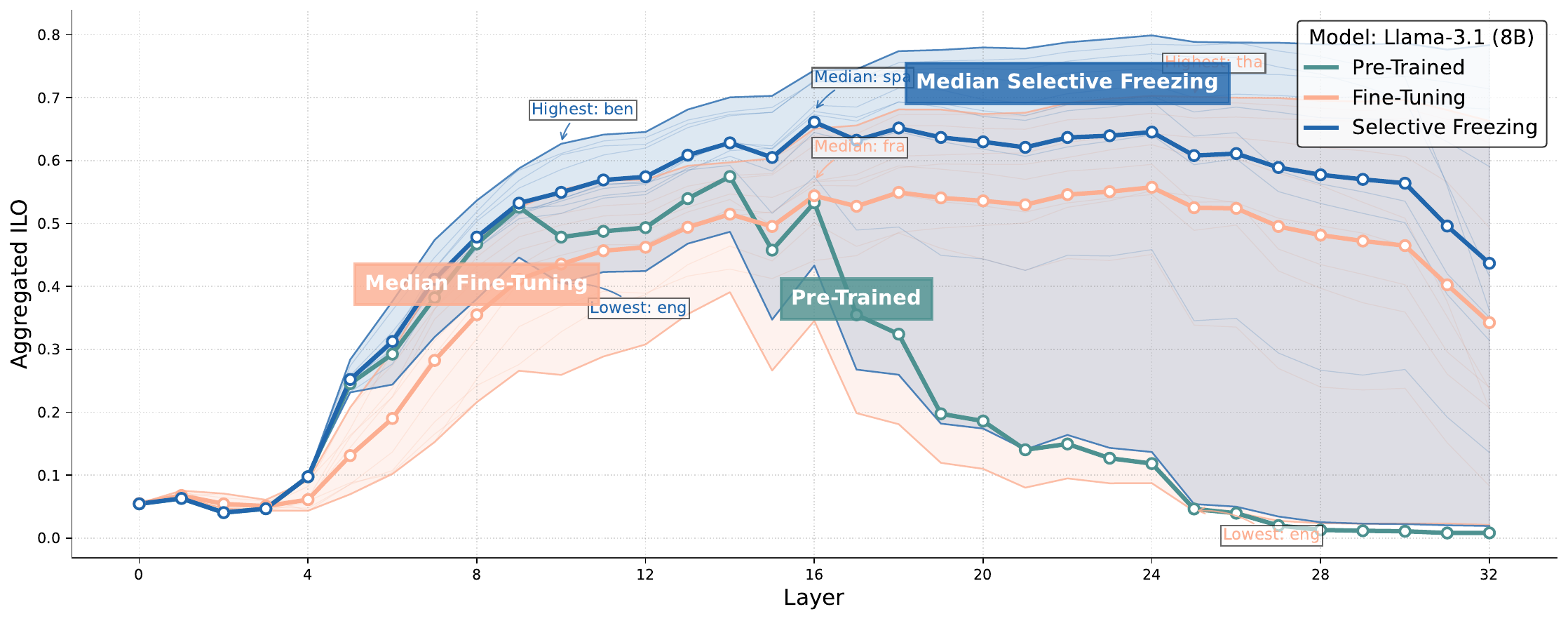}
      \vspace{-5pt}
      \caption{Fine-tuned with the first 4 layers being frozen}
      \vspace{-2pt}
  \end{subfigure}
  \begin{subfigure}[t]{\linewidth}
    \centering
      \includegraphics[clip, width=0.82\linewidth]{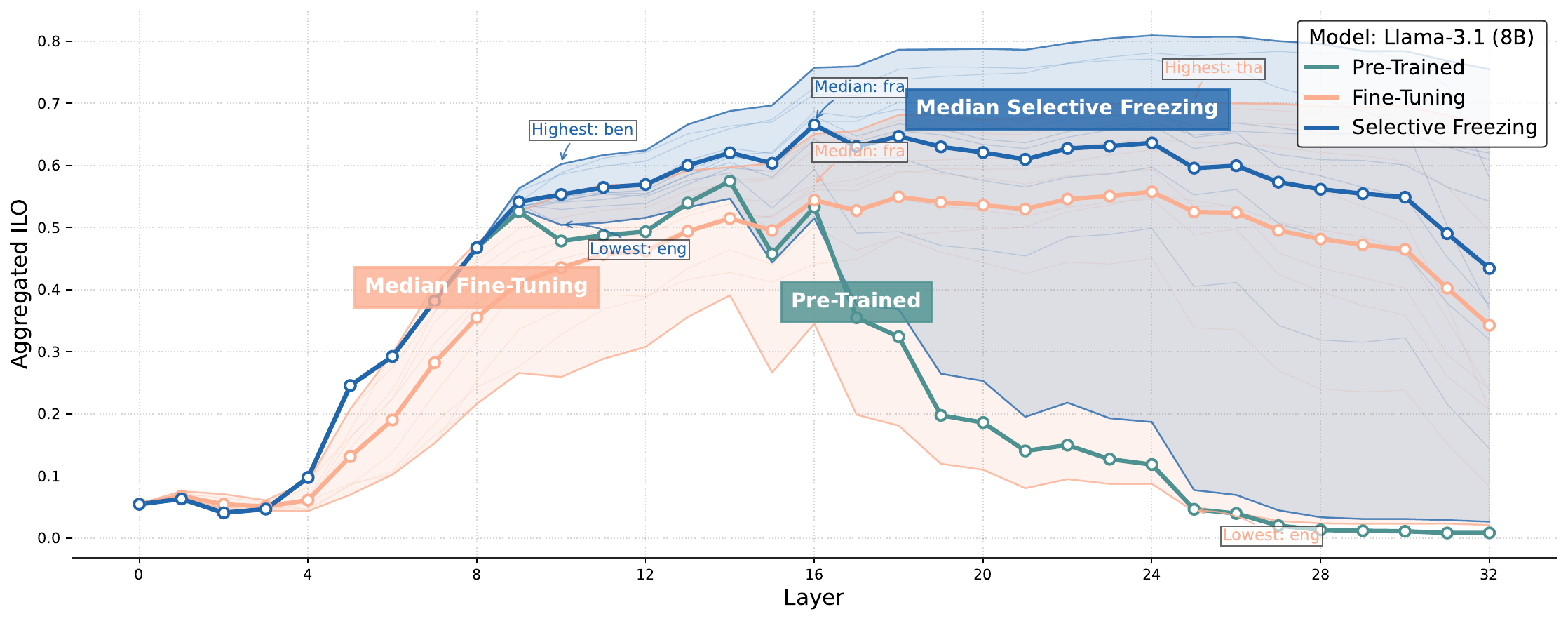}
      \vspace{-5pt}
      \caption{Fine-tuned with the first 8 layers being frozen}
      \vspace{-2pt}
  \end{subfigure}
  \begin{subfigure}[t]{\linewidth}
    \centering
      \includegraphics[clip, width=0.82\linewidth]{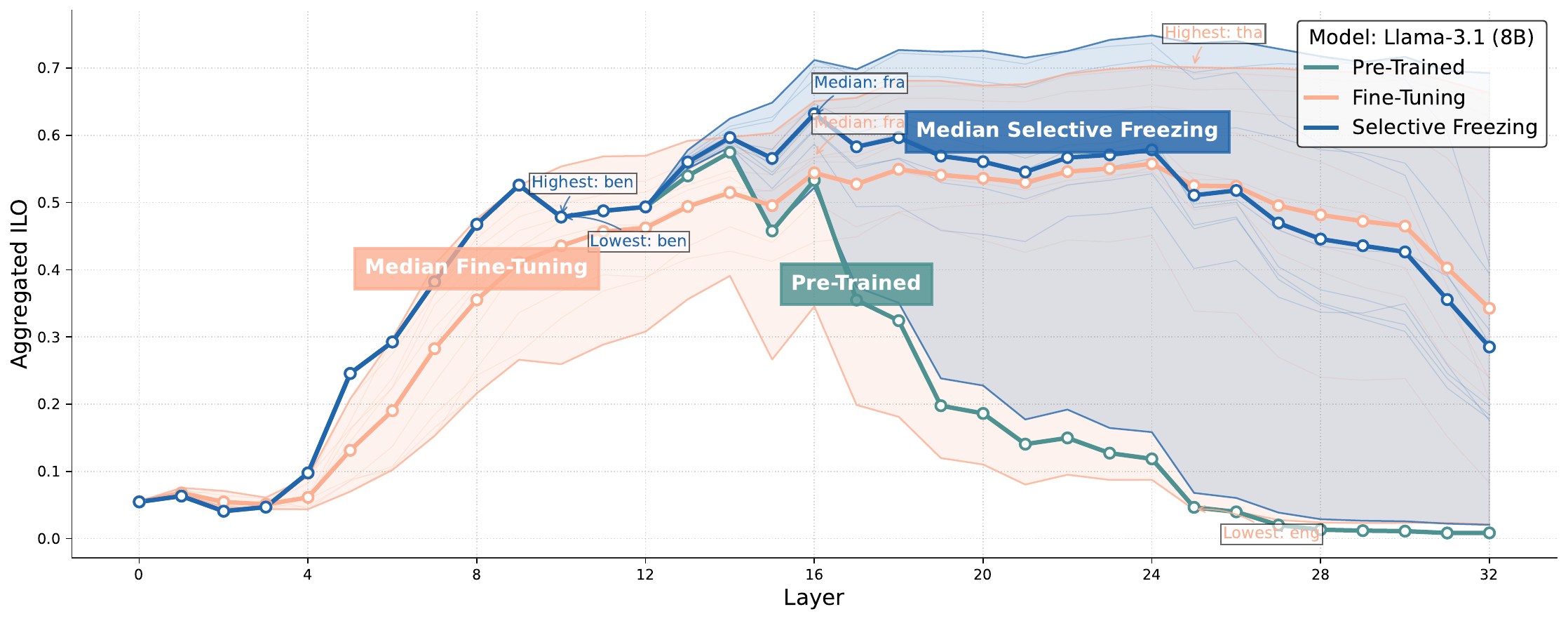}
      \vspace{-5pt}
      \caption{Fine-tuned with the first 12 layers being frozen}
      \vspace{-2pt}
  \end{subfigure}
  \begin{subfigure}[t]{\linewidth}
    \centering
      \includegraphics[clip, width=0.82\linewidth]{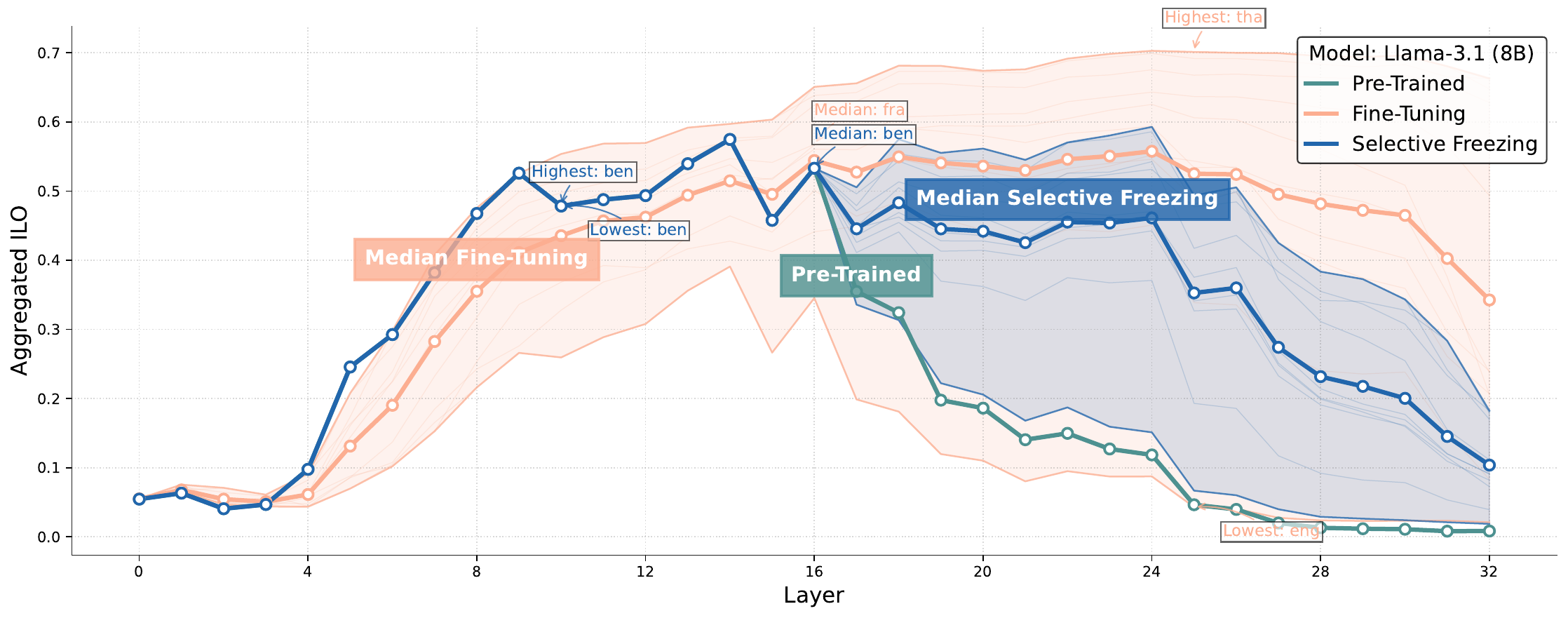}
      \vspace{-5pt}
      \caption{Fine-tuned with the first 16 layers being frozen}
      \vspace{-2pt}
  \end{subfigure}
  \caption{Ablation study on frozen layer selections, analyzed through layer-wise \(\bar{\operatorname{\abbrvmetric{}}}_{\mathcal{L}}\) scores for all of the source-languages in the single-language training on Llama-3.1 (8B) in \textbf{pre-trained}, \textbf{fine-tuning}, and \textbf{selective freezing} modes. Decrease in alignment from single-language \textbf{fine-tuning} is seen in the early layers. On the contrary, freezing the first 4, 8, and 12 layers maintains and improves the semantic alignment across layers. However, while freezing the first 16 layers preserves alignment in the frozen layers, the subsequent layers exhibit lower alignments compared to the fine-tuned models.
  }
  \label{fig:abl_layer_freeze_ilo}
\end{figure*}

We perform experiments on selective freezing of the first 4, 8, 12, and 16 layers of Llama-3.1 (8B). Our motivation stems from prior works that have demonstrated that multilingual language models tend to align their representations in the early layers~\cite{muller2021first, zhao2024how}, which guided our decision to focus on these layers. We denote the aggregated results in Table~\ref{tab:abl_layer_freeze_agg}, the complete results in Table~\ref{tab:abl_layer_freeze_complete}, and visualize the aggregated ILO scores in Figures~\ref{fig:abl_layer_freeze_ilo}. In general, fine-tuning with freezing the early layers enhances the cross-lingual generalization. Notably, the best overall performance was achieved when freezing the first 12 layers. Throughout the experiments, analysis of interlingual alignment using ILO reveal that freezing the first 4, 8, and 12 layers maintains and improves the semantic alignment across layers. In contrast, while freezing the first 16 layers preserves alignment in the frozen layers, the subsequent layers exhibit lower alignments compared to the fine-tuned models. 

Furthermore, across all settings, we observed improved transfer on languages within families and regions, with negligible degradation—and sometimes even improvements—in low-resource, cross-family, and cross-regional scenarios. When comparing the trade-offs between freezing the first 8 layers versus the first 12 layers, we found that the performance gain in the source language is mixed. In the latter setting, the task performances in languages such as English, Russian, French, German, and Bengali improved, while in Spanish, Chinese, Japanese, Swahili, and Thai, they instead decreased. Moreover, the multilingual performance from fine-tuning with English mostly dropped, except for certain gains in English, Russian, Thai, and Bengali. Lastly, the aggregate multilingual performance when freezing the first 16 layers is closer to that of fine-tuning, showcasing the impact of lower interlingual alignment previously indicated from the observation on the analysis using ILO.

\begin{table}[]
\centering
\resizebox{0.8\linewidth}{!}{%
\begin{tabular}{lcrr}
\toprule
\multirow{2}{*}{\textbf{Method}} & \multirow{2}{*}{\textbf{Frozen Layers}} & \multicolumn{2}{c}{\textbf{Average}} \\ \cmidrule(lr){3-4}
 &  & \multicolumn{1}{c}{\textbf{All}} & \multicolumn{1}{c}{\textbf{XL}} \\ \midrule
Fine-tuning & - & 17.7\% & 16.0\% \\ \midrule
\multirow{2}{*}{\shortstack[l]{Selective \\ Freezing}} & First 4 & 21.6\% & 20.2\% \\
 & First 8 & 22.4\% & 21.2\% \\
 & First 12 & \textbf{23.1\%} & \textbf{22.1\%} \\
 & First 16 & 19.0\% & 18.0\%  \\ \bottomrule
\end{tabular}}
\caption{Aggregated results on the ablation study on the cross-lingual transfer performance on MGSM for Llama-3.1 (8B) fine-tuned with the selective freezing strategy varied on the frozen layers. Freezing the first 4, 8, 12, and 16 layers enhanced the cross-lingual generalization, with the best performance achieved when freezing the first 12 layers.}
\label{tab:abl_layer_freeze_agg}
\vspace{-10pt}
\end{table}

\begin{table*}[!ht]
\centering
\resizebox{0.94\linewidth}{!}{%
\begin{tabular}{l|c|rrrrrrrrrrr|rr}
\toprule
& & \multicolumn{11}{c}{\textbf{Accuracy}} & \multicolumn{2}{|c}{\textbf{Average}} \\
\cmidrule(lr){3-13}\cmidrule(lr){14-15}
\multirow{-2.3}{*}{\textbf{Method}} & \multirow{-2.2}{*}{\shortstack{\textbf{Training} \\ \textbf{languages}}} & \multicolumn{1}{c}{\textbf{ben}} & \multicolumn{1}{c}{\textbf{tha*}} & \multicolumn{1}{c}{\textbf{swh}} & \multicolumn{1}{c}{\textbf{tel*}} & \multicolumn{1}{c}{\textbf{jpn}} & \multicolumn{1}{c}{\textbf{zho}} & \multicolumn{1}{c}{\textbf{deu}} & \multicolumn{1}{c}{\textbf{fra}} & \multicolumn{1}{c}{\textbf{rus}} & \multicolumn{1}{c}{\textbf{spa}} & \multicolumn{1}{c}{\textbf{eng}} & \multicolumn{1}{|c}{\textbf{All}} & \multicolumn{1}{c}{\textbf{XL}} \\ \midrule
Pre-trained & mixed & 11.6\% & 12.0\% & 7.2\% & 0.0\% & 10.4\% & 8.8\% & 16.0\% & 12.4\% & 14.0\% & 11.6\% & 17.6\% & 10.3\% & \multicolumn{1}{c}{-} \\ \midrule
 & ben & \cellcolor[HTML]{00FFFF}\textbf{23.2\%} & \cellcolor[HTML]{EA9999}4.8\% & \cellcolor[HTML]{EA9999}1.2\% & 3.2\% & \cellcolor[HTML]{EA9999}10.0\% & 9.6\% & \cellcolor[HTML]{EA9999}10.8\% & 13.6\% & \cellcolor[HTML]{EA9999}11.6\% & 14.8\% & \cellcolor[HTML]{EA9999}12.8\% & 10.5\% & 9.2\% \\
 & tha* & \cellcolor[HTML]{EA9999}1.6\% & \cellcolor[HTML]{00FFFF}\textbf{32.8\%} & \cellcolor[HTML]{EA9999}4.4\% & 1.6\% & 14.4\% & 14.8\% & 17.2\% & 19.2\% & 18.0\% & 20.4\% & 25.6\% & 15.5\% & 13.7\% \\
 & swh & \cellcolor[HTML]{EA9999}3.2\% & \cellcolor[HTML]{EA9999}6.4\% & \cellcolor[HTML]{00FFFF}\textbf{30.8\%} & 2.8\% & 11.2\% & 12.4\% & 20.4\% & 19.6\% & 14.8\% & 22.4\% & 26.8\% & 15.5\% & 14.0\% \\
 & jpn & \cellcolor[HTML]{EA9999}3.6\% & \cellcolor[HTML]{EA9999}7.2\% & \cellcolor[HTML]{EA9999}2.8\% & 1.2\% & \cellcolor[HTML]{00FFFF}\textbf{32.8\%} & 21.6\% & 19.6\% & 18.0\% & 18.4\% & 22.4\% & 28.8\% & 16.0\% & 14.4\% \\
 & zho & \cellcolor[HTML]{EA9999}0.8\% & \cellcolor[HTML]{EA9999}7.2\% & \cellcolor[HTML]{EA9999}2.4\% & 1.6\% & 22.0\% & \cellcolor[HTML]{00FFFF}\textbf{34.8\%} & 19.6\% & 19.6\% & 21.6\% & 21.2\% & 27.6\% & 16.2\% & 14.4\% \\
 & deu & \cellcolor[HTML]{EA9999}8.0\% & 16.4\% & 8.0\% & \textbf{4.0\%} & 19.2\% & 19.6\% & \cellcolor[HTML]{00FFFF}\textbf{37.6\%} & \textbf{34.4\%} & 23.6\% & 28.8\% & 36.4\% & 21.5\% & 19.8\% \\
 & fra & \cellcolor[HTML]{EA9999}4.8\% & \cellcolor[HTML]{EA9999}11.6\% & \cellcolor[HTML]{EA9999}4.0\% & 3.2\% & 16.0\% & 16.8\% & 31.6\% & \cellcolor[HTML]{00FFFF}\textbf{34.4\%} & 25.6\% & 34.4\% & 35.6\% & 19.8\% & 18.4\% \\
 & rus & \cellcolor[HTML]{EA9999}4.0\% & 14.0\% & \cellcolor[HTML]{EA9999}4.0\% & 1.2\% & 17.2\% & 16.4\% & 29.6\% & 28.4\% & \cellcolor[HTML]{00FFFF}\textbf{34.0\%} & 30.0\% & 26.4\% & 18.7\% & 17.1\% \\
 & spa & \cellcolor[HTML]{EA9999}4.8\% & 16.0\% & \cellcolor[HTML]{EA9999}2.8\% & 2.4\% & 14.4\% & 19.6\% & 28.4\% & 30.8\% & 31.2\% & \cellcolor[HTML]{00FFFF}\textbf{38.4\%} & 38.4\% & 20.7\% & 18.9\% \\
\multirow{-10}{*}{Fine-tuning} & eng & \cellcolor[HTML]{EA9999}6.4\% & 14.4\% & \cellcolor[HTML]{EA9999}6.0\% & 2.4\% & 18.8\% & 24.4\% & 37.2\% & 27.2\% & 33.6\% & 33.2\% & \cellcolor[HTML]{00FFFF}\textbf{43.2\%} & \textbf{22.4\%} & \textbf{20.4\%} \\ \midrule
 & ben & \cellcolor[HTML]{00FFFF}\textbf{22.8\%} & \cellcolor[HTML]{EA9999}8.8\% & 8.0\% & 12.4\% & 14.8\% & 10.0\% & \cellcolor[HTML]{EA9999}12.0\% & 12.4\% & 14.4\% & 16.4\% & \cellcolor[HTML]{EA9999}14.8\% & 13.3\% & 12.4\% \\
 & tha* & \cellcolor[HTML]{EA9999}10.4\% & \cellcolor[HTML]{00FFFF}\textbf{31.2\%} & \cellcolor[HTML]{EA9999}5.2\% & 9.2\% & 17.2\% & 20.0\% & 19.6\% & 18.4\% & 15.2\% & 19.6\% & 28.8\% & 17.7\% & 16.4\% \\
 & swh & \cellcolor[HTML]{EA9999}9.6\% & 15.2\% & \cellcolor[HTML]{00FFFF}{\ul \textbf{38.4\%}} & 11.2\% & 11.6\% & 17.6\% & 26.0\% & 23.2\% & 16.4\% & 28.0\% & 26.4\% & 20.3\% & 18.5\% \\
 & jpn & 14.4\% & 12.0\% & 10.4\% & 11.2\% & \cellcolor[HTML]{00FFFF}{\ul \textbf{36.4\%}} & 24.8\% & 23.2\% & 19.2\% & 24.8\% & 19.6\% & 25.2\% & 20.1\% & 18.5\% \\
 & zho & 11.6\% & 15.6\% & 10.8\% & 6.8\% & 20.0\% & \cellcolor[HTML]{00FFFF}\textbf{36.0\%} & 27.6\% & 26.0\% & 19.6\% & 29.2\% & 29.2\% & 21.1\% & 19.6\% \\
 & deu & 14.8\% & 20.8\% & 10.4\% & 10.4\% & 14.8\% & 19.6\% & \cellcolor[HTML]{00FFFF}{\ul \textbf{38.8\%}} & 32.0\% & 27.2\% & 31.2\% & 38.4\% & 23.5\% & 22.0\% \\
 & fra & 14.8\% & 18.8\% & 8.8\% & 10.0\% & 20.8\% & 23.6\% & 34.4\% & \cellcolor[HTML]{00FFFF}{\ul \textbf{38.0\%}} & 31.6\% & 35.6\% & 37.6\% & 24.9\% & 23.6\% \\
 & rus & 15.6\% & 16.4\% & 10.0\% & 10.4\% & 21.2\% & 20.8\% & 27.6\% & 26.8\% & \cellcolor[HTML]{00FFFF}38.0\% & 26.0\% & 37.6\% & 22.8\% & 21.2\% \\
 & spa & 15.6\% & 17.2\% & 11.2\% & 8.0\% & 20.4\% & 21.6\% & 31.6\% & 32.8\% & 34.4\% & \cellcolor[HTML]{00FFFF}\textbf{38.0\%} & 35.6\% & 24.2\% & 22.8\% \\
\multirow{-10}{*}{\begin{tabular}[c]{@{}l@{}}Selective Freezing \\ First 4 Layers\end{tabular}} & eng & 17.2\% & 25.6\% & 13.2\% & \textbf{13.2\%} & 23.6\% & 28.0\% & 36.0\% & 34.8\% & \textbf{38.8\%} & 36.4\% & \cellcolor[HTML]{00FFFF}\textbf{41.2\%} & \textbf{28.0\%} & \textbf{26.7\%} \\ \midrule
 & ben & \cellcolor[HTML]{00FFFF}\textbf{23.2\%} & \cellcolor[HTML]{EA9999}9.2\% & 8.8\% & 10.0\% & 17.6\% & 11.6\% & 18.0\% & 16.4\% & 17.6\% & 18.4\% & 20.8\% & 15.6\% & 14.8\% \\
 & tha* & 14.0\% & \cellcolor[HTML]{00FFFF}{\ul \textbf{35.2\%}} & 12.4\% & 12.4\% & 16.4\% & 20.8\% & 24.8\% & 20.8\% & 16.8\% & 18.0\% & 28.0\% & 20.0\% & 18.4\% \\
 & swh & \cellcolor[HTML]{EA9999}8.4\% & 13.6\% & \cellcolor[HTML]{00FFFF}\textbf{30.0\%} & 8.4\% & 15.2\% & 12.8\% & 20.8\% & 19.2\% & 16.8\% & 24.8\% & 29.2\% & 18.1\% & 16.9\% \\
 & jpn & 15.6\% & 15.2\% & 12.0\% & 14.0\% & \cellcolor[HTML]{00FFFF}\textbf{30.0\%} & 27.2\% & 24.8\% & 22.8\% & 23.2\% & 24.0\% & 28.0\% & 21.5\% & 20.7\% \\
 & zho & 15.6\% & 21.2\% & 10.4\% & 10.4\% & 22.0\% & \cellcolor[HTML]{00FFFF}{\ul \textbf{40.8\%}} & 23.6\% & 20.4\% & 21.6\% & 25.2\% & 34.8\% & 22.4\% & 20.5\% \\
 & deu & 18.0\% & 18.4\% & 8.4\% & 16.0\% & 22.4\% & 24.0\% & \cellcolor[HTML]{00FFFF}34.0\% & 31.2\% & 27.6\% & 32.0\% & 38.4\% & 24.6\% & 23.6\% \\
 & fra & \textbf{23.2\%} & 19.2\% & 13.2\% & 14.0\% & 18.8\% & 20.0\% & 30.4\% & \cellcolor[HTML]{00FFFF}\textbf{35.2\%} & 30.8\% & 33.2\% & 37.6\% & 25.1\% & 24.0\% \\
 & rus & 17.2\% & 18.4\% & 10.8\% & 14.4\% & 15.2\% & 18.0\% & 29.6\% & 24.4\% & \cellcolor[HTML]{00FFFF}\textbf{38.0\%} & 29.6\% & 36.8\% & 22.9\% & 21.4\% \\
 & spa & 17.2\% & 18.4\% & 11.6\% & 14.0\% & 20.4\% & 22.8\% & 31.6\% & 31.6\% & 28.8\% & \cellcolor[HTML]{00FFFF}38.0\% & 36.4\% & 24.6\% & 23.3\% \\
\multirow{-10}{*}{\begin{tabular}[c]{@{}l@{}}Selective Freezing \\ First 8 Layers\end{tabular}} & eng & 18.8\% & 23.2\% & 19.6\% & \textbf{17.6\%} & 26.4\% & 29.6\% & \textbf{36.8\%} & 32.4\% & 36.4\% & {\ul \textbf{40.0\%}} & \cellcolor[HTML]{00FFFF}\textbf{42.0\%} & \textbf{29.3\%} & \textbf{28.1\%} \\ \midrule
 & ben & \cellcolor[HTML]{00FFFF}{\ul \textbf{26.4\%}} & 12.8\% & 11.6\% & 14.4\% & 13.6\% & 14.8\% & 19.6\% & 20.0\% & 20.0\% & 17.6\% & \cellcolor[HTML]{EA9999}17.2\% & 17.1\% & 16.2\% \\
 & tha* & 14.8\% & \cellcolor[HTML]{00FFFF}\textbf{34.0\%} & 12.0\% & 12.4\% & 15.6\% & 21.6\% & 25.2\% & 22.0\% & 20.4\% & 24.4\% & 32.4\% & 21.3\% & 20.1\% \\
 & swh & \cellcolor[HTML]{EA9999}9.2\% & 16.4\% & \cellcolor[HTML]{00FFFF}\textbf{22.8\%} & 5.6\% & 14.0\% & 12.4\% & 18.4\% & 23.6\% & 19.2\% & 20.4\% & 27.6\% & 17.2\% & 16.7\% \\
 & jpn & 16.0\% & 17.6\% & 12.0\% & 11.2\% & \cellcolor[HTML]{00FFFF}\textbf{27.2\%} & 28.8\% & 24.4\% & 23.2\% & 24.0\% & 24.4\% & 29.6\% & 21.7\% & 21.1\% \\
 & zho & 17.2\% & 17.2\% & 12.4\% & 12.0\% & 22.4\% & \cellcolor[HTML]{00FFFF}\textbf{34.8\%} & 29.6\% & 22.4\% & 27.6\% & 23.6\% & 37.2\% & 23.3\% & 22.2\% \\
 & deu & 12.8\% & 22.8\% & 14.4\% & 17.6\% & 20.0\% & 25.6\% & \cellcolor[HTML]{00FFFF}\textbf{36.0\%} & 29.6\% & 27.6\% & 32.8\% & 39.2\% & 25.3\% & 24.2\% \\
 & fra & 14.8\% & 24.8\% & 18.4\% & 12.0\% & 21.2\% & 21.2\% & 33.6\% & \cellcolor[HTML]{00FFFF}\textbf{37.2\%} & 32.0\% & \textbf{36.8\%} & 36.8\% & 26.3\% & 25.2\% \\
 & rus & 20.4\% & 19.6\% & 11.6\% & {\ul \textbf{18.8\%}} & 22.0\% & 19.6\% & 28.8\% & 25.2\% & \cellcolor[HTML]{00FFFF}38.4\% & 28.8\% & 32.0\% & 24.1\% & 22.7\% \\
 & spa & 20.0\% & 24.0\% & 17.6\% & 16.8\% & 18.0\% & 27.2\% & 33.6\% & 33.6\% & 29.6\% & \cellcolor[HTML]{00FFFF}34.0\% & 36.4\% & 26.4\% & 25.7\% \\
\multirow{-10}{*}{\begin{tabular}[c]{@{}l@{}}Selective Freezing \\ First 12 Layers\end{tabular}} & eng & 20.4\% & 24.0\% & 18.0\% & 16.4\% & 20.4\% & 26.4\% & 35.2\% & 30.0\% & {\ul \textbf{43.6\%}} & 32.4\% & \cellcolor[HTML]{00FFFF}{\ul \textbf{46.8\%}} & \textbf{28.5\%} & \textbf{26.7\%} \\ \midrule
 & ben & \cellcolor[HTML]{00FFFF}\textbf{24.0\%} & 13.6\% & \cellcolor[HTML]{EA9999}6.4\% & 10.4\% & 11.2\% & \cellcolor[HTML]{EA9999}7.6\% & 16.8\% & 16.0\% & 15.2\% & 13.6\% & \cellcolor[HTML]{EA9999}16.0\% & 13.7\% & 12.7\% \\
 & tha* & 11.6\% & \cellcolor[HTML]{00FFFF}\textbf{27.2\%} & 9.6\% & 10.4\% & 12.4\% & 15.6\% & 19.6\% & 14.4\% & 21.2\% & 19.6\% & 27.6\% & 17.2\% & 16.2\% \\
 & swh & \cellcolor[HTML]{EA9999}10.8\% & \cellcolor[HTML]{EA9999}10.8\% & \cellcolor[HTML]{00FFFF}\textbf{20.4\%} & 8.0\% & 11.6\% & 10.4\% & 18.0\% & 20.4\% & 14.8\% & 19.6\% & 21.2\% & 15.1\% & 14.6\% \\
 & jpn & 14.8\% & 13.6\% & 9.6\% & 6.0\% & \cellcolor[HTML]{00FFFF}\textbf{26.4\%} & 22.4\% & 23.2\% & 17.2\% & 14.8\% & 22.0\% & 26.8\% & 17.9\% & 17.0\% \\
 & zho & 12.8\% & 15.2\% & \cellcolor[HTML]{EA9999}6.0\% & 8.0\% & 15.6\% & \cellcolor[HTML]{00FFFF}\textbf{27.2\%} & 23.2\% & 16.0\% & 24.0\% & 21.6\% & 31.6\% & 18.3\% & 17.4\% \\
 & deu & \cellcolor[HTML]{EA9999}10.4\% & 19.6\% & 9.2\% & 9.6\% & 15.6\% & 20.4\% & \cellcolor[HTML]{00FFFF}\textbf{34.0\%} & 23.6\% & 24.4\% & 25.2\% & 34.8\% & 20.6\% & 19.3\% \\
 & fra & 18.4\% & 14.8\% & 12.0\% & 12.8\% & 14.4\% & 20.4\% & 25.6\% & \cellcolor[HTML]{00FFFF}\textbf{35.6\%} & 27.6\% & 30.4\% & 32.4\% & 22.2\% & 20.9\% \\
 & rus & 12.0\% & 18.0\% & 10.0\% & 12.4\% & 13.2\% & 20.0\% & 26.4\% & 22.8\% & \cellcolor[HTML]{00FFFF}27.6\% & 23.6\% & 29.2\% & 19.6\% & 18.8\% \\
 & spa & \cellcolor[HTML]{EA9999}11.2\% & 22.0\% & 14.0\% & \textbf{14.8\%} & 12.8\% & 20.4\% & 25.6\% & 29.2\% & 30.0\% & \cellcolor[HTML]{00FFFF}\textbf{30.8\%} & 32.0\% & 22.1\% & 21.2\% \\
\multirow{-10}{*}{\begin{tabular}[c]{@{}l@{}}Selective Freezing \\ First 16 Layers\end{tabular}} & eng & 16.0\% & 16.8\% & 12.4\% & 10.4\% & 17.6\% & 25.6\% & 31.2\% & 30.0\% & \textbf{34.0\%} & 26.0\% & \cellcolor[HTML]{00FFFF}\textbf{40.4\%} & \textbf{23.7\%} & \textbf{22.0\%} \\ \bottomrule
\end{tabular}}
\caption{Ablation study on the cross-lingual transfer performance on MGSM for Llama-3.1 (8B) fine-tuned with the selective freezing strategy varied on the frozen layers. ``XL'' denotes average on languages that were not fine-tuned. Diagonal entries in \sethlcolor{highlightBlue}\hl{blue highlights} correspond to source language performances. \sethlcolor{highlightRed}\hl{Red highlights} indicate decrease from pre-trained baseline. \textbf{Bold} and {\ul underline} respectively denote the best within group and within column. The (*) marks languages classified as low-resource in Flores-200.}
\label{tab:abl_layer_freeze_complete}
\end{table*}

\end{document}